\documentclass[lettersize,journal]{IEEEtran}
\usepackage{cite}
\usepackage{amsmath,amssymb,amsfonts}
\usepackage{amssymb} 
\usepackage{graphicx}
\usepackage{algorithm}
\usepackage{algpseudocode}
\usepackage{hyperref}
\usepackage{textcomp}
\usepackage[table,xcdraw]{xcolor}
\usepackage{colortbl}
\usepackage{xcolor}
\usepackage{booktabs}
\usepackage{array} 
\usepackage{subcaption}
\usepackage{float}
\usepackage{ifthen}
\usepackage{caption} 
\title{Efficient Fireworks Algorithm Equipped with an Explosion Mechanism based on Student's T-distribution}

\author{
Shipeng Cen,
Ying Tan
}

\begin{document}

\maketitle

\begin{abstract}
Many real-world problems can be transformed into optimization problems, which can be classified into convex and non-convex. Although convex problems are almost completely studied in theory, many related algorithms to many non-convex problems do not work well and we need more optimization techniques. As a swarm intelligence optimization algorithm, the Fireworks Algorithm(FWA) has been widely studied and applied to many real-world scenarios, even including large language model fine-tuning. But the current fireworks algorithm still has a number of problems. Firstly, as a heuristic algorithm, its performance on convex problems cannot match the SOTA results, and can even be said to be unsatisfactory; secondly, the sampling methods (explosion) of most FWA variants are still uniform sampling, which is actually inefficient in high dimensional cases. This work of ours proposes a new student's t-distribution based FWA(TFWA) with a solid theoretical foundation, which fully utilizes the advantage that student's t-distribution can adjust the parameters (degrees of freedom) and thus adjust the exploitation capability. We have fully experimented on mainstream benchmarks CEC2013 and CEC2017, which proves that TFWA not only becomes the strongest variant of the fireworks algorithm, but also achieves results comparable to SOTA on the test set, and its performance is far superior to that of the SOTA algorithm in some scenarios with a large number of extreme points.
\end{abstract}


\section{Introduction}
 In recent years, many evolutionary and swarm intelligence optimization algorithms have been proposed and widely applied across diverse fields, including engineering, logistics, artificial intelligence, and finance, where optimization problems are becoming increasingly complex\cite{b1,b2,b3,b4}. Algorithms like Genetic Algorithms (GA)\cite{c1}, Particle Swarm Optimization (PSO)\cite{c2}, Ant Colony Optimization (ACO)\cite{c3}, Differential Evolution (DE)\cite{c4}, and Fireworks Algorithm (FWA)\cite{c5} are particularly well-suited for solving complex, high-dimensional, and nonlinear problems. These methods are inspired by natural processes such as genetic evolution, social interactions among particles, and foraging behavior in ant colonies. Unlike traditional optimization methods, which can struggle with local optima or computational inefficiencies in complex landscapes, evolutionary and swarm-based algorithms offer flexibility, robustness, and the ability to effectively explore vast solution spaces. As a result, they have become essential tools in optimizing intricate systems across various real-world applications, and their development continues to evolve to meet new challenges.

Among them, FWAs have been widely studied and applied with great attention due to their simplicity, efficiency and parallelism. It is a heuristic algorithm that simulates firework explosions for sampling. In recent years, at the algorithmic level, many efficient variants such as EDEFWA\cite{c6}, LoTFWA\cite{c7}, MGFWA\cite{c8}, etc. have evolved through the improvement of various types of operators. On the application side, the fireworks algorithm even has been successfully applied to LORA fine-tuning\cite{c10} for large language models, achieving better results than the traditional zero-order algorithm.\cite{c9}

Nonetheless, the Fireworks Algorithm variants do not particularly stand out in terms of its optimization capabilities, mainly because its underlying sampling approach generally employs uniform sampling, which proves to be inefficient in high-dimensional spaces. In order to overcome this inherent weakness, this work proposes a student's t-distribution based explosive sampling approach by means of rigorous theoretical derivations from natural gradient descent methods. In addition to that, we propose a degree-of-freedom adjustment mechanism to further enhance the sampling capability. Sufficient experiments on CEC2013\cite{c11} and CEC2017\cite{c12} benchmarks prove that our method is not only a SOTA variant of the current fireworks algorithm variants, but also able to achieve a performance comparable to or even surpassing SOTA on major benchmarks. To the best of our knowledge, this is the first complete natural gradient descent framework based on student's t-distribution.

The article is structured as follows. In Section II, we introduce FWA, the differences among distributions and Natural Gradient Descent\cite{c27}. In Section III, we will introduce the whole TFWA algorithm and theoretical derivation. In section IV, we mainly conduct experimental comparisons, including excellent fireworks algorithm variants and SOTA optimization algorithms on benchmarks, and we also analyze all kinds of mechanisms of TFWA. In Section V, we summarize the whole work.

\section{Related works}
\subsection{Fireworks Algorithm}
The Fireworks Algorithm (FWA), introduced in 2010 by Professor Tan Ying and colleagues, is a swarm intelligence optimization algorithm inspired by the explosion process of fireworks. Initially designed as a novel approach to handle complex optimization problems, FWA quickly gained attention due to its unique mechanisms and flexibility. The current FWA variants have been used in very many optimization scenarios, including various task scheduling, matrix decomposition, and even LoRA fine-tuning of large language models, which have shown notable performance.

FWA consist of three crucial operators, namely explosion, mutation and selection. The \textbf{Explosion} operator is used to sample around a given point according to a certain rule, and all variants of the fireworks algorithm use almost exclusively hypercube sampling. The \textbf{Mutation} operator is based on explosive sampling and produces one or more new solutions by further analyzing and organizing the sampled solutions and fitness. Common types of mutation are Gaussian mutation, guiding vector mutation, and so on. The \textbf{Selection} operator is used to make choices among current sparks to produce the next generation of firework position, usually using roulette or greedy methods. It is easy to see that the explosion operator plays the most crucial role in these three operators.

Classical variants of fireworks algorithms include the early AFWA\cite{c13}, dynFWA\cite{c14}, GFWA\cite{c15} and the current mainstream LoTFWA\cite{c7}, MGFWA\cite{c8}, etc., with MGFWA being the best previous variant. There are also some works investigating the explosion operator, such as FWASSP\cite{c16} and SF-FWA\cite{c17}, but they just follow the sampling method of CMA-ES\cite{c18} and do not depart from the underlying Gaussian distribution. Our proposed TFWA bridges the research gap in this area and takes advantage of the multiple swarms of fireworks algorithms.

\subsection{Differences of Sampling Among Distributions}
We will start with normal Gaussian and Cauchy distribution. The well-known standard Gaussian function (one-dimensional) is represented by the following probability density expression:
\begin{equation}
    f_{g}(x) = \frac{1}{\sqrt{2\pi}}exp(-\frac{x^2}{2})
\label{eq1}
\end{equation}
Here, 0 is the mean (expectation), and 1 is the standard deviation. This function describes the shape of the standard Gaussian distribution, where the majority of the data is concentrated around the mean, and the probability decreases as the distance from the mean increases.
The probability density function of the standard Cauchy distribution is given by the following expression:
\begin{equation}
    f_{c}(x) = \frac{1}{\pi(1+x^2)}
\label{eq2}
\end{equation}
This function describes the characteristics of the Cauchy distribution, where the probability density decreases as $x$ increases, but its tails decay more slowly compared to the Gaussian distribution, leading to heavier tails.
\begin{figure}
    \centering
    \includegraphics[width=1.0\linewidth]{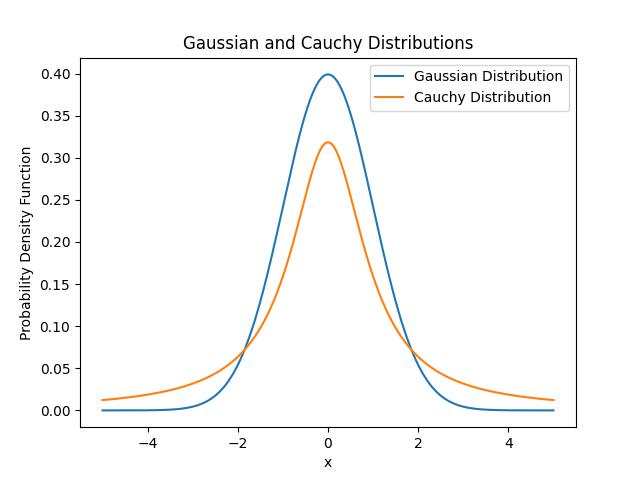}
    \caption{Standard Gaussian and Cauchy Distribution}
    \label{fig:1}
\end{figure}
The characteristics of Gaussian and Cauchy distributions are intuitively clear geometrically in Fig \ref{fig:1}. We can provide a mathematical proof through the following equation, indicating that the probabilities of their tail distributions are infinitesimal quantities of completely different orders.

In the following equations, we assume $a \geq 0$.

\begin{equation}
\begin{split}
Tail_{Gaussian}(a) &= \int_a^{+\infty} \frac{1}{\sqrt{2\pi}}exp(-\frac{x^2}{2}) \,dx \\
    &\leq \frac{1}{\sqrt{2\pi}} \int_a^{+\infty} \frac{x}{a}exp(-\frac{x^2}{2}) \,dx  \\
    &\leq \ \frac{1}{2a}e^{-\frac{a^2}{2}}
\end{split}
\label{eq3}
\end{equation}
\begin{equation}
\begin{split}
Tail_{Cauchy}(a) &= \int_a^{+\infty} \frac{1}{\pi(1+x^2)} \,dx \\
                &\geq \frac{a}{\pi(1+a^2)} \\
                &\geq O(\frac{1}{a})
\end{split}
\label{eq4}
\end{equation}
It's easy to see that $lim_{a\rightarrow\infty} \frac{Tail_{Cauchy}(a)}{Tail_{Gaussian}(a)} = +\infty$. Therefore, there is a fundamental difference in the tail distributions of these two. The Gaussian distribution has undergone extensive research and application within CMA-ES\cite{c18}. Regarding the Cauchy distribution, there is some work. But the majority of it is concentrated on the mutation operators\cite{c29,c30,c31,c32} rather than \textbf{strict} sampling distribution like CMA-ES in evolutionary computation. There is one work\cite{c28} viewing Cauchy distribution from Natural gradient descent. But its experiments don't go far enough, due to its weak baseline and insufficient functions.

The tail distributions of these two distributions are quantitatively different, and we hope that our FWA can incorporate the strengths of both; using a single distribution is not perfect. In fact, the Cauchy distribution and the Gaussian distribution happen to be the most special cases of the student's t-distribution. First, the probability density function of the student's t-distribution is as follows:
\begin{equation}
\begin{split}
    f_{t}(x)= \frac{\Gamma(\frac{v}{2} + \frac{d}{2})}{\Gamma(\frac{v}{2})}\frac{{|\Sigma|}^{-\frac{1}{2}}}{(v\pi)^{\frac{d}{2}}}[1 + \frac{1}{v}(x-\mu)^{T}\Sigma^{-1}(x-\mu)]^{-\frac{d+v}{2}}
\end{split}
\label{eq5}
\end{equation}
where $v$ represents degrees of freedom, $d$ is the dimension of the vector $x$(variable of our problem), $\Sigma$ represents scale matrix, $\Gamma(x)=\int_{t=0}^{\infty}{t^{x-1}e^{-t}}dt$. For the special case when $d=1$, $\mu = 0$ and $\sigma = 1$, we have the following density function:
\begin{equation}
\begin{split}
    f_{t}(x)= \frac{\Gamma(\frac{v}{2} + \frac{1}{2})}{\Gamma(\frac{v}{2})}\frac{1}{(v\pi)^{\frac{1}{2}}}[1 + \frac{1}{v}x^2]^{-\frac{1+v}{2}}
\end{split}
\label{eq6}
\end{equation}

Note that as the variable $v$ tends to positive infinity, the right side of equation (6) will approach equation (1), which is not difficult by using $e^x \sim 1 + x$ when $x\rightarrow0$. And as the variable $v$ tends to 1, the right side of equation (6) will approach equation (2).

According to the above discussion, we find that the student's t-distribution is a generalization of both the Cauchy distribution and the Gaussian distribution. In other words, we can adjust the parameters of the student's t-distribution, $v$ in equation(5), to achieve sampling characteristics that cannot be achieved solely with the Cauchy or Gaussian distributions.

Some important conclusions about student's t-distribution in equation (5) we may use:
\begin{equation}
    E_{x \sim f_t(x)}[x]=\mu,
    Var_{x \sim f_t(x)}[x] = \frac{v}{v-2}\Sigma
\label{eq7}
\end{equation}
It should be noted that the variance of the t-distribution becomes infinite when the degrees of freedom $v$ are less than 2. It must be noted that the matrix in equation (5) is not a covariance matrix. Therefore, in practical applications, it is typically required that the degrees of freedom be at least 2. Due to the special nature of the T-distribution, which will be used as the basic distribution for our explosion sampling, we will present the theoretical background below and in this work, we will use $T(m, C, df)$ to represent a  student's t-distribution.
\subsection{Natural Gradient Descent}
Gradient descent has a significant drawback, namely, it converges very slowly for ill-conditioned problems. For instance, in the case of a quadratic function
\begin{equation}
    f(x)=\frac{1}{2}x^TAx
\label{eq8}
\end{equation}
where the matrix $A$ has a large condition number, the algorithm may exhibit a slow, oscillating convergence, requiring a considerable number of iterations to reach the optimal point. Even with precise line search, finding the extremum in the gradient direction at each step, numerous iterations may still be necessary to reach the optimum. 

Since the Fisher information matrix can be proven to be an unbiased estimate of the Hessian matrix, one mainstream interpretation of the theoretical framework for CMA-ES\cite{c18} is to obtain sampled information , estimate the inverse of the Fisher information matrix, and then use it to correct the information of the underlying gradient. This process yields the final parameter update gradient.\cite{c19} The specific method for estimating and using the Fisher information matrix is as follows:
\begin{equation}
    F(\theta)=E_{x \sim p(x|\theta)}[\nabla \log p(x|\theta) \nabla \log p(x|\theta)^T  ]
\label{eq9}
\end{equation}
First our goal is minimize or maximize the following equation:
\begin{equation}
    J(\theta) = E_{x\sim p(x|\theta)}[f(x)] = \int_{x}{f(x)p(x|\theta)}dx
\label{eq10}
\end{equation}
By using (9), we can calculate our gradients adjusted by the Fisher information.
\begin{equation}
    \nabla J(\theta)=\int_{x}{f(x)\nabla p(x|\theta)}dx
\label{eq11}
\end{equation}
\begin{equation}
\begin{split}
    \tilde{\nabla} J(\theta) &=F(\theta)^{-1}\nabla J(\theta) \\
    &=\int_{x}{F(\theta)^{-1}f(x)\nabla p(x|\theta)}dx \\
    &=E_{x\sim p(x|\theta)}[f(x)F(\theta)^{-1}\nabla \log p(x|\theta)]\\
    & \approx \sum_{i=1}^{\lambda}\frac{f(x_i)}
    {\lambda}F(\theta)^{-1}\nabla \log p(x_{i}|\theta)
\end{split}
\label{eq12}
\end{equation}
As the underlying distribution has been replaced, we can no longer use the updates for
the mean and covariance from the Gaussian distribution. It requires us to derive them anew from the ground up.

\section{Proposed Algorithm}
In this section, we will employ the framework of natural gradient descent to compute the updates for the mean and covariance of the student's t-distribution in explosion of TFWA.  We will strive to provide the derivation steps and list the necessary references. We will start from equation ($5$).
\begin{equation}
    \begin{split}
    ln[p(x)]  &= ln[\frac{\Gamma(\frac{v}{2} + \frac{d}{2})}{\Gamma(\frac{v}{2})}] - \frac{1}{2}ln(|\Sigma|)  \\
    & - \frac{d+v}{2}ln(1 + \frac{1}{v}(x-\mu)^{T}\Sigma^{-1}(x-\mu))
    \end{split}
\label{eq13}
\end{equation}
Notice, where $M_{i,j}$ is algebraic cofactor of element $\Sigma_{i,j}$ and $tr(A)$ represents finding the trace of the matrix $A$:
\begin{equation}
    \begin{split}
        \frac{\partial{ln(|\Sigma|)}}{\partial{\theta}} &= \frac{1}{|\Sigma|}*\sum_{i,j}{M_{i,j}\frac{\partial{\Sigma_{i,j}}}{\theta}} \\
        &= \frac{1}{|\Sigma|}*tr(\frac{\partial{\Sigma}}{\partial{\theta}}*M^{T})=tr(\frac{\partial{\Sigma}}{\partial{\theta}}*\Sigma^{-1})
    \end{split}
\label{eq14}
\end{equation}
We have:
\begin{equation}
\begin{split}
    & \frac{\partial{ln[p(x)]}}{\partial{\theta}}\\ 
    &=-\frac{1}{2}tr(\frac{\partial{\Sigma}}{\partial{\theta}}\Sigma^{-1})-\frac{d+v}{2}\frac{1}{v+(x-u)^{T}\Sigma^{-1}(x-u)} \\
    & *[-2(\frac{\partial{u}}{\partial{\theta}})^{T}\Sigma^{-1}(x-u)+(x-u)^{T}\frac{\partial{\Sigma^{-1}}}{\partial{\theta}}(x-u)]
\end{split}
\label{eq15}
\end{equation}
Now we want to calculate Fisher Information Matrix $F$, where $F_{i,j}=E_{x}[\nabla ln[p(x|\theta_{i})]^{T}\nabla ln[p(x|\theta_{j})]^{T}]$, after which we can get the true gradient from $g = F^{-1}(\theta)\nabla ln[p(x|\theta)]$
With equation $(15)$, we can calculate the following thing:
\begin{equation}
\begin{split}
    F_{ij}[\theta]= & E_{x}[\frac{1}{4}tr(\frac{\partial{\Sigma}}{\partial{\theta_{i}}}*\Sigma^{-1})tr(\frac{\partial{\Sigma}}{\partial{\theta_{j}}}*\Sigma^{-1}) 
    \\ &+(\frac{d+v}{v+s})^{2}(\frac{\partial{\mu}}{\partial{\theta_{i}}})^{T}\Sigma^{-1}(x-u)(\frac{\partial{\mu}}{\partial{\theta_{j}}})^{T}\Sigma^{-1}(x-u)
    \\ & +\frac{1}{4}(\frac{d+v}{v+s})^{2}(x-u)^{T}\frac{\partial{\Sigma^{-1}}}{\partial{\theta_{i}}}(x-u)
    \\ &*(x-u)^{T}\frac{\partial{\Sigma^{-1}}}{\partial{\theta_{j}}}(x-u)
    \\ &+ \frac{d+v}{2(v+s)}tr(\frac{\partial{\Sigma}}{\partial{\theta_{i}}}*\Sigma^{-1})(x-u)^{T}\frac{\partial{\Sigma^{-1}}}{\partial{\theta_{j}}}(x-u)]
\end{split}
\label{eq16}
\end{equation}
Where $s = (x-u)^{T}\Sigma^{-1}(x-u)$, and we have split the $F_{i,j}(\theta)$ into 4 parts to calculate as shown in Equation$(16)$. And we can calculate all the parts to get the following consequence. You can refer to \textbf{supplementary material} for the complete mathematical derivation.
\begin{equation}
\begin{split}
    F_{i,j}(\theta) &=\frac{1}{2}\frac{d+v}{d+v+2}tr(\frac{\partial \Sigma}{\partial \theta_{i}}\Sigma^{-1}\frac{\partial \Sigma}{\partial \theta_{j}}\Sigma^{-1}) \\
    & - \frac{1}{2(d+v+2)}tr(\frac{\partial \Sigma}{\partial \theta_{i}}\Sigma^{-1})tr(\frac{\partial \Sigma}{\partial \theta_{j}}\Sigma^{-1}) \\
    & +\frac{d+v}{d+v+2}(\frac{\partial{\mu}}{\partial{\theta_{i}}})^{T}\Sigma^{-1}(\frac{\partial{\mu}}{\partial{\theta_{j}}})
\end{split}
\label{eq17}
\end{equation}

According to the NGD rules introduced in Equation \ref{eq12}, We get(you can refer to supplementary material for details):
\begin{equation}
    F_{C}^{-1}\nabla_{C}lnP(x)=\frac{d+v+2}{d+v}(\frac{d+v}{v+s}(x-u)(x-u)^{T}-\Sigma)
\label{eq18}
\end{equation}
\begin{equation}
    F_{m}^{-1}\nabla_{m}lnP(x)=\frac{d+v+2}{s+v}(x-u)
\label{eq19}
\end{equation}
In the classical CMA-ES algorithm, rank weights obtained based on fitness are used. In this work, we fuse the rank-based weights with the weights based on Equations $(18)$ and $(19)$ as the mean and covariance update weights for the student's t-distribution in TFWA. 

After obtaining the way the underlying student's t-distribution is maintained and updated, we propose \textbf{Algorithm 1}, which presents a complete demonstration of the TFWA's explosion sampling approach, with all the parameters that may be used listed in Table \ref{table:1}. As we mentioned before, an important reason for choosing the student's t-distribution is that it is possible to balance the local exploitation and global exploration of the sampling process by adjusting the degrees of freedom of the student's t-distribution.

We involve a very important mechanism in Algorithm 1 called the \textbf{AdjusDegreeOfFreedom}, see \textbf{Algorithm 2} for the definition. The idea behind the design of this function is that if a better solution is obtained for a particular firework sampling at the current degrees of freedom, then in general there exists a region of non-zero measurements that is better compared to the current solution, and we will increase the degrees of freedom to make the student's t-distribution closer to a Gaussian distribution, at which point the overall sampling way will be closer to CMA-ES, reinforcing the ability to exploit; however, if the sampling does not result in a better solution, then we will keep the degrees of freedom unchanged for global exploration capabilities rather than continue to increase them.

\begin{algorithm}
\caption{T-distribution Based Explosion of FWA} 
\begin{algorithmic}[1] 
    \Require Object function $f$, current T-distribution T(mean $m$, covariance $C$, degree of freedom $df$), factor to update degree of freedom $factor$, Num of sparks $\lambda$, $p_{c}$, $p_{s}$, current best fitness $f_{best}$;  
\Ensure  New T-distribution's mean $m'$, covariance $C'$, degree of freedom $df'$, $pc'$, $ps'$ 
\State Initialize variables\\ 
\quad $w_{i}\gets max[log(0.5 + \frac{\lambda}{2}) - (1 + i), 0],0 \leq i \leq \lambda - 1$ \\
\quad Normalize w\\
\quad $distribuiton \leftarrow T(m, C, df)$

\State Explosion for sparks \\
\quad $pop \leftarrow distribution.sample(number = \lambda)$ \\
\quad $fit \leftarrow f(pop) $ \\
\quad $w'_{i} \leftarrow \frac{dim + df + 2}{df + pop[i]^TC^{-1}pop[i]},0 \leq i \leq \lambda - 1$ \\
\quad $w'' \leftarrow \frac{w \odot w'}{sum(w \odot w')}$ 
\State Update parameters \\
\quad $m' \leftarrow w'' * pop$ \\
\quad $cov_{i} \leftarrow (pop_{i}-m_{i})^T(pop_{i}-m_{i}),0 \leq i \leq \lambda - 1$\\
\quad $C' \leftarrow (1 - c_{1a} - c_{\mu} * sum(w'')) * C  + c_{1} * p_{c}p_{c}^{T} + \sum_{i=1}^{\lambda}{\frac{w''_{i} * c_{\mu}}{scale^2}*cov_{i}},0 \leq i \leq \lambda - 1$ \\
\quad $p_{c}' \leftarrow (1-c_{c})*p_{c} + c_{cn} * H * (m' - m )$\\
\quad $p_{s}' \leftarrow (1-c_{s})*p_{s} + c_{sn} * C^{-1}(m'-m)$\\
\quad $scale \leftarrow scale * exp(min(1, \frac{c_{n}}{2}(\frac{\sum{p_{s}'^2}}{dim}-1)))$ \\
\quad $df' \leftarrow AdjustDegreeOfFreedom(df, fit, f_{best}, factor)$
\State \textbf{Return} $m', C', df', pc', ps'$ \ 
\end{algorithmic}
\end{algorithm}

\begin{algorithm}
\caption{AdjustDegreeOfFreedom} 
\begin{algorithmic}[1] 
\Require degree of freedom $df$,  current fitness $fit$, current best fitness $f_{best}$, factor to update degree of freedom $factor$;  
\Ensure new degree of freedom $df'$
\If{$fit < f_{best}$}
    \State $df' \leftarrow max(df * factor, df + 1)$
    \State $df' \leftarrow min(df', \frac{INTMAX}{2})$
\Else
    \State $df' \leftarrow df$
\EndIf \\
$f_{best} \leftarrow fit$ \# we only compare with the best fit of the last generation
\State \textbf{Return} $df'$
\end{algorithmic}
\end{algorithm}

\begin{table}[ht]
\centering
\caption{Important Parameters used in TFWA}
\begin{tabular}{|c|c|}
\hline
\textbf{Name} & \textbf{value}  \\
\hline
$\mu_{eff} $ &  $\frac{(\sum{w})^2}{\sum{w^2}}$\\
\hline
$c_{c}$ &  $\frac{4+\frac{\mu_{eff}}{dim}}{4 + dim + 2 * \frac{\mu_{eff}}{dim}}$\\
\hline
$c_{s} $ &   $\frac{2 + \mu_{eff}}{dim + \mu_{eff} + 5}$\\
\hline
$c_{cn} $ &   $\frac{\sqrt{c_{c}(2-c_{c})\mu_{eff}}}{scale}$\\
\hline
$c_{sn} $ &   $\frac{\sqrt{c_{s}(2-c_{s})\mu_{eff}}}{scale}$\\
\hline
$H $ &   $INT(\frac{\sum(p_{s}^2)}{dim * (1-(1-c_{s})^{2{\#eval}+1})} \leq {2 + \frac{4}{dim+1}})$\\
\hline
$c_{1} $ &  $\frac{2}{(dim+1.3)^2 + \mu_{eff}}$ \\
\hline
$c_{1a} $ &   $c_{1}(1-(1-H^2)c_{c}(2-c_{c}))$\\
\hline
$c_{\mu} $ &  $min(1-c_{1}, \frac{2(\mu_{eff} + \frac{1}{\mu_{eff}} - 2)}{\mu_{eff} + (dim+2)^2})$ \\
\hline
\end{tabular}
\label{table:1}
\end{table}

\begin{algorithm}
\caption{Loser-out Tournament Mechanism in LoTFWA} 
\begin{algorithmic}[1] 
\Require Object function $f$,  number of fireworks $N$, small positive constant $EPS$, max generation $g_{max}$, current generation $g$  
\For{$i = 1$ to $N$}
    \If{$f(\mathbf{m}_i^g) < f(\mathbf{m}_i^{g-1}) - EPS $ }
        \State $\delta_i^g \gets f(\mathbf{m}_i^{g-1}) - f(\mathbf{m}_i^g)$
    \EndIf
    \If{$\delta_i^g \cdot (g_{\max} - g) < f(\mathbf{X}_i^g) - \min_j \{ f(\mathbf{m}_j^g) \}$}
        \State $m_{i} \sim Uniform(\frac{lb}{2}, \frac{ub}{2})$
        \State $Firework_{i} \leftarrow T(mean = m_{i}, cov = I, df = factor[i]])$
        \State $y_{i} \leftarrow f(m_{i})$
    \EndIf
\EndFor
\end{algorithmic}
\end{algorithm}

\begin{algorithm}
\caption{TFWA} 
\begin{algorithmic}[1] 
\Require  Object function $f$, number of sparks $\lambda$, number of fireworks $N$, factor array to update degree of freedom $factors$  
\Ensure best solution $x^*$, best fitness $y^*$ 

\For{$i = 1$ to $N$}
    \State $m_{i} \sim Uniform(\frac{lb}{2}, \frac{ub}{2})$
    \State $Firework_{i} \leftarrow T(mean = m_{i},  cov = I, df = factors[i]])$
    \State $y_{i} \leftarrow f(m_{i})$
    \State $p_{c_{i}}, p_{s_{i}},scale_{i} \leftarrow 0, 0, f.ub - f.lb$
\EndFor

\While{terminal criterion is not met}
\For{$i = 1$ to $N$}
    \State $Firework_{i}$ explode according to \textbf{Algorithm 1}
\EndFor

Restart fireworks according to \textbf{Algorithm 3} 

\EndWhile

\State \textbf{Return} $x^*, y^*$ \ 
\end{algorithmic}
\end{algorithm}
At the same time we follow the restart strategy in LoTFWA\cite{c7}, you can refer to the pseudo-code of \textbf{Algorithm 3}, which is mainly designed to improve the efficiency of the evaluation and to reduce useless evaluations by restarting unpromising fireworks. In addition, a very important feature in TFWA is the elimination of the mutation operation for generating guiding sparks, the reason for this is that in the framework of uniformly distributed sampling, the quality of the generated solutions is not high, the guiding vector indicate the direction of the function descent, which makes the quality of the solutions that come out of the mutation much higher than that generated by the explosions; however, in the sampling framework of the current student's t-distributed sampling framework, the quality of the generated solutions is greatly improved, and perhaps in a certain generation, the choice of bootstrap fireworks in the middle can temporarily improve the quality of the solution, but in the long run it will affect the updating of the whole distribution and harm the quality of the final solution. Eventually we get \textbf{TFWA} and you can refer to \textbf{Algorithm 4} for details.

\section{Experiments and Analysis}
For the convenience of testing, we chose two benchmark function test sets, CEC2013 and CEC2017, and our experiments mainly chose the cases of 30 and 50 dimensions. On each test benchmark, the number of evaluations in dimension $d$ is limited to $10000d$ and each experiment is repeated 30 times. Our experiments consist of two parts, a comparison with the mainstream good variants of FWA and a comparison with other types of excellent optimization algorithms. Sufficient experiments prove that TFWA is not only the best current variant of the fireworks algorithm (especially the qualitative breakthrough in the single-mode problem), but also achieves a performance that is comparable to SOTA or even surpasses SOTA in some scenarios. In the following experiments, we choose our number of fireworks $N=2$, factor array to update degree of freedom $[1.05, 10]$, initial degree of freedom of student's t-distribution $df = 5$, number of sparks $\lambda = \frac{10d}{N}$.
\subsection{Compare with FWA variants}
We chose two variants of the algorithm, LoTFWA and MGFWA, where the latter further improves on the former by improving the mechanism of guiding sparks to more fully utilize the information from each generation. 

To be more persuasive, we not only compare the algorithms together and do a one-to-one wilcoxon rank sum test where we count the win/lose/tie among algorithms, displayed in Table \ref{table:2}, \ref{table:3}, \ref{table:4} .

Each element in Table \ref{table:2} is meant to be the result of a 1-to-1 wilcoxon rank sum test between the algorithm for the corresponding row and the algorithm for the corresponding column. For example, 4/17/7 in the second column of the first row means that on the 30-dimensional CEC2013 benchmark, LoTFWA has a lead in 4 functions, a lag in 17 functions, and a tie in 7 functions compared to MGFWA. From Tables  \ref{table:2}, \ref{table:3}, \ref{table:4} we can see that TFWA has made great progress compared to LoTFWA and MGFWA, and has become the strongest version of the fireworks algorithm, both on the CEC2013 and CEC2017 benchmarks. We will now discuss the experimental results carefully.

One of the most obvious results is that TFWA achieves almost globally optimal solutions on the unimodal problems in both test sets, well ahead of the two FWA variants of the algorithm. We have mentioned that when the degrees of freedom of the underlying T-distribution are increasing, the T-distribution gradually turns into a Gaussian distribution, and thus the whole algorithm will be equipped with the local mining ability of CMAES, so it can solve the convex problems readily.

On the other hand, TFWA is also ahead of the original two variants of the algorithm in the multimode problem (especially at CEC2017), which is mainly attributed to the fact that high-dimensional spaces are not suitable for uniform sampling of hypercube explosions, as sampling is not efficient in this way. It is worth noting that both LoTFWA and MGFWA utilize the mechanism of guiding vectors to estimate the direction of the local descent of the function. The mechanism of bootstrap vectors is very useful both for the previous FWA variants in unimodal and multimodal modes, but this role is based on the weak local sampling ability. Since student's t-distribution can be well adapted to the whole problem and its local sampling ability is much stronger than the previous approach, the bootstrap vector mechanism becomes a shortcoming at this point, so we choose to remove it.

\begin{table}[h]
    \centering
    \scriptsize
    \caption{Win/Lose/Tie Comparison Among LoTFWA, MGFWA and TFWA on CEC2013 and CEC2017, dim = 30}
    \resizebox{\linewidth}{!}{ 
    \begin{tabular}{lcccccc}
        \toprule
        & \multicolumn{3}{c}{CEC2013} & \multicolumn{3}{c}{CEC2017} \\
        \cmidrule(lr){2-4} \cmidrule(lr){5-7}
        & LoTFWA & MGFWA & TFWA
        & LoTFWA & MGFWA & TFWA \\
        \midrule
        LoTFWA
        & - &  4/17/7 & 5/21/2
        & - & 6/13/11 & 2/26/2 \\
        
        MGFWA
        & 17/4/7 & - & 5/19/4
        & 13/6/11 & - & 0/27/3   \\
        
        TFWA
        & 21/5/2 & 19/5/4  & -
        & 26/2/2  & 27/0/3  & - \\
        \bottomrule
    \end{tabular}
    }
    \label{table:2}
\end{table}

\begin{figure}[h]
  \centering
  \begin{subfigure}[b]{0.4\textwidth}
    \includegraphics[width=\textwidth]{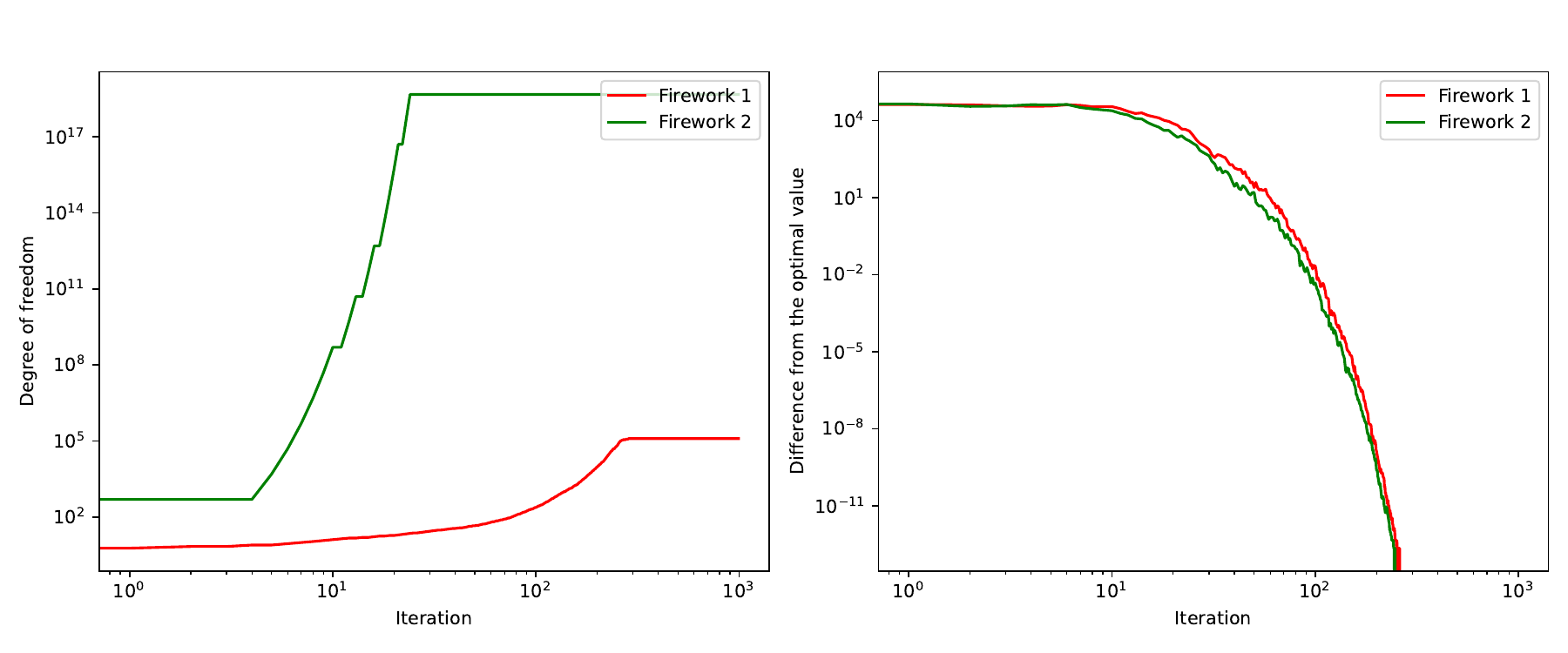}
    \caption{Sphere Function}
    \label{fig:sub1}
  \end{subfigure}
  
  \begin{subfigure}[b]{0.4\textwidth}
    \includegraphics[width=\textwidth]{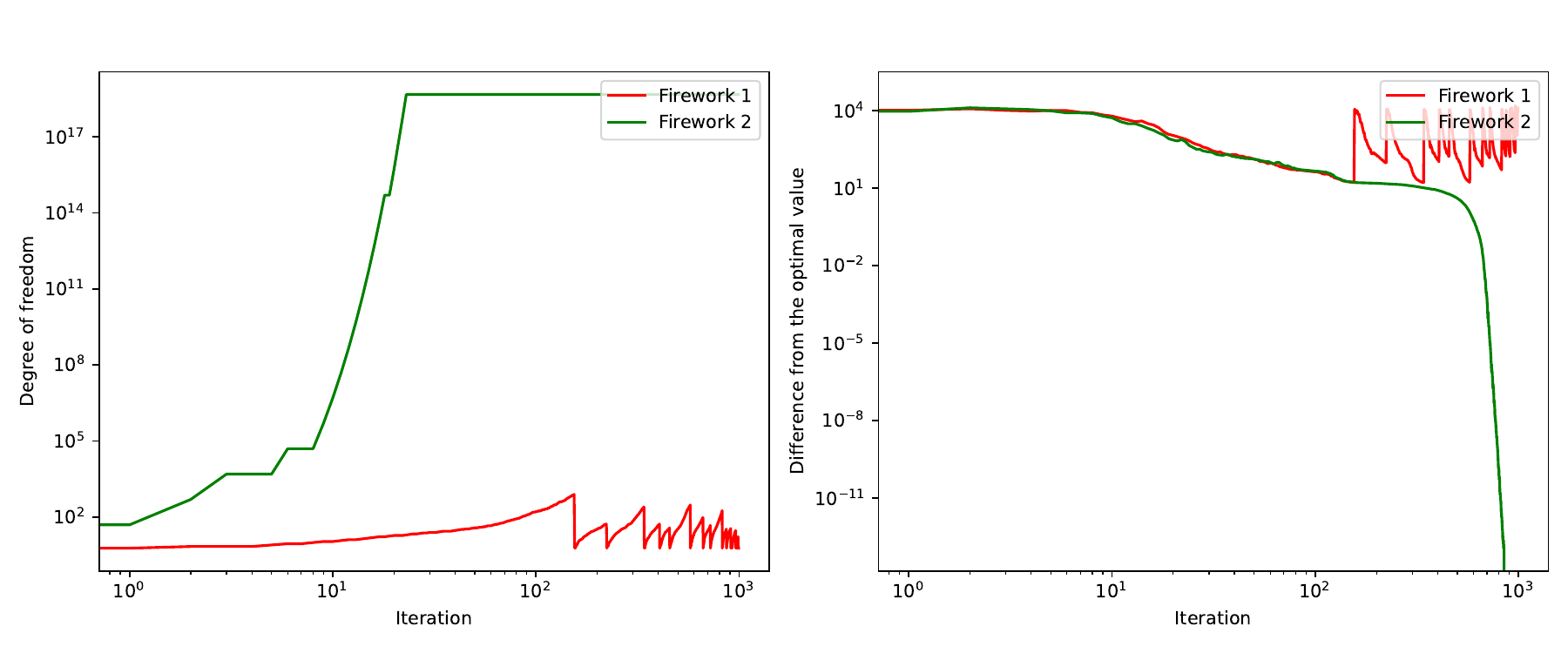}
    \caption{Rotated Rosenbrock’s Function}
    \label{fig:sub2}
  \end{subfigure}

  \begin{subfigure}[b]{0.4\textwidth}
    \includegraphics[width=\textwidth]{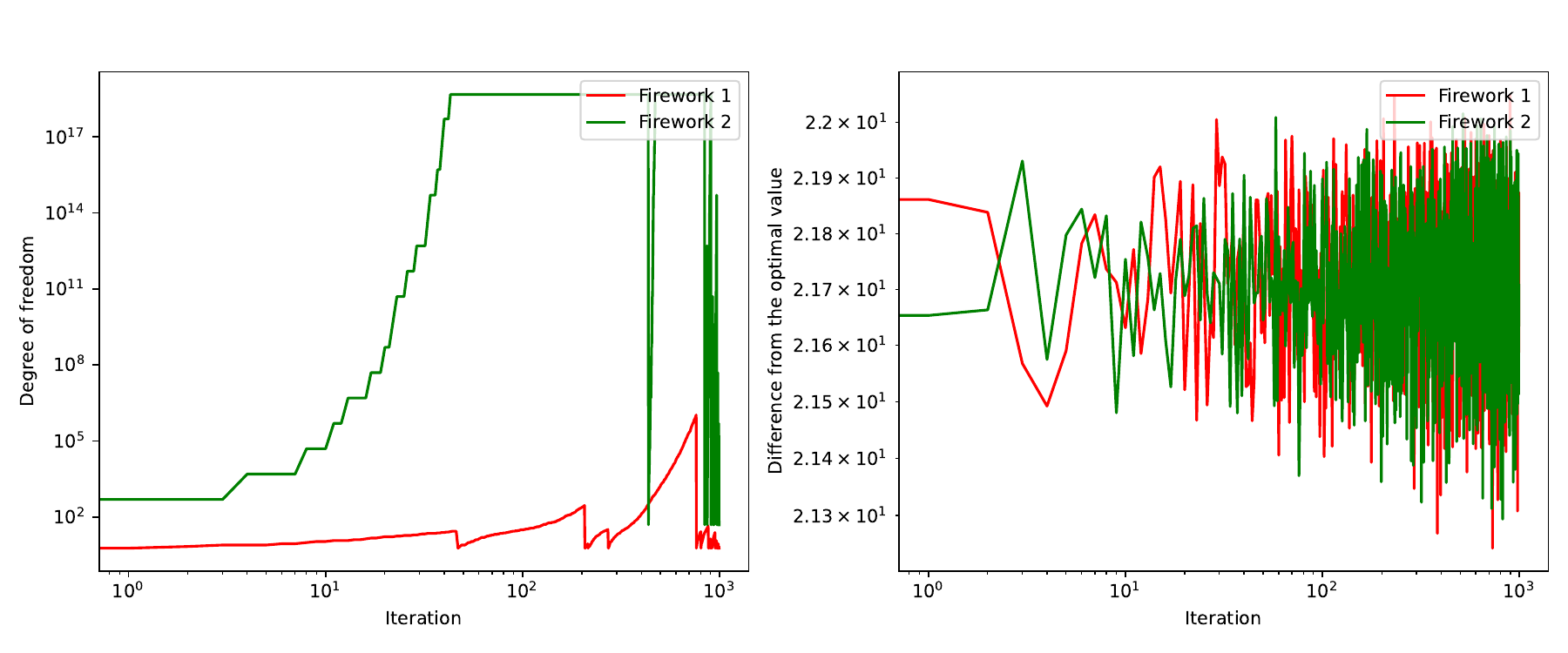}
    \caption{Rotated Ackley’s Function}
    \label{fig:sub3}
  \end{subfigure}
  
  \begin{subfigure}[b]{0.4\textwidth}
    \includegraphics[width=\textwidth]{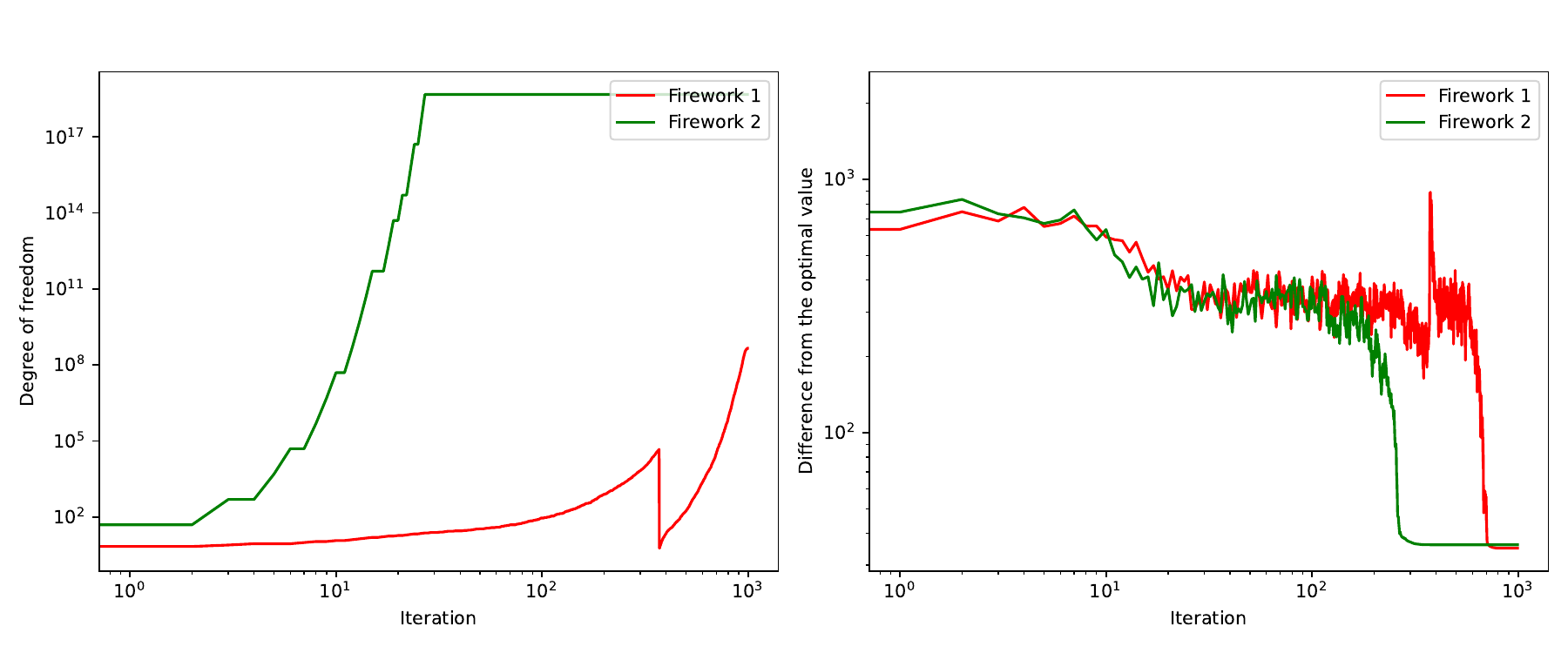}
    \caption{Lunacek Bi$\_$Rastrigin Function }
    \label{fig:sub4}
  \end{subfigure}
  
  \caption{Visualization of TFWA optimization progress on functions of CEC2013 benchmark}
  \label{fig:2}
\end{figure}

\begin{table*}[h]
    \centering
    \fontsize{7pt}{7.3pt}\selectfont 
    \caption{Comparing LoTFWA, MGFWA and TFWA on CEC2013, dim = 30}
    \begin{tabular}{|c|c|c|c|c|c|c|}
        \hline
\textbf{idx} & \textbf{LoTFWA.mean} & \textbf{LoTFWA.std} & \textbf{MGFWA.mean} & \textbf{MGFWA.std} & \textbf{TFWA.mean} & \textbf{TFWA.std} \\
\hline
1 & \textbf{0.000e+00} & \textbf{7.190e-14} & \textbf{0.000e+00} & \textbf{1.245e-13} & \textbf{0.000e+00} & \textbf{0.000e+00} \\
\hline
2 & 1.211e+06 & 5.309e+05 & 1.589e+06 & 5.255e+05 & \textbf{0.000e+00} & \textbf{0.000e+00} \\
\hline
3 & 1.946e+07 & 1.632e+07 & 6.259e+06 & 7.821e+06 & \textbf{0.000e+00} & \textbf{0.000e+00} \\
\hline
4 & 2.132e+03 & 7.903e+02 & 1.363e+03 & 5.134e+02 & \textbf{0.000e+00} & \textbf{0.000e+00} \\
\hline
5 & 3.452e-03 & 4.780e-04 & 6.916e-03 & 1.538e-03 & \textbf{0.000e+00} & \textbf{1.249e-07} \\
\hline
6 & 1.455e+01 & 5.999e+00 & 1.482e+01 & 1.740e-01 & \textbf{0.000e+00} & \textbf{0.000e+00} \\
\hline
7 & 4.979e+01 & 8.795e+00 & 2.752e+01 & 8.428e+00 & \textbf{3.270e-04} & \textbf{1.761e-03} \\
\hline
8 & \textbf{2.084e+01} & \textbf{8.804e-02} & 2.087e+01 & 6.271e-02 & 2.094e+01 & 6.130e-02 \\
\hline
9 & 1.459e+01 & 2.230e+00 & 1.002e+01 & 1.692e+00 & \textbf{4.100e+00} & \textbf{1.425e+00} \\
\hline
10 & 4.773e-02 & 3.139e-02 & 3.031e-02 & 1.785e-02 & \textbf{0.000e+00} & \textbf{0.000e+00} \\
\hline
11 & 6.357e+01 & 1.111e+01 & 2.643e+01 & 5.742e+00 & \textbf{5.240e+00} & \textbf{1.582e+00} \\
\hline
12 & 6.829e+01 & 1.189e+01 & 2.579e+01 & 6.390e+00 & \textbf{4.444e+00} & \textbf{1.578e+00} \\
\hline
13 & 1.340e+02 & 2.565e+01 & 6.073e+01 & 1.126e+01 & \textbf{5.413e+00} & \textbf{3.690e+00} \\
\hline
14 & 2.447e+03 & 2.819e+02 & 2.458e+03 & 2.884e+02 & \textbf{1.048e+03} & \textbf{4.511e+02} \\
\hline
15 & 2.761e+03 & 2.472e+02 & 2.391e+03 & 2.980e+02 & \textbf{5.424e+02} & \textbf{1.415e+02} \\
\hline
16 & \textbf{5.381e-02} & \textbf{1.879e-02} & 5.396e-02 & 1.669e-02 & 1.596e+00 & 1.125e+00 \\
\hline
17 & 6.215e+01 & 8.253e+00 & 5.592e+01 & 5.189e+00 & \textbf{3.447e+01} & \textbf{1.102e+00} \\
\hline
18 & 6.575e+01 & 7.405e+00 & \textbf{5.878e+01} & \textbf{5.722e+00} & 8.404e+01 & 6.489e+01 \\
\hline
19 & 3.235e+00 & 7.663e-01 & \textbf{2.476e+00} & \textbf{3.645e-01} & 2.637e+00 & 3.683e-01 \\
\hline
20 & 1.354e+01 & 9.475e-01 & \textbf{1.304e+01} & \textbf{1.214e+00} & 1.315e+01 & 2.331e+00 \\
\hline
21 & \textbf{2.000e+02} & \textbf{9.340e-02} & 2.133e+02 & 3.399e+01 & 2.933e+02 & 2.494e+01 \\
\hline
22 & 2.940e+03 & 3.900e+02 & 2.927e+03 & 3.841e+02 & \textbf{5.157e+02} & \textbf{2.623e+02} \\
\hline
23 & 3.329e+03 & 3.363e+02 & 3.001e+03 & 3.625e+02 & \textbf{6.381e+02} & \textbf{3.951e+02} \\
\hline
24 & 2.385e+02 & 1.070e+01 & 2.039e+02 & 3.312e+00 & \textbf{2.000e+02} & \textbf{0.000e+00} \\
\hline
25 & 2.723e+02 & 1.896e+01 & 2.467e+02 & 1.617e+01 & \textbf{2.000e+02} & \textbf{7.724e-02} \\
\hline
26 & 2.001e+02 & 2.336e-02 & \textbf{2.000e+02} & \textbf{1.243e-02} & 2.741e+02 & 2.115e+01 \\
\hline
27 & 6.661e+02 & 1.026e+02 & 3.417e+02 & 3.445e+01 & \textbf{3.000e+02} & \textbf{3.834e-02} \\
\hline
28 & \textbf{2.667e+02} & \textbf{7.452e+01} & 3.000e+02 & 3.728e-07 & 3.000e+02 & 5.646e-13 \\
\hline
        AvgRank & \multicolumn{2}{c|}{2.43} & \multicolumn{2}{c|}{1.96} & \multicolumn{2}{c|}{1.46} \\
        \hline
    \end{tabular}
    \label{table:3}
\end{table*}
\begin{table*}[h]
    \centering
    \fontsize{7pt}{7.3pt}\selectfont 
    \caption{Comparing LoTFWA, MGFWA and TFWA on CEC2017, dim = 30}
    \begin{tabular}{|c|c|c|c|c|c|c|}
        \hline
\textbf{idx} & \textbf{LoTFWA.mean} & \textbf{LoTFWA.std} & \textbf{MGFWA.mean} & \textbf{MGFWA.std} & \textbf{TFWA.mean} & \textbf{TFWA.std} \\
\hline
1 & 1.259e+02 & 2.206e+02 & 1.633e+02 & 2.010e+02 & \textbf{0.000e+00} & \textbf{0.000e+00} \\
\hline
2 & 6.286e-03 & 1.142e-03 & 1.627e-02 & 7.486e-03 & \textbf{0.000e+00} & \textbf{1.501e-07} \\
\hline
3 & 5.052e-04 & 6.334e-04 & 2.376e-02 & 2.613e-02 & \textbf{0.000e+00} & \textbf{0.000e+00} \\
\hline
4 & \textbf{5.336e+01} & \textbf{3.557e+01} & 7.995e+01 & 9.053e+00 & 7.896e+01 & 1.538e+01 \\
\hline
5 & 6.215e+01 & 1.123e+01 & 2.716e+01 & 6.273e+00 & \textbf{7.064e+00} & \textbf{2.045e+00} \\
\hline
6 & 5.553e-01 & 8.969e-01 & 1.971e-04 & 5.017e-04 & \textbf{0.000e+00} & \textbf{4.151e-14} \\
\hline
7 & 6.227e+01 & 7.398e+00 & 5.708e+01 & 5.964e+00 & \textbf{3.562e+01} & \textbf{1.497e+00} \\
\hline
8 & 6.364e+01 & 1.050e+01 & 2.863e+01 & 4.590e+00 & \textbf{6.069e+00} & \textbf{2.108e+00} \\
\hline
9 & 3.253e-02 & 1.752e-01 & \textbf{0.000e+00} & \textbf{5.003e-11} & \textbf{0.000e+00} & \textbf{0.000e+00} \\
\hline
10 & 2.538e+03 & 2.507e+02 & 2.225e+03 & 3.450e+02 & \textbf{8.128e+02} & \textbf{2.926e+02} \\
\hline
11 & 9.624e+01 & 2.924e+01 & 1.069e+02 & 3.462e+01 & \textbf{3.020e+01} & \textbf{2.166e+01} \\
\hline
12 & 4.544e+05 & 3.247e+05 & 4.688e+05 & 2.886e+05 & \textbf{1.256e+03} & \textbf{2.913e+02} \\
\hline
13 & 2.344e+04 & 8.940e+03 & 1.632e+04 & 4.265e+03 & \textbf{7.608e+01} & \textbf{1.943e+02} \\
\hline
14 & 8.298e+02 & 5.176e+02 & 7.639e+02 & 4.686e+02 & \textbf{9.465e+01} & \textbf{1.357e+01} \\
\hline
15 & 1.273e+04 & 4.938e+03 & 7.844e+03 & 1.977e+03 & \textbf{8.867e+01} & \textbf{6.362e+01} \\
\hline
16 & 5.248e+02 & 1.671e+02 & 4.956e+02 & 1.449e+02 & \textbf{2.364e+01} & \textbf{6.938e+01} \\
\hline
17 & 1.253e+02 & 4.950e+01 & 1.112e+02 & 3.969e+01 & \textbf{4.597e+01} & \textbf{7.136e+00} \\
\hline
18 & 5.528e+04 & 2.654e+04 & 4.314e+04 & 1.493e+04 & \textbf{9.607e+01} & \textbf{3.237e+01} \\
\hline
19 & 8.406e+04 & 5.924e+04 & 5.563e+04 & 4.543e+04 & \textbf{7.370e+01} & \textbf{1.326e+01} \\
\hline
20 & 2.303e+02 & 5.725e+01 & 2.097e+02 & 6.579e+01 & \textbf{7.597e+01} & \textbf{5.119e+01} \\
\hline
21 & 2.628e+02 & 2.895e+01 & 2.294e+02 & 5.116e+00 & \textbf{2.062e+02} & \textbf{1.721e+00} \\
\hline
22 & \textbf{9.784e+01} & \textbf{1.164e+01} & 1.000e+02 & 3.134e-09 & 1.000e+02 & 0.000e+00 \\
\hline
23 & 4.376e+02 & 2.113e+01 & 3.794e+02 & 6.575e+00 & \textbf{3.415e+02} & \textbf{5.794e+00} \\
\hline
24 & 4.950e+02 & 1.271e+01 & 4.391e+02 & 5.450e+00 & \textbf{4.079e+02} & \textbf{7.086e+00} \\
\hline
25 & 3.868e+02 & 9.873e-01 & 3.868e+02 & 5.140e-01 & \textbf{3.867e+02} & \textbf{1.737e-02} \\
\hline
26 & \textbf{4.548e+02} & \textbf{4.896e+02} & 1.278e+03 & 8.655e+01 & 6.042e+02 & 1.406e+02 \\
\hline
27 & 5.313e+02 & 1.039e+01 & 5.112e+02 & 3.150e+00 & \textbf{5.009e+02} & \textbf{8.759e+00} \\
\hline
28 & 3.400e+02 & 4.578e+01 & 3.537e+02 & 4.388e+01 & \textbf{3.107e+02} & \textbf{3.209e+01} \\
\hline
29 & 7.384e+02 & 8.809e+01 & 7.117e+02 & 6.214e+01 & \textbf{4.291e+02} & \textbf{1.185e+01} \\
\hline
30 & 3.358e+05 & 1.760e+05 & 4.690e+05 & 2.491e+05 & \textbf{2.628e+03} & \textbf{8.722e+02} \\
\hline
\textbf{AvgRank} & \multicolumn{2}{c|}{2.57} & \multicolumn{2}{c|}{2.27} & \multicolumn{2}{c|}{\textbf{1.10}} \\
\hline
\end{tabular}
\label{table:4}
\end{table*}
\subsection{Compare with other optimization algorithms}
For the experiments comparing with other algorithms, we chose the algorithms at their respective TOP level on the two test sets for comparison. Specifically, in the CEC2013 benchmark, we choose the NBIPOPaCMA\cite{c20}, SHADE\cite{c21}, NIPOPaCMA\cite{c20}, MVMO\cite{c22} and SPSO2011\cite{c23} algorithms for comparison; in the CEC2017 benchmark, we choose the CMA series of excellent algorithms, including EBOwithCMAR\cite{c24}, LSHADE-SPACMA\cite{c25} , RB-IPOP-CMA-ES\cite{c26}. For further description of these algorithms, please refer to Table \ref{table:5}.

In Table \ref{table:5} the first 3 algorithms are used for the comparison on CEC2017, which are ranked 1st, 4th and 7th in the public competition; the last 5 algorithms are used for the comparison on CEC2013, top 4 of which are ranked 1st, 4th, 5th and 6th in the public competition.

According to the official release, the top five \textbf{DO NOT} have any common algorithms on the two benchmarks, CEC2013 and CEC2017. However, from Tables \ref{table:6},\ref{table:7}, \ref{table:8}, we find that TFWA achieves very competitive results in average rank compared to their respective top algorithms in different benchmarks. In Table \ref{table:6}, \ref{table:7}, we can find that TFWA get best results on CEC2013 when dim equals 30, better than NBIPOPaCMA, which is the SOTA of the CEC2013 competition; and it performs as well as NBIPOPaCMA when dim equals 50. In Table \ref{table:8}, we compare TFWA with some CMA variants of the competiton, where EBOwithCMAR(a hybrid method) is the SOTA and gets the first place, and TFWA gets the second place, better than another two CMA variants.

It is worth mentioning that TFWA is significantly better than RB-IPOP-CMA-ES (sampling based on Gaussian distribution) in the later multimodal functions of CEC2017, which suggests that T-distributions do play a better role in exploring multimodal problems.

\begin{table*}[htbp]
\centering
\caption{List of Algorithms used to compare and Their Descriptions}
\rowcolors{1}{gray!20}{white}  
\resizebox{\linewidth}{!}{ 
\begin{tabular}{|>{\centering\arraybackslash}m{1cm}|>{\centering\arraybackslash}m{3cm}|>{\centering\arraybackslash}m{12cm}|}  
\hline
\textbf{ID} & \textbf{Algorithm Name} & \textbf{Description} \\
\hline
1 & EBOwithCMAR & Enhances the Butterfly Optimizer with covariance matrix adaptation, improving local search and convergence. \\
\hline
2 & LSHADE\_SPACMA & Combines LSHADE with CMA-ES hybridization, using semi-parameter adaptation to improve performance on complex problems. \\
\hline
3 & RB-IPOP-CMA-ES & A CMA-ES variant with increasing population size to handle bound-constrained optimization effectively. \\
\hline
4 & NBIPOPaCMA & Uses a bi-population restart strategy in CMA-ES to enhance performance on large-scale, multimodal optimization tasks. \\
\hline

5 & SHADE & Adaptive Differential Evolution that adjusts parameters based on past performance to enhance search efficiency. \\
\hline
6 & NIPOPaCMA & Applies adaptive restarts with increasing population sizes in CMA-ES to improve robustness on challenging problems. \\
\hline
7 & MVMO & Mean-Variance Mapping Optimization leverages statistical mappings to guide exploration and exploitation effectively. \\
\hline 
8 & SPSO2011 & An enhanced particle swarm optimization algorithm that adjusts velocity and position update strategies dynamically \\ 
\hline
\end{tabular}
}
\label{table:5}
\end{table*}

\begin{table*}[h]
\centering
\fontsize{6pt}{7.2pt}\selectfont 
\caption{Comparing on CEC2013, dim = 30}
\resizebox{\linewidth}{!}{ 
\begin{tabular}{|c|c|c|c|c|c|c|c|c|c|c|c|c|}
\hline
\textbf{idx} & \textbf{NBIPOPaCMA.mean} & \textbf{NBIPOPaCMA.std} & \textbf{NIPOPaCMA.mean} & \textbf{NIPOPaCMA.std} & \textbf{SHADE.mean} & \textbf{SHADE.std} & \textbf{MVMO.mean} & \textbf{MVMO.std} & \textbf{SPSO2011.mean} & \textbf{SPSO2011.std} & \textbf{TFWA.mean} & \textbf{TFWA.std} \\
\hline
1 & \textbf{0.000e+00} & \textbf{0.000e+00} & \textbf{0.000e+00} & \textbf{0.000e+00} & \textbf{0.000e+00} & \textbf{0.000e+00} & \textbf{0.000e+00} & \textbf{0.000e+00} & \textbf{0.000e+00} & \textbf{0.000e+00} & \textbf{0.000e+00} & \textbf{0.000e+00} \\ 
\hline
2 & \textbf{0.000e+00} & \textbf{0.000e+00} & \textbf{0.000e+00} & \textbf{0.000e+00} & 1.618e+04 & 1.066e+04 & 1.892e-05 & 8.786e-06 & 4.452e+05 & 1.280e+05 & \textbf{0.000e+00} & \textbf{0.000e+00} \\
\hline
3 & 9.310e-05 & 4.850e-04 & \textbf{1.167e-06} & \textbf{6.287e-06} & 2.664e+05 & 6.280e+05 & 1.946e-03 & 6.830e-03 & 4.639e+08 & 6.186e+08 & \textbf{0.000e+00} & \textbf{0.000e+00} \\
\hline
4 & \textbf{0.000e+00} & \textbf{0.000e+00} & \textbf{0.000e+00} & \textbf{0.000e+00} & \textbf{0.000e+00} & \textbf{3.048e+00} & \textbf{3.567e-07} & \textbf{1.626e-06} & 4.319e+04 & 4.000e+03 & \textbf{0.000e+00} & \textbf{0.000e+00} \\
\hline
5 & \textbf{0.000e+00} & \textbf{0.000e+00} & \textbf{0.000e+00} & \textbf{0.000e+00} & \textbf{0.000e+00} & \textbf{2.076e-14} & \textbf{0.000e+00} & \textbf{0.000e+00} & 5.732e-04 & 2.358e-05 & \textbf{0.000e+00} & \textbf{1.249e-07} \\
\hline
6 & \textbf{0.000e+00} & \textbf{0.000e+00} & \textbf{0.000e+00} & \textbf{0.000e+00} & \textbf{0.000e+00} & \textbf{1.534e+00} & \textbf{0.000e+00} & \textbf{0.000e+00} & 5.718e+01 & 2.018e+01 & \textbf{0.000e+00} & \textbf{0.000e+00} \\
\hline
7 & 3.929e+00 & 7.391e+00 & 6.888e+00 & 1.099e+01 & \textbf{0.000e+00} & \textbf{8.745e-01} & 2.847e+01 & 6.530e+00 & 1.015e+02 & 1.520e+01 & 3.270e-04 & 1.761e-03 \\
\hline
8 & 2.097e+01 & 2.333e-02 & 2.096e+01 & 3.255e-02 & \textbf{0.000e+00} & \textbf{8.845e-02} & 2.095e+01 & 2.492e-02 & 2.096e+01 & 2.306e-02 & 2.094e+01 & 6.130e-02 \\
\hline
9 & 4.165e+00 & 1.082e+00 & 3.638e+00 & 7.336e-01 & \textbf{0.000e+00} & \textbf{4.631e+00} & 1.609e+01 & 1.660e+00 & 3.157e+01 & 3.203e+00 & 4.100e+00 & 1.425e+00 \\
\hline
10 & \textbf{0.000e+00} & \textbf{0.000e+00} & \textbf{0.000e+00} & \textbf{0.000e+00} & \textbf{0.000e+00} & \textbf{3.776e-02} & 1.726e-03 & 3.128e-03 & 4.332e-01 & 1.186e-01 & \textbf{0.000e+00} & \textbf{0.000e+00} \\
\hline
11 & 3.914e+00 & 9.927e-01 & 1.703e+00 & 8.228e-01 & \textbf{0.000e+00} & \textbf{1.760e+00} & 8.291e+00 & 1.308e+00 & 1.230e+02 & 1.843e+01 & 5.240e+00 & 1.582e+00 \\
\hline
12 & 3.814e+00 & 9.292e-01 & 1.083e+00 & 8.156e-01 & \textbf{0.000e+00} & \textbf{6.700e+00} & 4.580e+01 & 1.034e+01 & 1.229e+02 & 3.335e+01 & 4.444e+00 & 1.578e+00 \\
\hline
13 & 3.694e+00 & 1.084e+00 & 1.507e+00 & 7.128e-01 & \textbf{0.000e+00} & \textbf{7.995e+00} & 8.386e+01 & 1.422e+01 & 2.192e+02 & 2.498e+01 & 5.413e+00 & 3.690e+00 \\
\hline
14 & 1.034e+03 & 2.897e+02 & \textbf{8.656e+02} & \textbf{1.750e+02} & 1.966e+03 & 2.568e+02 & 1.104e+03 & 3.436e+02 & 4.389e+03 & 4.568e+02 & 1.048e+03 & 4.511e+02 \\
\hline
15 & 9.469e+02 & 2.308e+02 & 8.447e+02 & 2.004e+02 & 5.300e+03 & 2.685e+02 & 3.467e+03 & 1.806e+02 & 4.254e+03 & 3.838e+02 & \textbf{5.424e+02} & \textbf{1.415e+02} \\
\hline
16 & 7.263e-01 & 1.109e+00 & 2.674e+00 & 1.481e-01 & 2.004e+02 & 8.846e-01 & \textbf{4.474e-01} & \textbf{7.719e-02} & 1.553e+00 & 1.899e-01 & 1.596e+00 & 1.125e+00 \\
\hline
17 & 3.544e+01 & 1.791e+00 & 3.525e+01 & 1.480e+00 & 3.592e+02 & 2.675e+00 & 5.385e+01 & 3.834e+00 & 1.288e+02 & 1.542e+01 & \textbf{3.447e+01} & \textbf{1.102e+00} \\
\hline
18 & 8.162e+01 & 5.055e+01 & 6.727e+01 & 5.320e+01 & 5.238e+02 & 7.347e+00 & \textbf{6.262e+01} & \textbf{6.796e+00} & 1.361e+02 & 1.903e+01 & 8.404e+01 & 6.489e+01 \\
\hline
19 & 2.446e+00 & 1.928e-01 & 2.699e+00 & 2.308e-01 & 5.042e+02 & 3.300e-01 & \textbf{2.235e+00} & \textbf{3.198e-01} & 1.185e+01 & 4.283e+00 & 2.637e+00 & 3.683e-01 \\
\hline
20 & 1.332e+01 & 1.699e-01 & 1.420e+01 & 6.613e-01 & 6.123e+02 & 3.206e-01 & \textbf{1.084e+01} & \textbf{3.242e-01} & 1.425e+01 & 3.264e-01 & 1.315e+01 & 2.331e+00 \\
\hline
21 & \textbf{2.000e+02} & \textbf{0.000e+00} & 2.700e+02 & 4.583e+01 & 9.977e+02 & 6.101e+01 & 2.314e+02 & 5.758e+01 & 3.383e+02 & 6.348e+01 & 2.933e+02 & 2.494e+01 \\
\hline
22 & 1.116e+03 & 3.809e+02 & 7.434e+02 & 3.422e+02 & 3.068e+03 & 1.812e+02 & 1.156e+03 & 3.867e+02 & 4.820e+03 & 4.591e+02 & \textbf{5.157e+02} & \textbf{2.623e+02} \\
\hline
23 & 8.446e+02 & 2.298e+02 & 8.560e+02 & 2.582e+02 & 6.336e+03 & 2.867e+02 & 3.792e+03 & 3.268e+02 & 5.374e+03 & 5.845e+02 & \textbf{6.381e+02} & \textbf{3.951e+02} \\
\hline
24 & \textbf{1.778e+02} & \textbf{2.887e+01} & 3.004e+02 & 1.955e+00 & 1.277e+03 & 3.542e+00 & 2.196e+02 & 4.733e+00 & 2.748e+02 & 8.882e+00 & 2.000e+02 & 0.000e+00 \\
\hline
25 & 2.250e+02 & 2.500e+00 & 2.999e+02 & 1.312e+00 & 1.393e+03 & 5.241e+00 & 2.641e+02 & 4.476e+00 & 3.063e+02 & 6.261e+00 & \textbf{2.000e+02} & \textbf{7.724e-02} \\
\hline
26 & \textbf{1.730e+02} & \textbf{3.098e+01} & 2.886e+02 & 4.116e+01 & 1.409e+03 & 3.012e+01 & 2.000e+02 & 2.026e-04 & 3.463e+02 & 5.010e+01 & 2.741e+02 & 2.115e+01 \\
\hline
27 & 5.209e+02 & 4.880e+01 & 1.177e+03 & 2.580e+02 & 2.383e+03 & 3.370e+01 & 6.344e+02 & 8.365e+01 & 1.081e+03 & 5.879e+01 & \textbf{3.000e+02} & \textbf{3.834e-02} \\
\hline
28 & \textbf{3.000e+02} & \textbf{0.000e+00} & \textbf{3.000e+02} & \textbf{0.000e+00} & 1.700e+03 & 1.761e-13 & 3.358e+02 & 1.930e+02 & 4.922e+02 & 5.952e+02 & \textbf{3.000e+02} & \textbf{5.646e-13} \\
\hline
\textbf{AvgRank} &  \multicolumn{2}{c|}{2.21}  &  \multicolumn{2}{c|}{2.54}  &  \multicolumn{2}{c|}{3.89}  &  \multicolumn{2}{c|}{3.07} &  \multicolumn{2}{c|}{5.14}  &  \multicolumn{2}{c|}{2.07} \\
\hline
\end{tabular}
}
\label{table:6}
\end{table*}

\begin{table*}[h]
\centering
\caption{Comparing on CEC2013, dim = 50}
\fontsize{6pt}{7.2pt}\selectfont 
\resizebox{\linewidth}{!}{ 
\begin{tabular}{|c|c|c|c|c|c|c|c|c|c|c|c|c|}
\hline
\textbf{idx} & \textbf{NBIPOPaCMA.mean} & \textbf{NBIPOPaCMA.std} & \textbf{NIPOPaCMA.mean} & \textbf{NIPOPaCMA.std} & \textbf{SHADE.mean} & \textbf{SHADE.std} & \textbf{MVMO.mean} & \textbf{MVMO.std} & \textbf{SPSO2011.mean} & \textbf{SPSO2011.std} & \textbf{TFWA.mean} & \textbf{TFWA.std} \\
\hline
1 & \textbf{0.000e+00} & \textbf{0.000e+00} & \textbf{0.000e+00} & \textbf{0.000e+00} & \textbf{0.000e+00} & \textbf{0.000e+00} & \textbf{0.000e+00} & \textbf{0.000e+00} & \textbf{0.000e+00} & \textbf{0.000e+00} & \textbf{0.000e+00} & \textbf{0.000e+00} \\
\hline
2 & \textbf{0.000e+00} & \textbf{0.000e+00} & \textbf{0.000e+00} & \textbf{0.000e+00} & 3.421e+04 & 7.379e+03 & 1.209e-03 & 7.102e-04 & 8.165e+05 & 1.251e+05 & 1.201e-04 & 6.467e-04 \\
\hline
3 & 3.088e+01 & 1.554e+02 & 3.191e+01 & 1.493e+02 & 1.474e+06 & 2.352e+06 & 6.164e-04 & 2.053e-03 & 1.172e+09 & 1.028e+09 & \textbf{0.000e+00} & \textbf{0.000e+00} \\
\hline
4 & \textbf{0.000e+00} & \textbf{0.000e+00} & \textbf{0.000e+00} & \textbf{0.000e+00} & 2.389e-03 & 1.344e-03 & \textbf{0.000e+00} & \textbf{9.533e-07} & 5.690e+04 & 6.773e+03 & 2.792e-04 & 1.504e-03 \\
\hline
5 & \textbf{0.000e+00} & \textbf{0.000e+00} & \textbf{0.000e+00} & \textbf{0.000e+00} & \textbf{0.000e+00} & \textbf{0.000e+00} & \textbf{0.000e+00} & \textbf{7.885e-09} & 1.022e-03 & 2.718e-05 & \textbf{0.000e+00} & \textbf{3.766e-07} \\
\hline
6 & \textbf{0.000e+00} & \textbf{0.000e+00} & \textbf{0.000e+00} & \textbf{0.000e+00} & 4.364e+01 & 1.020e+00 & 4.345e+01 & 1.421e-14 & 6.360e+01 & 2.626e+01 & 4.345e+01 & 2.355e-10 \\
\hline
7 & 8.198e+00 & 5.378e+00 & 1.948e+01 & 4.716e+01 & 2.952e+01 & 6.655e+00 & 4.873e+01 & 4.779e+00 & 9.829e+01 & 1.250e+01 & \textbf{1.236e-03} & \textbf{6.640e-03} \\
\hline
8 & 2.115e+01 & 1.726e-02 & 2.114e+01 & 2.035e-02 & \textbf{2.102e+01} & \textbf{1.235e-01} & 2.113e+01 & 2.448e-02 & 2.114e+01 & 1.857e-02 & 2.114e+01 & 3.080e-02 \\
\hline
9 & 8.638e+00 & 1.665e+00 & 8.017e+00 & 1.215e+00 & 5.678e+01 & 9.722e-01 & 3.570e+01 & 2.766e+00 & 6.038e+01 & 4.853e+00 & \textbf{7.939e+00} & \textbf{3.742e+00} \\
\hline
10 & \textbf{0.000e+00} & \textbf{0.000e+00} & \textbf{0.000e+00} & \textbf{0.000e+00} & 9.689e-02 & 2.845e-02 & 2.465e-04 & 1.328e-03 & 5.396e-01 & 2.193e-01 & \textbf{0.000e+00} & \textbf{0.000e+00} \\
\hline
11 & 7.636e+00 & 1.715e+00 & 2.787e+00 & 1.039e+00 & \textbf{0.000e+00} & \textbf{0.000e+00} & 4.216e+01 & 8.145e+00 & 2.619e+02 & 2.967e+01 & 6.865e+00 & 2.509e+00 \\
\hline
12 & 7.031e+00 & 1.452e+00 & \textbf{2.124e+00} & \textbf{1.111e+00} & 6.578e+01 & 8.134e+00 & 1.031e+02 & 1.708e+01 & 2.711e+02 & 3.740e+01 & 7.131e+00 & 2.465e+00 \\
\hline
13 & 1.044e+01 & 5.388e+00 & \textbf{2.029e+00} & \textbf{9.105e-01} & 1.584e+02 & 1.352e+01 & 2.017e+02 & 1.387e+01 & 4.708e+02 & 4.175e+01 & 7.025e+00 & 4.318e+00 \\
\hline
14 & 1.715e+03 & 4.705e+02 & 1.501e+03 & 3.834e+02 & \textbf{4.664e-02} & \textbf{1.476e-02} & 2.866e+03 & 5.146e+02 & 7.837e+03 & 5.492e+02 & 1.367e+03 & 5.031e+02 \\
\hline
15 & 1.845e+03 & 4.087e+02 & \textbf{1.671e+03} & \textbf{3.423e+02} & 7.099e+03 & 2.648e+02 & 6.525e+03 & 2.638e+02 & 8.751e+03 & 7.259e+02 & 1.754e+03 & 1.897e+03 \\
\hline
16 & 1.470e+00 & 1.615e+00 & 3.558e+00 & 1.616e-01 & 1.418e+00 & 1.277e-01 & \textbf{1.258e+00} & \textbf{1.251e-01} & 2.217e+00 & 3.260e-01 & 3.295e+00 & 5.776e-01 \\
\hline
17 & 5.858e+01 & 2.947e+00 & 5.894e+01 & 2.318e+00 & \textbf{5.079e+01} & \textbf{7.105e-15} & 1.112e+02 & 9.932e+00 & 3.498e+02 & 4.969e+01 & 9.134e+01 & 8.760e+01 \\
\hline
18 & 1.749e+02 & 9.963e+01 & 2.796e+02 & 1.002e+02 & 1.448e+02 & 9.047e+00 & \textbf{1.158e+02} & \textbf{9.826e+00} & 3.379e+02 & 4.700e+01 & 2.721e+02 & 1.193e+02 \\
\hline
19 & 4.868e+00 & 3.287e-01 & 4.792e+00 & 2.044e-01 & \textbf{2.830e+00} & \textbf{1.679e-01} & 4.451e+00 & 3.454e-01 & 4.485e+01 & 8.532e+00 & 5.492e+00 & 3.104e+00 \\
\hline
20 & 2.322e+01 & 5.089e-01 & 2.394e+01 & 7.531e-01 & 1.981e+01 & 4.030e-01 & 1.995e+01 & 4.053e-01 & 2.328e+01 & 5.574e-01 & \textbf{1.943e+01} & \textbf{1.576e+00} \\
\hline
21 & \textbf{2.000e+02} & \textbf{0.000e+00} & 4.502e+02 & 3.630e+02 & 1.084e+03 & 9.714e+01 & 2.905e+02 & 2.367e+02 & 1.056e+03 & 1.209e+02 & 7.217e+02 & 3.607e+02 \\
\hline
22 & 1.789e+03 & 5.200e+02 & 1.308e+03 & 3.678e+02 & \textbf{1.500e+01} & \textbf{8.749e+00} & 3.271e+03 & 6.880e+02 & 1.003e+04 & 1.006e+03 & 1.877e+03 & 2.381e+03 \\
\hline
23 & 2.159e+03 & 7.390e+02 & 1.589e+03 & 6.118e+02 & 8.068e+03 & 4.002e+02 & 7.140e+03 & 3.492e+02 & 1.125e+04 & 8.885e+02 & \textbf{1.480e+03} & \textbf{1.295e+03} \\
\hline
24 & 2.530e+02 & 6.311e+00 & 3.847e+02 & 2.406e+00 & 2.398e+02 & 8.456e+00 & 2.477e+02 & 6.000e+00 & 3.540e+02 & 9.832e+00 & \textbf{2.000e+02} & \textbf{1.097e-05} \\
\hline
25 & \textbf{2.509e+02} & \textbf{2.732e+00} & 3.847e+02 & 1.803e+00 & 3.618e+02 & 9.857e+00 & 3.317e+02 & 7.741e+00 & 4.153e+02 & 1.577e+01 & 2.804e+02 & 6.781e+00 \\
\hline
26 & \textbf{2.008e+02} & \textbf{4.191e+00} & 3.500e+02 & 8.241e+01 & 2.978e+02 & 8.331e+01 & 2.059e+02 & 3.198e+01 & 4.392e+02 & 1.070e+01 & 2.997e+02 & 1.549e+00 \\
\hline
27 & 8.083e+02 & 3.369e+01 & 2.139e+03 & 1.850e+01 & 1.100e+03 & 2.962e+02 & 1.111e+03 & 7.992e+01 & 1.790e+03 & 1.117e+02 & \textbf{3.000e+02} & \textbf{1.497e-13} \\
\hline
28 & \textbf{4.000e+02} & \textbf{0.000e+00} & 6.917e+02 & 8.751e+02 & 4.983e+02 & 5.295e+02 & 5.002e+02 & 5.394e+02 & 1.272e+03 & 1.588e+03 & \textbf{4.000e+02} & \textbf{2.117e-13} \\
\hline
\textbf{AvgRank} & \multicolumn{2}{c|}{2.43}  & \multicolumn{2}{c|}{3.00}  & \multicolumn{2}{c|}{3.11} & \multicolumn{2}{c|}{3.21}  & \multicolumn{2}{c|}{5.57}  & \multicolumn{2}{c|}{2.43} \\
\hline
\end{tabular}
}
\label{table:7}
\end{table*}

\begin{table*}[h]
\centering
\caption{Comparing on CEC2017, dim = 30}
\resizebox{\linewidth}{!}{ 
\begin{tabular}{|c|c|c|c|c|c|c|c|c|}
\hline
\textbf{idx} & \textbf{EBOwithCMAR.mean} & \textbf{EBOwithCMAR.std} & \textbf{LSHADE\_SPACMA.mean} & \textbf{LSHADE\_SPACMA.std} & \textbf{RB-IPOP-CMA-ES.mean} & \textbf{RB-IPOP-CMA-ES.std} & \textbf{TFWA.mean} & \textbf{TFWA.std} \\
\hline
1 & \textbf{0.000e+00} & \textbf{0.000e+00} & \textbf{0.000e+00} & \textbf{0.000e+00} & \textbf{0.000e+00} & \textbf{1.963e-08} & \textbf{0.000e+00} & \textbf{0.000e+00} \\
\hline
2 & \textbf{0.000e+00} & \textbf{0.000e+00} & 3.333e-02 & 1.795e-01 & \textbf{0.000e+00} & \textbf{0.000e+00} & \textbf{0.000e+00} & \textbf{1.501e-07} \\
\hline
3 & \textbf{0.000e+00} & \textbf{0.000e+00} & \textbf{0.000e+00} & \textbf{0.000e+00} & \textbf{0.000e+00} & \textbf{0.000e+00} & \textbf{0.000e+00} & \textbf{0.000e+00} \\
\hline
4 & 5.875e+01 & 9.973e-01 & \textbf{5.856e+01} & \textbf{0.000e+00} & 6.078e+01 & 2.722e+00 & 7.896e+01 & 1.538e+01 \\
\hline
5 & 3.781e+00 & 1.485e+00 & 4.863e+00 & 2.507e+00 & \textbf{2.469e+00} & \textbf{1.154e+00} & 7.064e+00 & 2.045e+00 \\
\hline
6 & \textbf{0.000e+00} & \textbf{0.000e+00} & \textbf{0.000e+00} & \textbf{0.000e+00} & \textbf{0.000e+00} & \textbf{1.491e-08} & \textbf{0.000e+00} & \textbf{4.151e-14} \\
\hline
7 & \textbf{3.397e+01} & \textbf{6.425e-01} & 3.433e+01 & 6.146e-01 & 3.514e+01 & 1.028e+00 & 3.562e+01 & 1.497e+00 \\
\hline
8 & 2.885e+00 & 9.732e-01 & 4.313e+00 & 8.658e-01 & \textbf{2.687e+00} & \textbf{1.501e+00} & 6.069e+00 & 2.108e+00 \\
\hline
9 & \textbf{0.000e+00} & \textbf{0.000e+00} & \textbf{0.000e+00} & \textbf{0.000e+00} & \textbf{0.000e+00} & \textbf{0.000e+00} & \textbf{0.000e+00} & \textbf{0.000e+00} \\
\hline
10 & 1.551e+03 & 1.389e+02 & 1.594e+03 & 1.671e+02 & 1.805e+03 & 4.424e+02 & \textbf{8.128e+02} & \textbf{2.926e+02} \\
\hline
11 & \textbf{6.643e+00} & \textbf{1.080e+01} & 2.880e+01 & 2.706e+01 & 6.710e+01 & 4.620e+01 & 3.020e+01 & 2.166e+01 \\
\hline
12 & \textbf{6.414e+02} & \textbf{1.755e+02} & 8.071e+02 & 2.495e+02 & 1.273e+03 & 1.826e+02 & 1.256e+03 & 2.913e+02 \\
\hline
13 & 1.891e+01 & 2.471e+00 & \textbf{1.714e+01} & \textbf{1.359e+00} & 1.918e+02 & 5.033e+02 & 7.608e+01 & 1.943e+02 \\
\hline
14 & \textbf{2.328e+01} & \textbf{1.185e+00} & 2.439e+01 & 1.444e+00 & 1.328e+02 & 2.814e+01 & 9.465e+01 & 1.357e+01 \\
\hline
15 & \textbf{4.999e+00} & \textbf{1.794e+00} & 5.784e+00 & 2.028e+00 & 3.466e+02 & 1.179e+02 & 8.867e+01 & 6.362e+01 \\
\hline
16 & 6.300e+01 & 6.617e+01 & 3.332e+01 & 1.854e+01 & 6.693e+02 & 1.774e+02 & \textbf{2.364e+01} & \textbf{6.938e+01} \\
\hline
17 & \textbf{3.447e+01} & \textbf{3.206e+00} & 3.510e+01 & 6.057e+00 & 1.898e+02 & 8.300e+01 & 4.597e+01 & 7.136e+00 \\
\hline
18 & \textbf{2.280e+01} & \textbf{8.780e-01} & 2.445e+01 & 1.720e+00 & 2.196e+02 & 1.122e+02 & 9.607e+01 & 3.237e+01 \\
\hline
19 & \textbf{9.538e+00} & \textbf{1.531e+00} & 1.177e+01 & 1.227e+00 & 1.550e+02 & 5.215e+01 & 7.370e+01 & 1.326e+01 \\
\hline
20 & \textbf{4.027e+01} & \textbf{6.111e+00} & 1.162e+02 & 5.037e+01 & 3.749e+02 & 7.716e+01 & 7.597e+01 & 5.119e+01 \\
\hline
21 & \textbf{2.039e+02} & \textbf{1.128e+00} & 2.095e+02 & 2.860e+00 & 2.150e+02 & 3.746e+00 & 2.062e+02 & 1.721e+00 \\
\hline
22 & \textbf{1.000e+02} & \textbf{0.000e+00} & \textbf{1.000e+02} & \textbf{0.000e+00} & 1.072e+03 & 7.618e+02 & \textbf{1.000e+02} & \textbf{0.000e+00} \\
\hline
23 & 3.534e+02 & 2.768e+00 & 3.577e+02 & 2.050e+00 & 3.531e+02 & 5.700e+00 & \textbf{3.415e+02} & \textbf{5.794e+00} \\
\hline
24 & 4.254e+02 & 1.019e+00 & 4.310e+02 & 1.508e+00 & 4.214e+02 & 1.462e+00 & \textbf{4.079e+02} & \textbf{7.086e+00} \\
\hline
25 & 3.867e+02 & 1.569e-02 & \textbf{3.867e+02} & \textbf{8.873e-03} & 3.867e+02 & 1.413e-02 & 3.867e+02 & 1.737e-02 \\
\hline
26 & 7.356e+02 & 2.425e+02 & 9.857e+02 & 3.778e+01 & \textbf{5.229e+02} & \textbf{1.931e+02} & 6.042e+02 & 1.406e+02 \\
\hline
27 & 5.051e+02 & 2.019e+00 & 5.088e+02 & 4.156e+00 & 5.190e+02 & 7.920e+00 & \textbf{5.009e+02} & \textbf{8.759e+00} \\
\hline
28 & 3.141e+02 & 3.606e+01 & 3.190e+02 & 4.248e+01 & 3.152e+02 & 3.690e+01 & \textbf{3.107e+02} & \textbf{3.209e+01} \\
\hline
29 & 4.404e+02 & 7.110e+00 & 4.527e+02 & 1.017e+01 & 5.377e+02 & 1.130e+02 & \textbf{4.291e+02} & \textbf{1.185e+01} \\
\hline
30 & \textbf{2.011e+03} & \textbf{3.976e+01} & 2.040e+03 & 5.947e+01 & 3.385e+03 & 2.359e+03 & 2.628e+03 & 8.722e+02 \\
\hline
\textbf{AvgRank} &  \multicolumn{2}{c|}{1.63}  &  \multicolumn{2}{c|}{2.30}  &  \multicolumn{2}{c|}{2.90}  &  \multicolumn{2}{c|}{2.17}  \\
\hline
\end{tabular}
}
\label{table:8}
\end{table*}

\subsection{Analysis of Degree of Freedom Adjustment and Restart Mechanisms}

Freedom adjustment and restart mechanisms play important roles in TFWA. We show four typical variations of degrees of freedom and fitness values in Fig.\ref{fig:2}. Each subfigure in the Fig.\ref{fig:2} is divided into left and right parts, where the left side represents the change in the degrees of freedom of the N fireworks individuals, and the right side represents the change in the difference between the fitness value and the optimal value of the N fireworks individuals.

The Fig.\ref{fig:sub1} shows the normal evolution of two individual fireworks on a sphere function, where one firework's student's t-distribution's degrees of freedom increases fast, the other increases slowly, and both optimize to the global optimum; (note that the green line is much steeper, which is in line with the discussion on the local exploitation ability of the student's t-distribution). The same curve also appears in the rotated high-conditioned elliptic Function.  

The Fig.\ref{fig:sub2} shows that on the rotated Rosenbrock’s Function, one firework has already optimized to the optimal solution case (the green line), the other firework is restarted several times (the red line), but in the end it is not optimized to the optimal solution. 

The Fig.\ref{fig:sub3} shows that on the Rotated Ackley’s Function, both fireworks have restarted several times, with the green line exhibiting a relatively higher degree of freedom. In the right graph, we can see that both fireworks experienced fluctuations during the optimization process, while continuously updating the optimal value, which demonstrates the importance of restarts and adjustments in the degree of freedom.

The Fig.\ref{fig:sub4} shows that on the Lunacek Bi$\_$Rastrigin Function only the firework-1 restarted once, while both fireworks experienced multiple fluctuations. Although the degree of freedom for the firework-2 is much higher than that for the firework-1, implying that firework-2 has a stronger exploitation capability and firework-1 has a stronger exploration capability, the final results show that firework-1 achieved a better solution than that firework-2 after the restart. This indicates the necessity of the restart mechanism.

In a word, not only the T-distribution is necessary, but also the adjustment mechanism of the degrees of freedom and the restart mechanism are also critical. For more information, you can refer to supplementary material.

\section{Conclusion}
 This work proposes the student's t-distribution based fireworks algorithm, TFWA, in the natural gradient perspective, which not only has a solid theoretical foundation, but also proves to have the ability to match or even surpass the SOTA algorithm on the major test benchmarks in experiments.

Our contributions can be listed as follows.
\begin{itemize}
    \item The T-distribution based sampling is analyzed under the natural gradient framework, and a new underlying explosion mechanism suitable for the fireworks algorithm is derived
    \item A degree-of-freedom adjustment mechanism is proposed, which well balances the algorithm's ability to mine and explore; it well solves the disadvantage of previous fireworks algorithms on single-mode problems
    \item The restart mechanism in the previous variant LoTFWA is used to further improve the optimization performance of the algorithm
\end{itemize}

In this work, we propose a brand new FWA variant TFWA equipped with student's t-distribution based explosion. On the one hand, TFWA has a solid theoretical foundation, and on the other hand, ample experiments have shown that TFWA has surpassed the current variants of FWA to become the new best variant. Moreover, TFWA has also achieved results on major mainstream function test benchmarks that are on par with, or even exceed, the current state-of-the-art optimization algorithms. What's more noteworthy is that TFWA has outperformed the majority of CMAES variants (and even hybrid algorithms), which also demonstrates the superiority and potential of sampling based on the student's t-distribution.

 
%

\vspace{12pt}

\newpage
\onecolumn
\appendix

\section{Derivation for Equation 17-19}
We will give the derivation of the equation 17-19
\begin{equation}
    p(x)  = \frac{\Gamma(\frac{v}{2} + \frac{d}{2})}{\Gamma(\frac{v}{2})} * \frac{{|\Sigma|}^{-\frac{1}{2}}}{(v * \pi)^{\frac{d}{2}}} * [1 + \frac{1}{v}*(x-\mu)^{T}\Sigma^{-1}(x-\mu)]^{-\frac{d+v}{2}}
\end{equation}

\begin{equation}
    ln[p(x)]  = ln[\frac{\Gamma(\frac{v}{2} + \frac{d}{2})}{\Gamma(\frac{v}{2})}] - \frac{1}{2}*ln(|\Sigma|) - \frac{d+v}{2}ln(1 + \frac{1}{v}*(x-\mu)^{T}\Sigma^{-1}(x-\mu))
\end{equation}
Notice:
\begin{equation}
    \frac{\partial{ln(|\Sigma|)}}{\partial{\theta}} = \frac{1}{|\Sigma|}*\sum_{i,j}{M_{i,j}\frac{\partial{\Sigma_{i,j}}}{\theta}} = \frac{1}{|\Sigma|}*tr(\frac{\partial{\Sigma}}{\partial{\theta}}*M^{T})=tr(\frac{\partial{\Sigma}}{\partial{\theta}}*\Sigma^{-1})
\end{equation}
We have:
\begin{equation}
\begin{split}
    \frac{\partial{ln[p(x)]}}{\partial{\theta}}= &-\frac{1}{2}tr(\frac{\partial{\Sigma}}{\partial{\theta}}*\Sigma^{-1})-\frac{d+v}{2}*\frac{1}{v+(x-u)^{T}\Sigma^{-1}(x-u)}* \\
    & [-2(\frac{\partial{u}}{\partial{\theta}})^{T}\Sigma^{-1}(x-u)+(x-u)^{T}\frac{\partial{\Sigma^{-1}}}{\partial{\theta}}(x-u)]
\end{split}
\end{equation}
Now we want to calculate Fisher Information Matrix $F$, where $F_{i,j}=E_{x}[\nabla ln[p(x|\theta_{i})]^{T}\nabla ln[p(x|\theta_{j})]^{T}]$, after which we can get the true gradient from $g = F^{-1}(\theta)\nabla ln[p(x|\theta)]$
With equation $(2)$, we can calculate the following thing:
\begin{equation}
\begin{split}
    F_{ij}[\theta]= & E_{x}[\frac{1}{4}tr(\frac{\partial{\Sigma}}{\partial{\theta_{i}}}*\Sigma^{-1})tr(\frac{\partial{\Sigma}}{\partial{\theta_{j}}}*\Sigma^{-1}) +
    \\ &(\frac{d+v}{v+s})^{2}(\frac{\partial{\mu}}{\partial{\theta_{i}}})^{T}\Sigma^{-1}(x-u)(\frac{\partial{\mu}}{\partial{\theta_{j}}})^{T}\Sigma^{-1}(x-u)+
    \\ & \frac{1}{4}(\frac{d+v}{v+s})^{2}(x-u)^{T}\frac{\partial{\Sigma^{-1}}}{\partial{\theta_{i}}}(x-u)(x-u)^{T}\frac{\partial{\Sigma^{-1}}}{\partial{\theta_{j}}}(x-u)+
    \\ & \frac{d+v}{2(v+s)}tr(\frac{\partial{\Sigma}}{\partial{\theta_{i}}}*\Sigma^{-1})(x-u)^{T}\frac{\partial{\Sigma^{-1}}}{\partial{\theta_{j}}}(x-u)]
\end{split}
\end{equation}
Where $s = (x-u)^{T}\Sigma^{-1}(x-u)$, and we will split the $F_{i,j}(\theta)$ into 4 parts to calculate as shown in Equation$(5)$.
\subsection{Part1}
We choose to keep the form as it is in Equation$(5)$.
\begin{equation}
    Part_{1} = \frac{1}{4}tr(\frac{\partial{\Sigma}}{\partial{\theta_{i}}}*\Sigma^{-1})tr(\frac{\partial{\Sigma}}{\partial{\theta_{j}}}*\Sigma^{-1})
\end{equation}
\subsection{Part2}
\begin{equation}
    \begin{split}
        Part_{2}= & E_{x}[(\frac{d+v}{v+s})^{2}(\frac{\partial{\mu}}{\partial{\theta_{i}}})^{T}\Sigma^{-1}(x-u)(\frac{\partial{\mu}}{\partial{\theta_{j}}})^{T}\Sigma^{-1}(x-u)]\\
        = & (\frac{d+v}{v})^{2}(\frac{\partial{\mu}}{\partial{\theta_{i}}})^T\Sigma^{-1}E_{x}[\frac{(x-u)(x-u)^{T}}{(1+\frac{s}{v})^{2}}]\Sigma^{-1}\frac{\partial{\mu}}{\partial{\theta_{j}}}
    \end{split}
\end{equation}
\begin{equation}
    \begin{split}
        E_{x}[\frac{(x-u)(x-u)^{T}}{(1+\frac{s}{v})^{2}}]
         = &\int_{x}\frac{\Gamma(\frac{v}{2} + \frac{d}{2})}{\Gamma(\frac{v}{2})} * \frac{{|\Sigma|}^{-\frac{1}{2}}}{(v * \pi)^{\frac{d}{2}}} * [1 + \frac{s}{v}]^{-\frac{d+v}{2}}*\frac{(x-u)(x-u)^{T}}{(1+\frac{s}{v})^{2}}\mathrm{d}x \\
        = &\int_{x}\frac{\Gamma(\frac{v}{2} + \frac{d}{2})}{\Gamma(\frac{v}{2})} * \frac{{|\Sigma|}^{-\frac{1}{2}}}{(v * \pi)^{\frac{d}{2}}} * [1 + \frac{s}{v}]^{-\frac{d+v+4}{2}}*(x-u)(x-u)^{T}\mathrm{d}x \\
        = &\frac{v(v+2)}{(d+v)(d+v+2)} * \int_{x} P_{t}(x|\mu, v+4,\Sigma^{'})(x-u)(x-u)^{T}\mathrm{d}x \\
        = & \frac{v^{2}}{(d+v)(d+v+2)}\Sigma \\
    \end{split}
\end{equation}
So we can get the following form:
\begin{equation}
    Part_{2}=\frac{d+v}{d+v+2}(\frac{\partial{\mu}}{\partial{\theta_{i}}})^{T}\Sigma^{-1}(\frac{\partial{\mu}}{\partial{\theta_{j}}})
\end{equation}
In CMA-ES, this equation(9) is shown below:
\begin{equation}
    Part_{2,CMAES}=(\frac{\partial{\mu}}{\partial{\theta_{i}}})^{T}\Sigma^{-1}(\frac{\partial{\mu}}{\partial{\theta_{j}}})
\end{equation}
\subsection{Part3}
\begin{equation}
    \begin{split}
    Part_{3}= &\frac{(d+v)^{2}}{v^{2}}\frac{\Gamma(\frac{v}{2} + \frac{d}{2})}{\Gamma(\frac{v}{2})} * \frac{{|\Sigma|}^{-\frac{1}{2}}}{(v * \pi)^{\frac{d}{2}}} \int_{y}y^{T}Ayy^{T}By[1+\frac{y^{T}\Sigma^{-1}y}{v}]^{-\frac{d+v+4}{2}}\mathrm{d}y\\
    = & \frac{(d+v)(v+2)}{4v(d+v+2)} * \int_{y}y^{T}Ayy^{T}ByP_{t}(0,v+4,\Sigma^{'})\mathrm{d}y
    \end{split}
\end{equation}
where $y^{T}Ay=(x-u)^{T}\frac{\partial{\Sigma^{-1}}}{\partial{\theta_{i}}}(x-u)$ and $y^{T}By=(x-u)^{T}\frac{\partial{\Sigma^{-1}}}{\partial{\theta_{j}}}(x-u)$. 
Further, we can let $(x-u)^{T}\Sigma^{-\frac{1}{2}}(x-u) = z^{T}A^{'}z$, where $A^{'}=\frac{v}{v+4}(-\Sigma^{-\frac{1}{2}}\frac{\partial{\Sigma}}{\partial{\theta_{i}}}\Sigma^{-\frac{1}{2}})$
\begin{equation}
    \begin{split}
        \int_{y}y^{T}Ayy^{T}ByP_{t}(0,v+4,\Sigma^{'})\mathrm{d}y
        = \int_{z}z^{T}A^{'}zz^{T}B^{'}zP_{t}(0,v+4,I)\mathrm{d}z 
    \end{split}
\end{equation}
\begin{equation}
    \begin{split}
        z^{T}A^{'}zz^{T}B^{'}z = \sum_{i,j,m,n}z_{i}a_{ij}z_{j}z_{m}b_{mn}z_{n}
    \end{split}
\end{equation}
According to the conclusion from \cite{c33,c34,c35,c36}, we have:
\begin{equation}
    E[\Pi_{i=1}^{p}Z_{i}^{k_{i}}]=V^{\frac{v}{2}}\frac{\Gamma(\frac{v-k}{2})}{\Gamma(\frac{v}{2})2^{k}}*\Pi_{i=1}^{p}\frac{k_{i}!}{\frac{k_{i}}{2}!}
\end{equation}
Where $k = \sum_{i}k_{i}$.
So we can get that:
\begin{equation}
\begin{split}
      E_{z}[\sum_{k}a_{kk}b_{kk}z_{k}^{4}]
    =&\sum_{k}a_{kk}b_{kk}\frac{3(v+4)^2}{v(v+2)}
\end{split}
\end{equation}
\begin{equation}
\begin{split}
      E_{z}[\sum_{i \neq j}(a_{ii}b_{jj}+a_{ij}b_{ij}+a_{ij}b_{ji})z_{i}^{2}z_{j}^{2}]
    = &(tr(A')tr(B')+tr(AB^{T})+tr(AB) - 3\sum_{i}(a_{ii}b_{ii}))\frac{(v+4)^2}{v(v+2)}
\end{split}
\end{equation}
Conclude from equation(12)~(16),we have:
\begin{equation}
\begin{split}
     Part_{3}=\frac{d+v}{4(d+v+2)}[tr(\frac{\partial \Sigma}{\partial \theta_{i}}\Sigma^{-1})tr(\frac{\partial \Sigma}{\partial \theta_{j}}\Sigma^{-1}) + 2tr(\frac{\partial \Sigma}{\partial \theta_{i}}\Sigma^{-1}\frac{\partial \Sigma}{\partial \theta_{j}}\Sigma^{-1})]
\end{split}
\end{equation}
\subsection{Part4}
\begin{equation}
\begin{split}
     E_{x}[\frac{d+v}{2(v+s)}tr(\frac{\partial{\Sigma}}{\partial{\theta_{i}}}*\Sigma^{-1})(x-u)^{T}\frac{\partial{\Sigma^{-1}}}{\partial{\theta_{j}}}(x-u)] 
    =  \frac{d+v}{2v}tr(\frac{\partial{\Sigma}}{\partial{\theta_{i}}}\Sigma^{-1})E_x[tr(\frac{(x-u)(x-u)^T}{1+\frac{s}{v}})]\frac{\partial{\Sigma^{-1}}}{\partial{\theta_{j}}}
\end{split}
\end{equation}
\begin{equation}
\begin{split}
    E_x[tr(\frac{(x-u)(x-u)^T}{1+\frac{s}{v}})]
    =&\int_{x} \frac{\Gamma(\frac{v}{2} + \frac{d}{2})}{\Gamma(\frac{v}{2})} * \frac{{|\Sigma|}^{-\frac{1}{2}}}{(v * \pi)^{\frac{d}{2}}}(1+\frac{(x-u)^{T}\Sigma^{-1}(x-u)}{v})^{-\frac{d+v+2}{2}}(x-u)(x-u)^{T}\mathrm{d}x \\  
    =& \frac{v}{d+v}\Sigma
\end{split}
\end{equation}
\begin{equation}
    Part_{4}=-\frac{1}{2}tr(\frac{\partial \Sigma}{\partial \theta_{i}}\Sigma^{-1})tr(\frac{\partial \Sigma}{\partial \theta_{j}}\Sigma^{-1})
\end{equation}
\subsection{Combine}
Now we get the expression for $F_{i,j}(\theta)$, which is shown below:
\begin{equation}
    F_{i,j}(\theta)=\frac{1}{2}\frac{d+v}{d+v+2}tr(\frac{\partial \Sigma}{\partial \theta_{i}}\Sigma^{-1}\frac{\partial \Sigma}{\partial \theta_{j}}\Sigma^{-1}) - \frac{1}{2(d+v+2)}tr(\frac{\partial \Sigma}{\partial \theta_{i}}\Sigma^{-1})tr(\frac{\partial \Sigma}{\partial \theta_{j}}\Sigma^{-1}) +
    \frac{d+v}{d+v+2}(\frac{\partial{\mu}}{\partial{\theta_{i}}})^{T}\Sigma^{-1}(\frac{\partial{\mu}}{\partial{\theta_{j}}})
\end{equation}
Note that the parameter theta of $F$ includes the mean $\mu$(we will use $m$ later) of the distribution at this point and the transformation matrix $\Sigma$ (symmetric matrix), we can think of $F$ as a matrix whose rows and columns are $d^2 + d$. At this point $F$ is a diagonal matrix, and we can simply find the inverse of its two bins. Here we make an approximation by considering the second term of $F_{i,j}$ as 0. This can be done because in general it can be assumed that the two-paradigm number of the transformation matrix is strictly bounded, so that the trace of its inverse matrix will be strictly bounded as well, and the term $d+v+2$ will be no less than $v$ (the degrees of freedom), and we can see in our experiments that $v$ will be large enough in general; and also for the convenience of the theoretical derivation.
According to equation $(4)$, we have:
\begin{equation}
    F_{m} = \frac{d+v}{d+v+2}\Sigma^{-1}
\end{equation}
\begin{equation}
    F_{m}^{-1} = \frac{d+v+2}{d+v}\Sigma
\end{equation}
\begin{equation}
    \nabla_{m}lnP(x) = \frac{d+v}{v+(x-m)^T\Sigma^{-1}(x-m)}\Sigma^{-1}(x-m)
\end{equation}
So,
\begin{equation}
    F_{m}^{-1}\nabla_{m}lnP(x)=\frac{d+v+2}{s+v}(x-m)
\end{equation}
Similarly,
\begin{equation}
    F_{\Sigma}^{-1} = -2\frac{d+v+2}{d+v}(\frac{\partial vech(\Sigma)}{\partial\Sigma^T})^{-1}(\frac{\partial vech(2\Sigma^{-1} - diag(\Sigma^{-1})}{\partial\Sigma^T})^{T}
\end{equation}
\begin{equation}
\begin{split}
    \nabla_{\Sigma}lnP(x) 
    =& -\frac{1}{2}tr(\frac{\partial{\Sigma}}{\partial{\theta}}*\Sigma^{-1}) - \frac{d+v}{2(v+s)})(x-m)^T\frac{\partial{\Sigma^{-1}}}{\partial{\theta}}(x-m) \\
    =& -\frac{1}{2}tr(\frac{\partial{\Sigma^{-1}}}{\partial{\Sigma}})(\frac{d+v}{v+s}(x-m)(x-m)^T-\Sigma) \\
    =& -\frac{1}{2}(\frac{\partial vech(2\Sigma^{-1} - diag(\Sigma^{-1})}{\partial\Sigma^T})^{T}(\frac{d+v}{v+s}(x-m)(x-m)^T-\Sigma)
\end{split}
\end{equation}
So,
\begin{equation}
    F_{\Sigma}^{-1}\nabla_{\Sigma}lnP(x)=\frac{d+v+2}{d+v}(\frac{d+v}{v+s}(x-m)(x-m)^T-\Sigma)
\end{equation}
Where, $s = (x-u)^{T}\Sigma^{-1}(x-u)$

\section{Visualization}
We give here plots of the change in fitness values and the change in degrees of freedom for all functions of our TFWA on the two benchmarks CEC2013 and CEC2017(dim=30).

\begin{figure}[htbp]
\centering
\begin{subfigure}[b]{0.49\textwidth}
\includegraphics[width=\linewidth]{pics/CEC13_1.pdf}
\caption{CEC13 function 1}
\label{fig1}
\end{subfigure}
\begin{subfigure}[b]{0.49\textwidth}
\includegraphics[width=\linewidth]{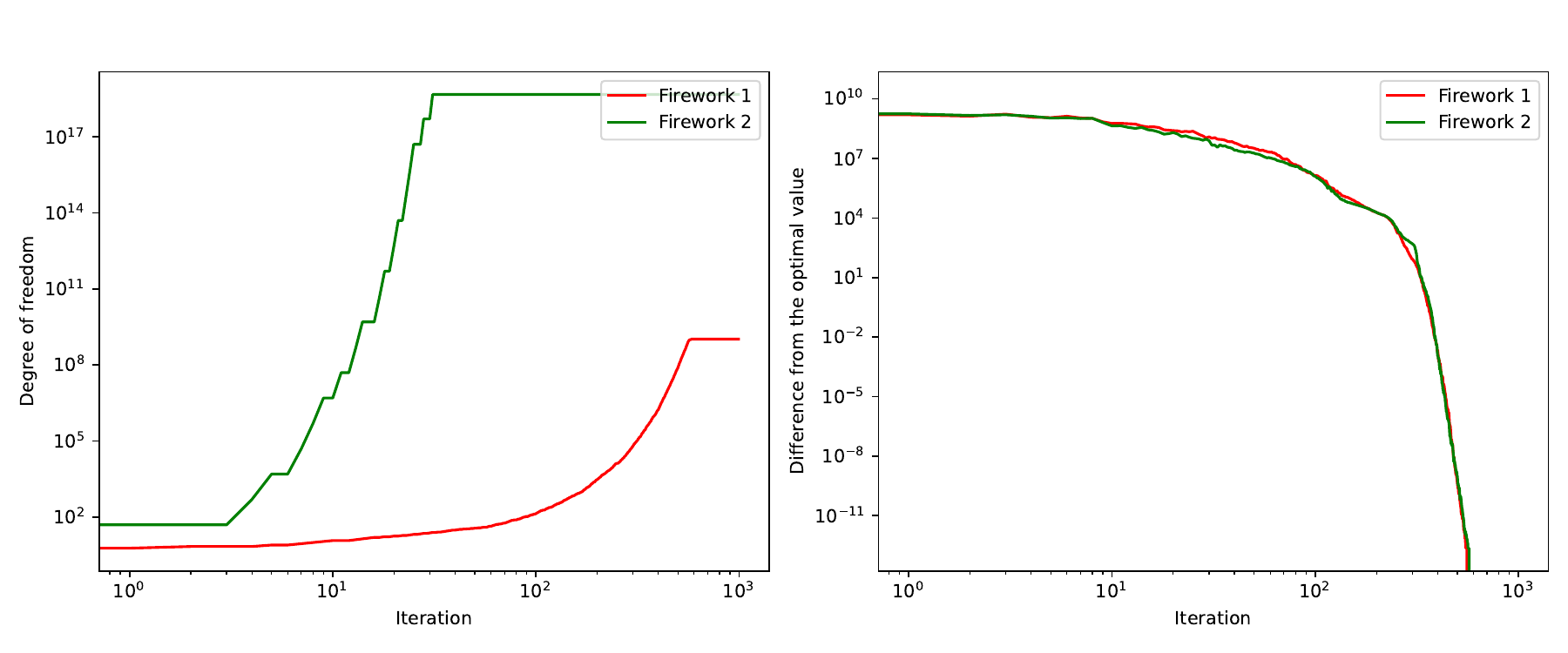}
\caption{CEC13 function 2}
\label{fig2}
\end{subfigure}
\begin{subfigure}[b]{0.49\textwidth}
\includegraphics[width=\linewidth]{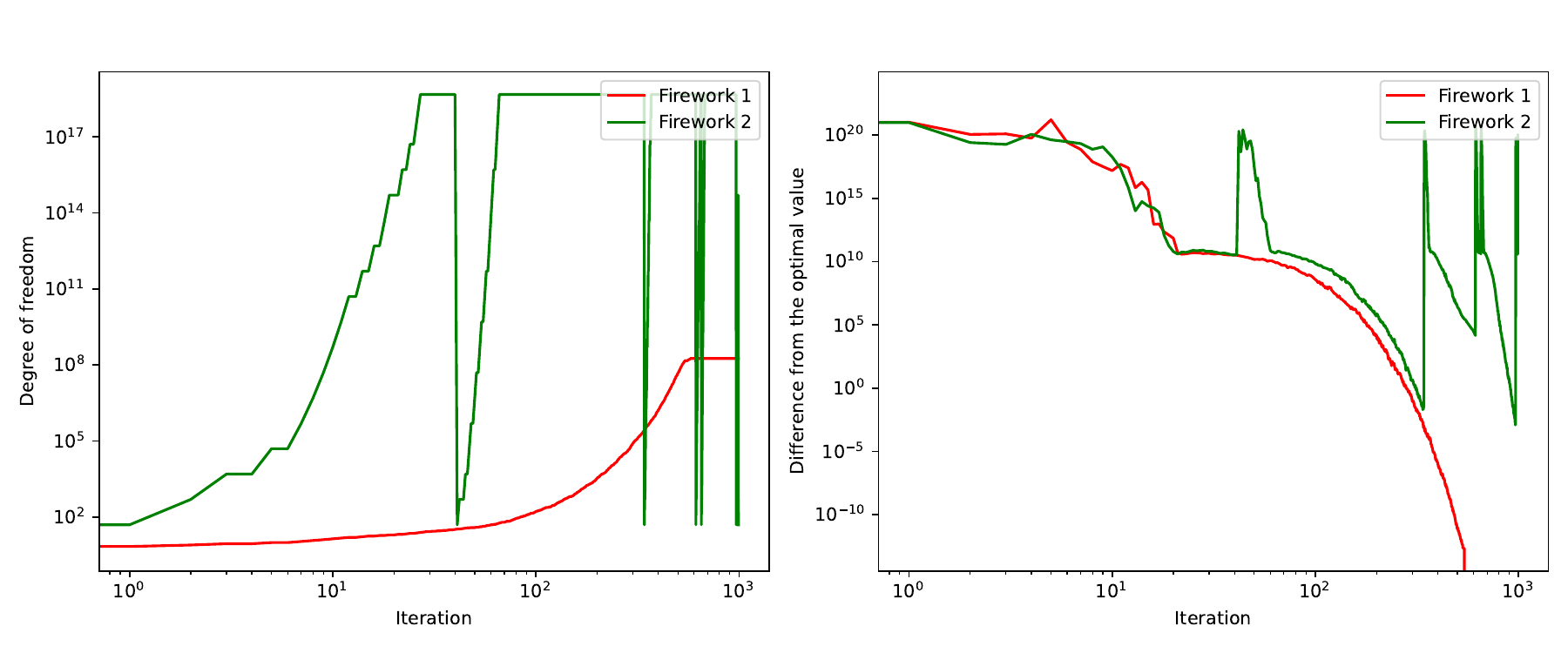}
\caption{CEC13 function 3}
\label{fig3}
\end{subfigure}
\begin{subfigure}[b]{0.49\textwidth}
\includegraphics[width=\linewidth]{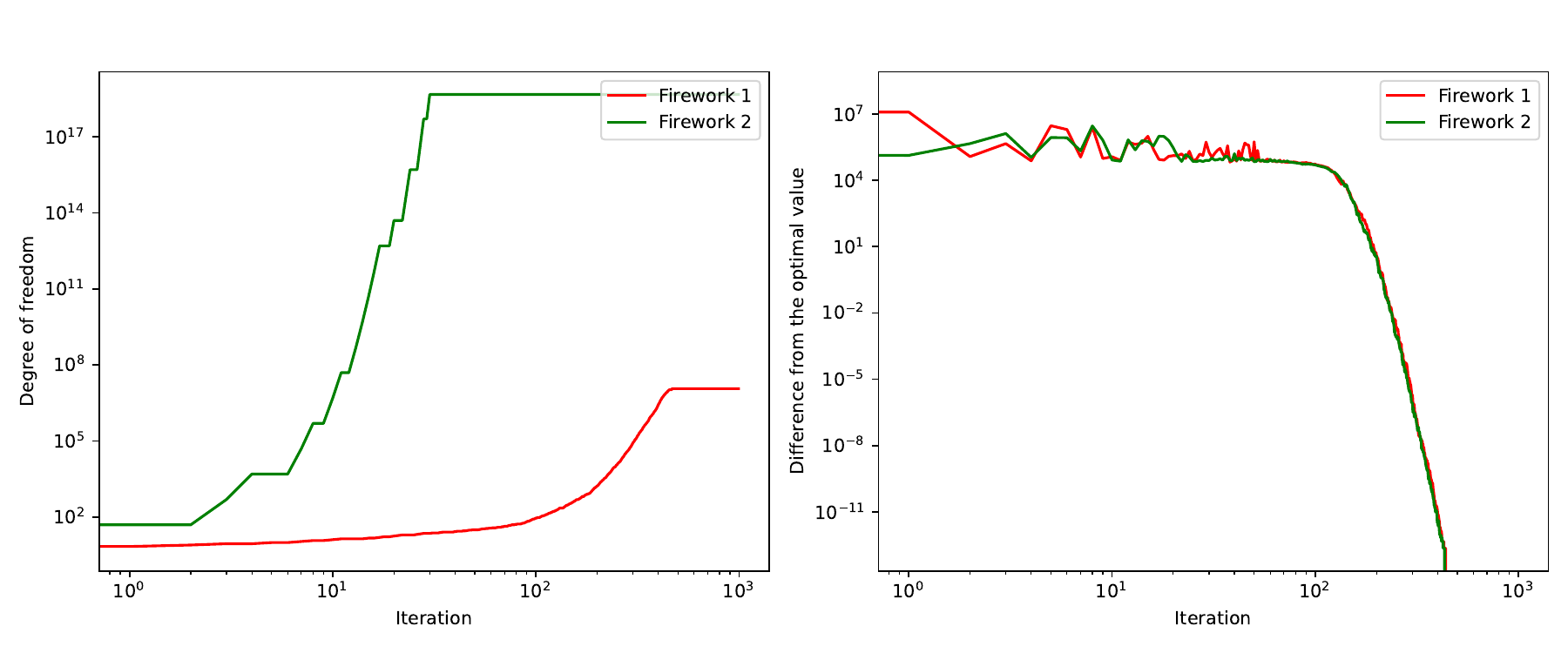}
\caption{CEC13 function 4}
\label{fig4}
\end{subfigure}
\begin{subfigure}[b]{0.49\textwidth}
\includegraphics[width=\linewidth]{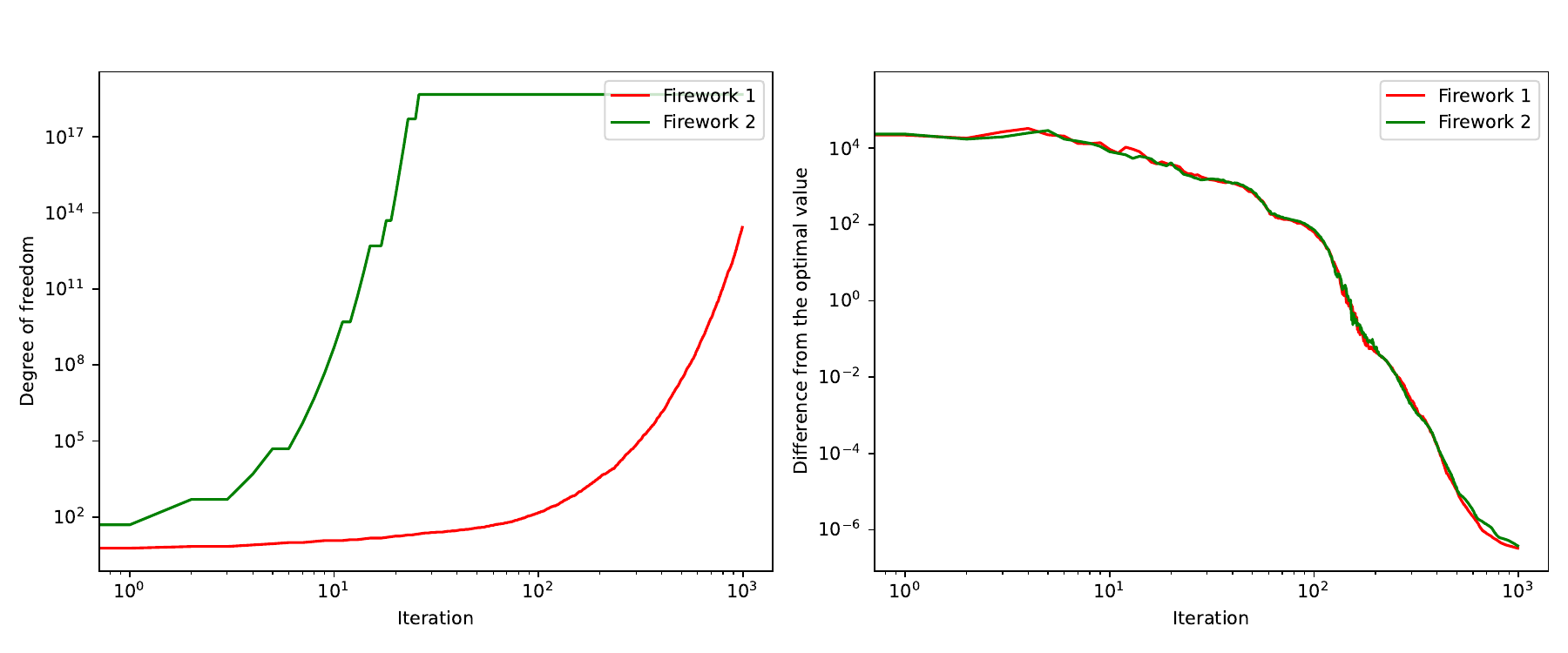}
\caption{CEC13 function 5}
\label{fig5}
\end{subfigure}
\begin{subfigure}[b]{0.49\textwidth}
\includegraphics[width=\linewidth]{pics/CEC13_6.pdf}
\caption{CEC13 function 6}
\label{fig6}
\end{subfigure}
\begin{subfigure}[b]{0.49\textwidth}
\includegraphics[width=\linewidth]{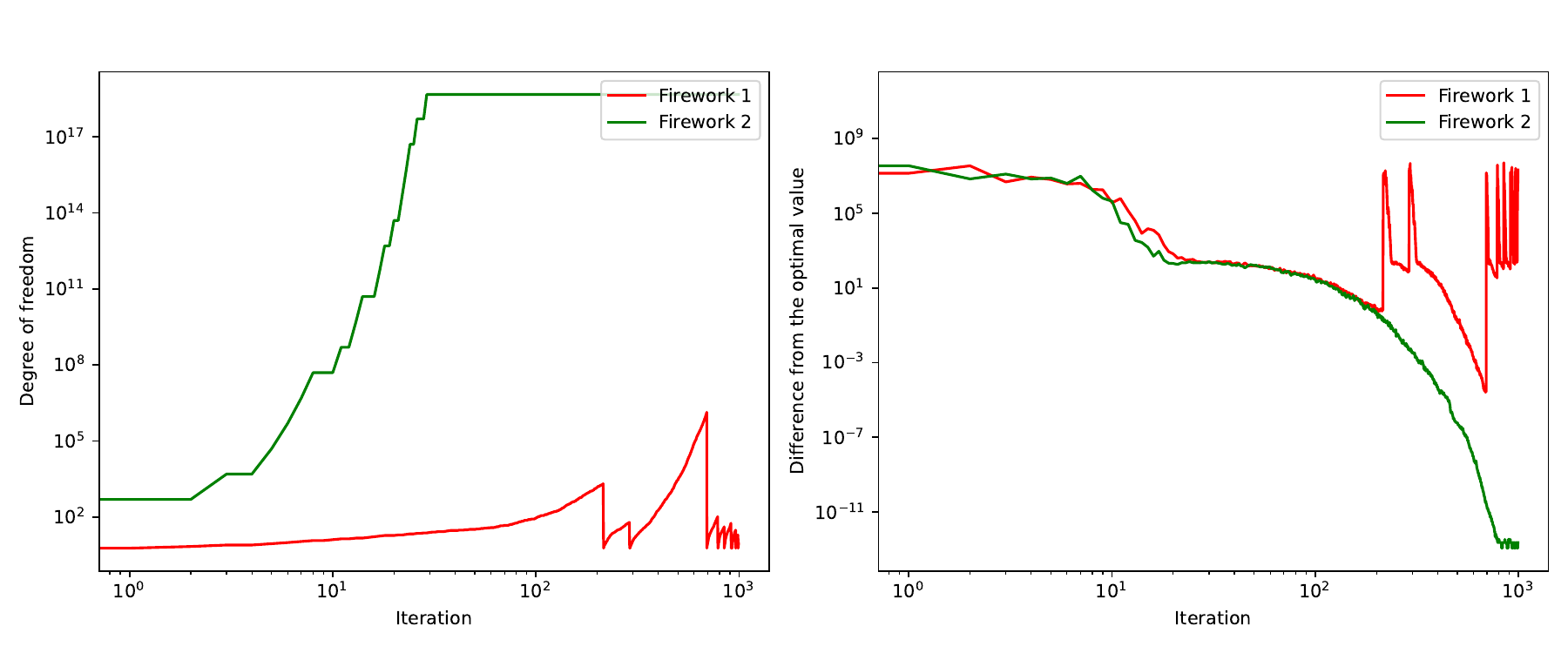}
\caption{CEC13 function 7}
\label{fig7}
\end{subfigure}
\begin{subfigure}[b]{0.49\textwidth}
\includegraphics[width=\linewidth]{pics/CEC13_8.pdf}
\caption{CEC13 function 8}
\label{fig8}
\end{subfigure}
\caption{CEC2013 f1-f8}
\end{figure}

\begin{figure}[htbp]
\centering
\begin{subfigure}[b]{0.49\textwidth}
\includegraphics[width=\linewidth]{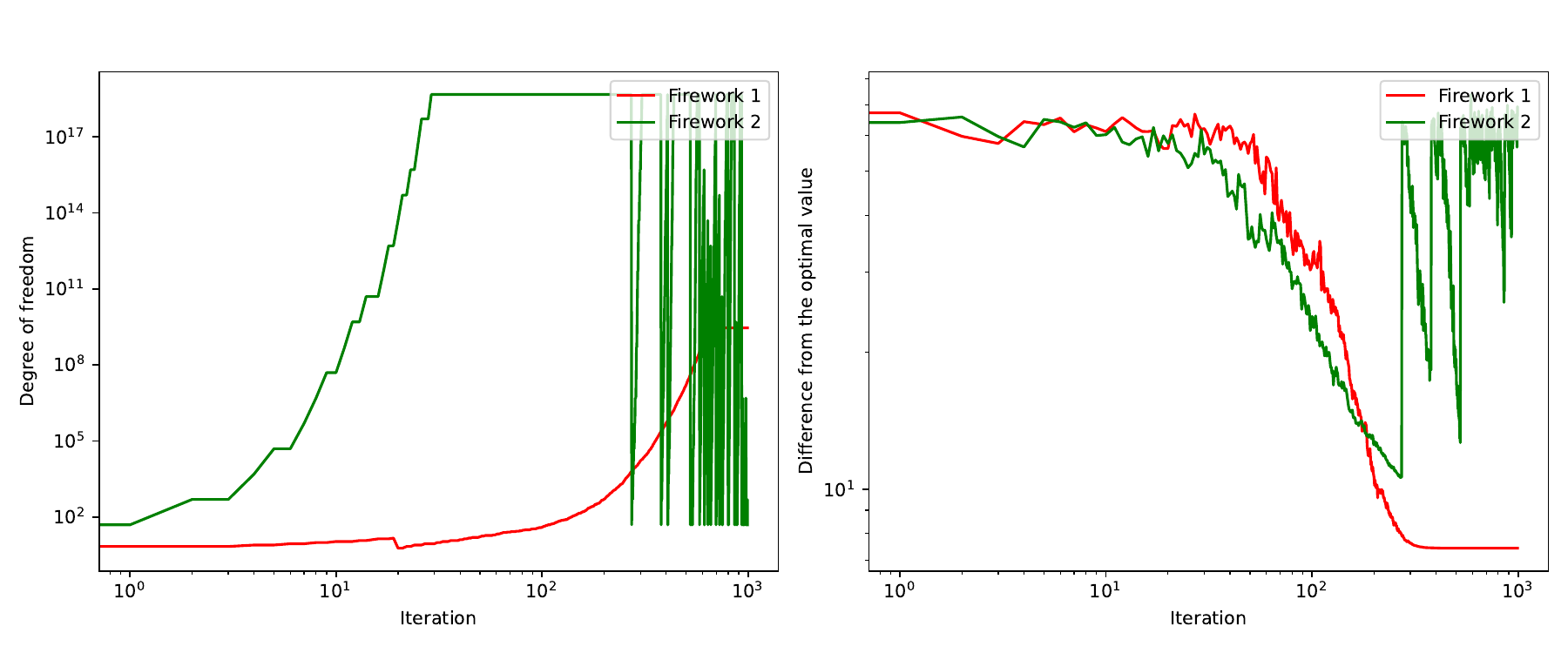}
\caption{CEC13 function 9}
\label{fig9}
\end{subfigure}
\begin{subfigure}[b]{0.49\textwidth}
\includegraphics[width=\linewidth]{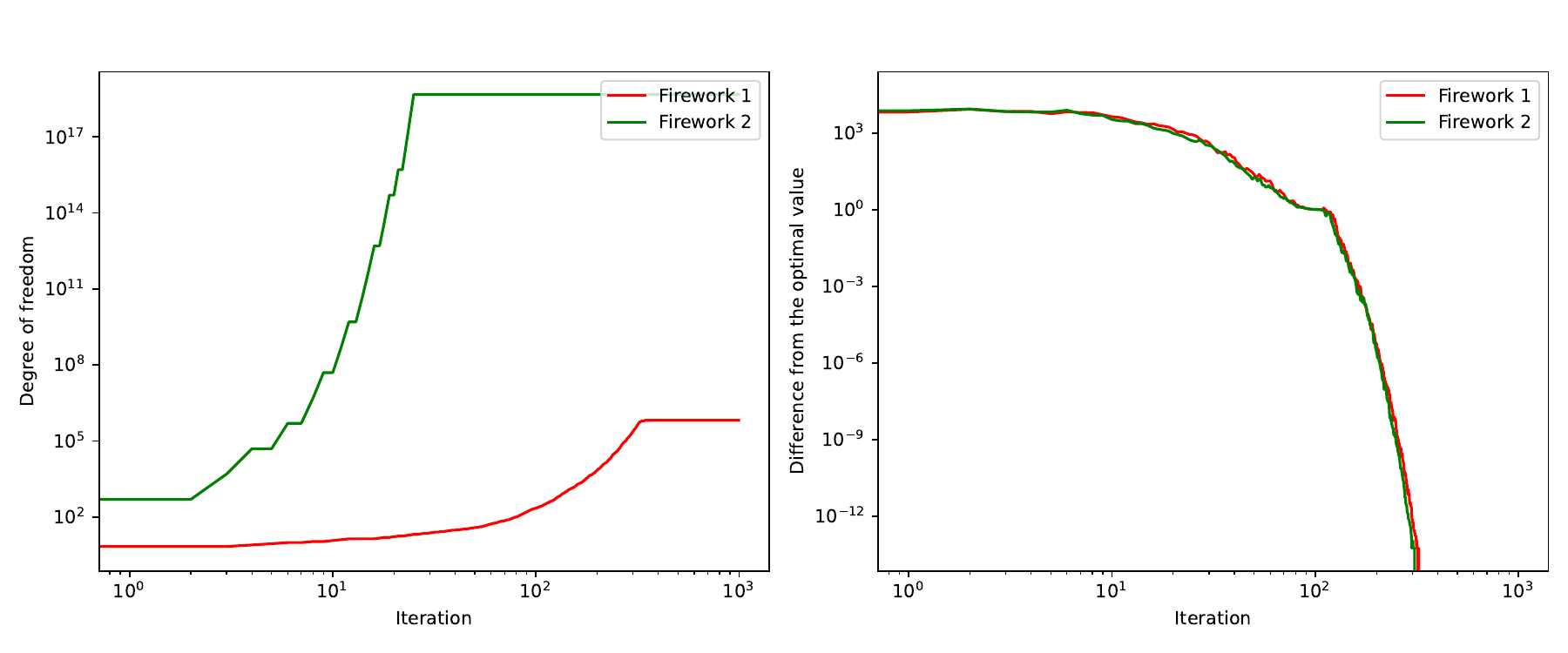}
\caption{CEC13 function 10}
\label{fig10}
\end{subfigure}
\begin{subfigure}[b]{0.49\textwidth}
\includegraphics[width=\linewidth]{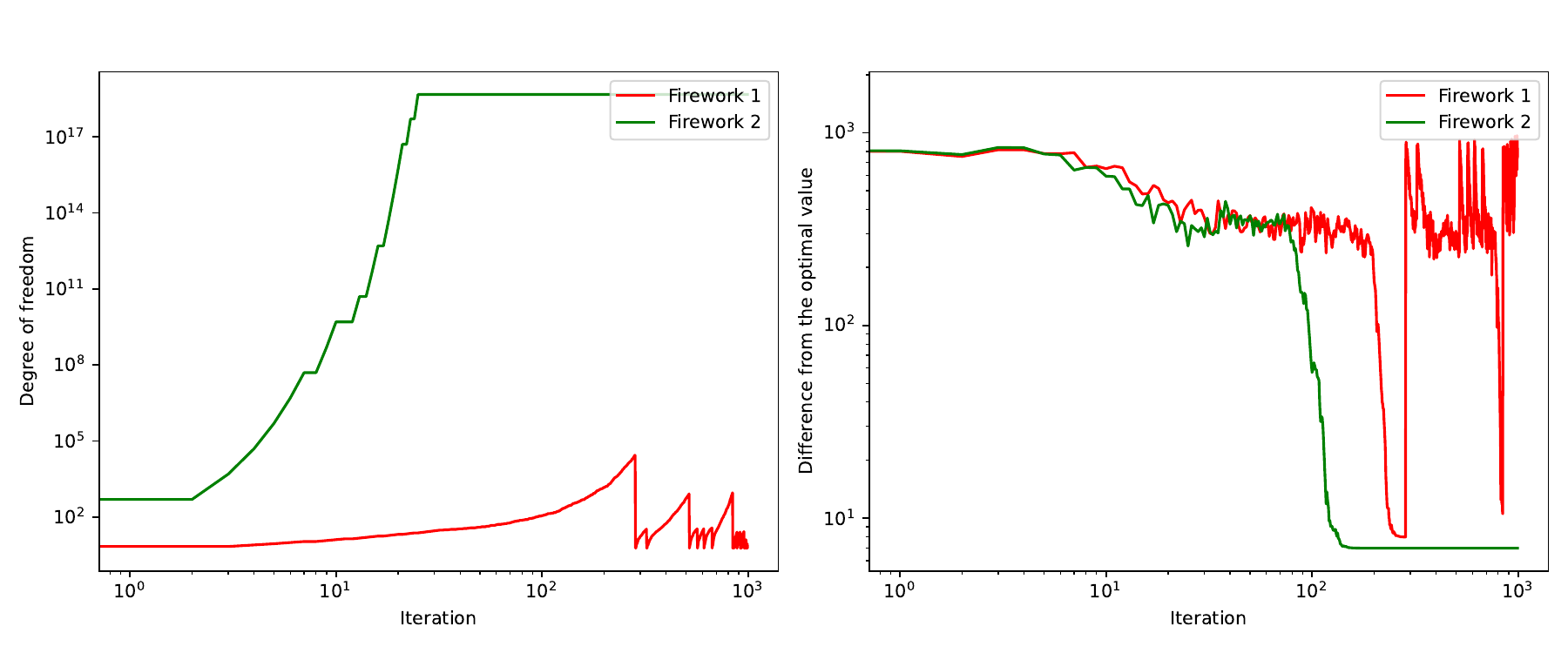}
\caption{CEC13 function 11}
\label{fig11}
\end{subfigure}
\begin{subfigure}[b]{0.49\textwidth}
\includegraphics[width=\linewidth]{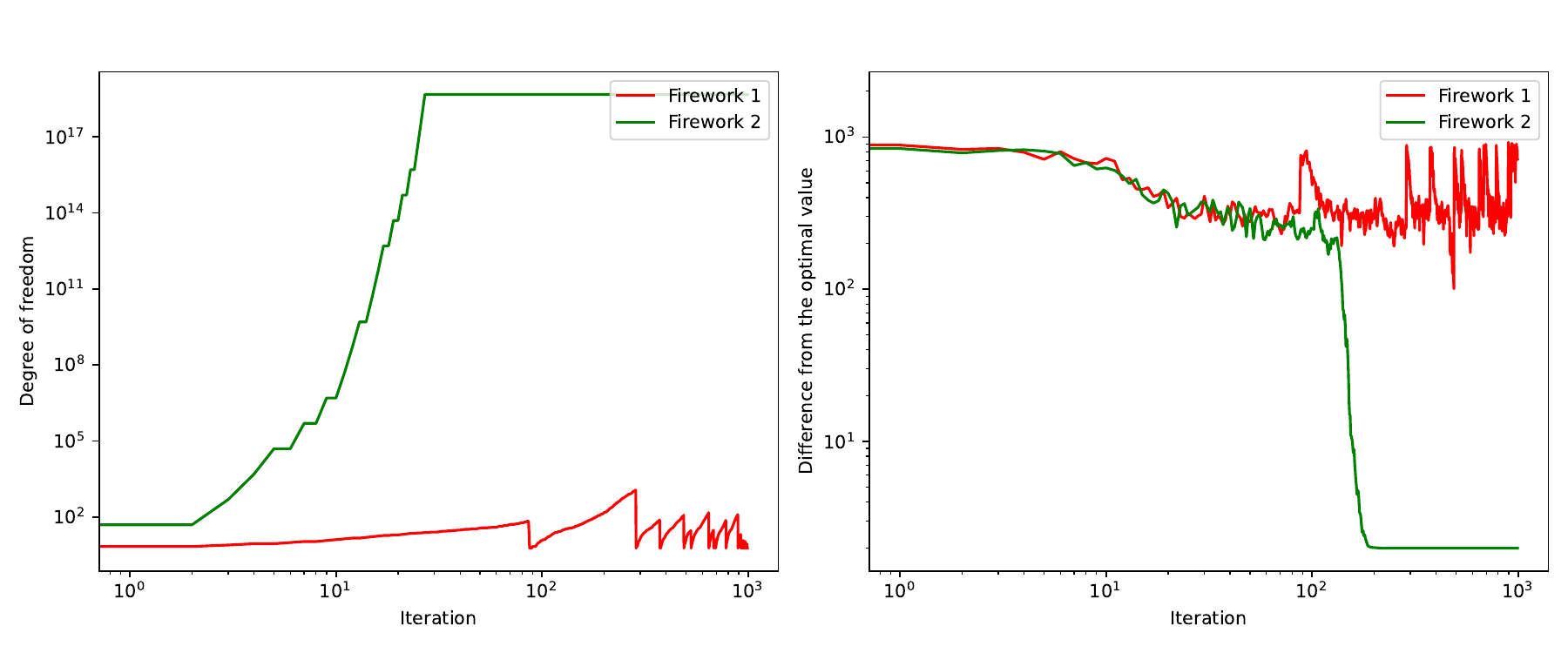}
\caption{CEC13 function 12}
\label{fig12}
\end{subfigure}
\begin{subfigure}[b]{0.49\textwidth}
\includegraphics[width=\linewidth]{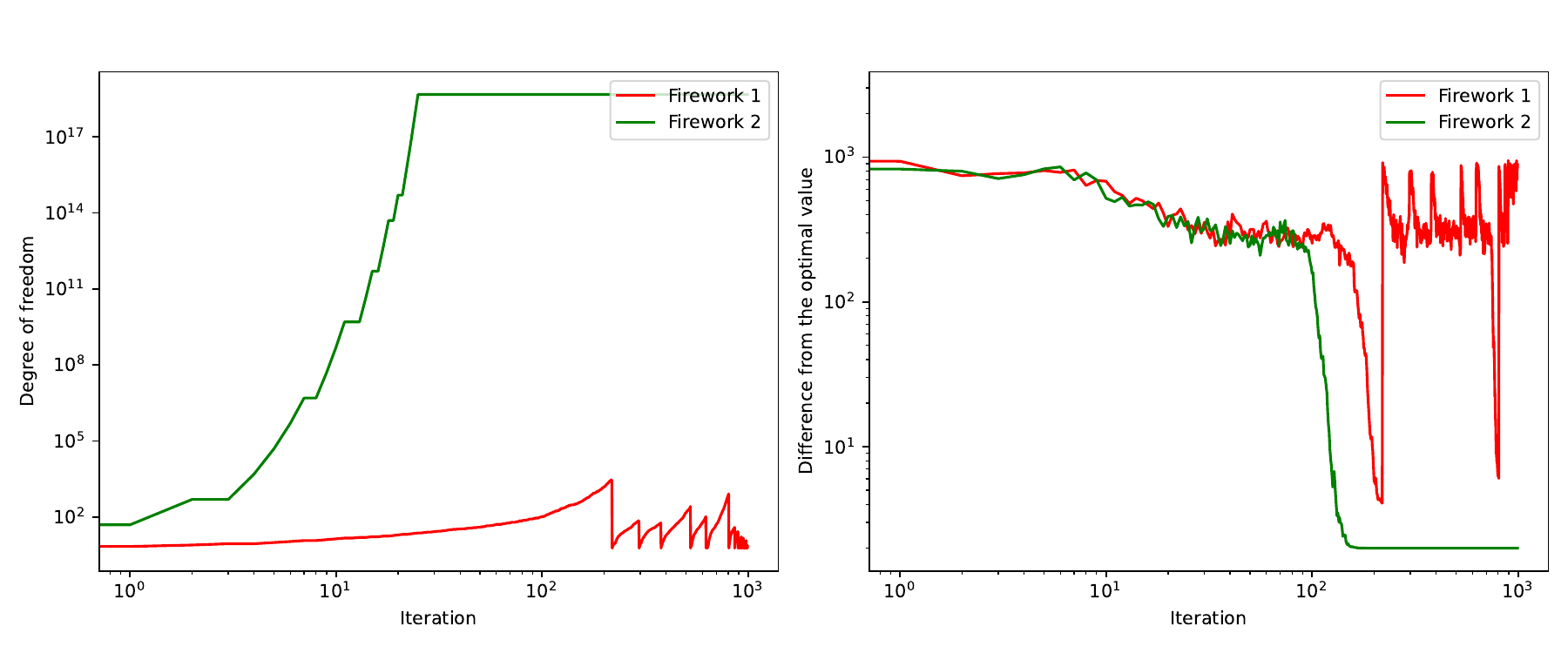}
\caption{CEC13 function 13}
\label{fig13}
\end{subfigure}
\begin{subfigure}[b]{0.49\textwidth}
\includegraphics[width=\linewidth]{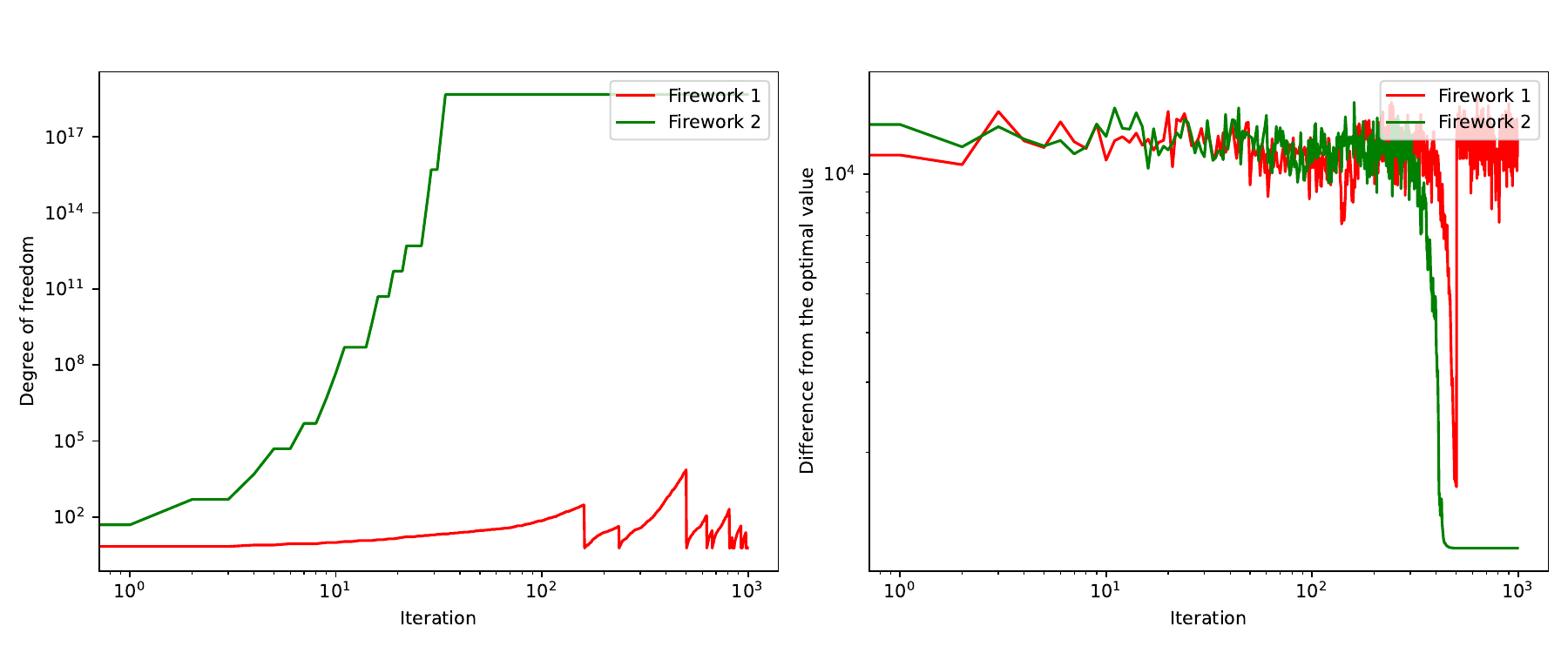}
\caption{CEC13 function 14}
\label{fig14}
\end{subfigure}
\begin{subfigure}[b]{0.49\textwidth}
\includegraphics[width=\linewidth]{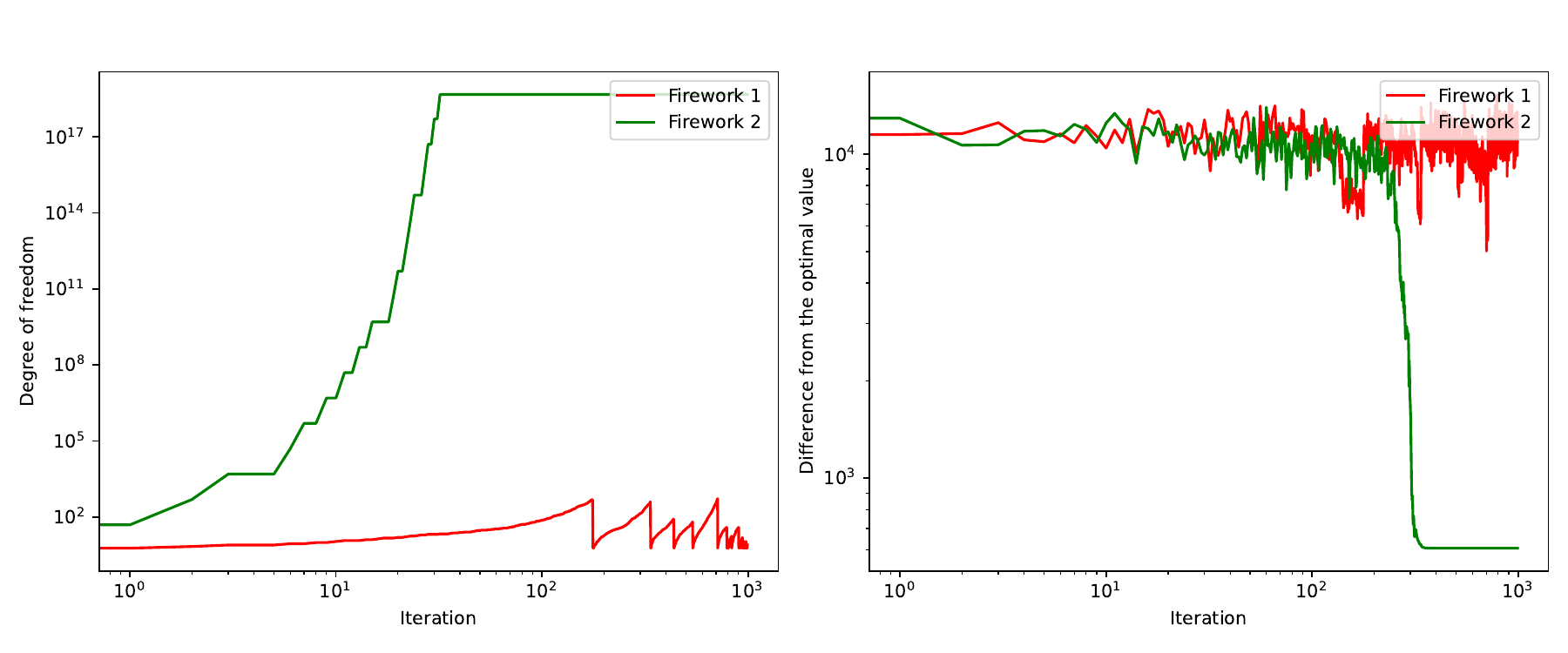}
\caption{CEC13 function 15}
\label{fig15}
\end{subfigure}
\begin{subfigure}[b]{0.49\textwidth}
\includegraphics[width=\linewidth]{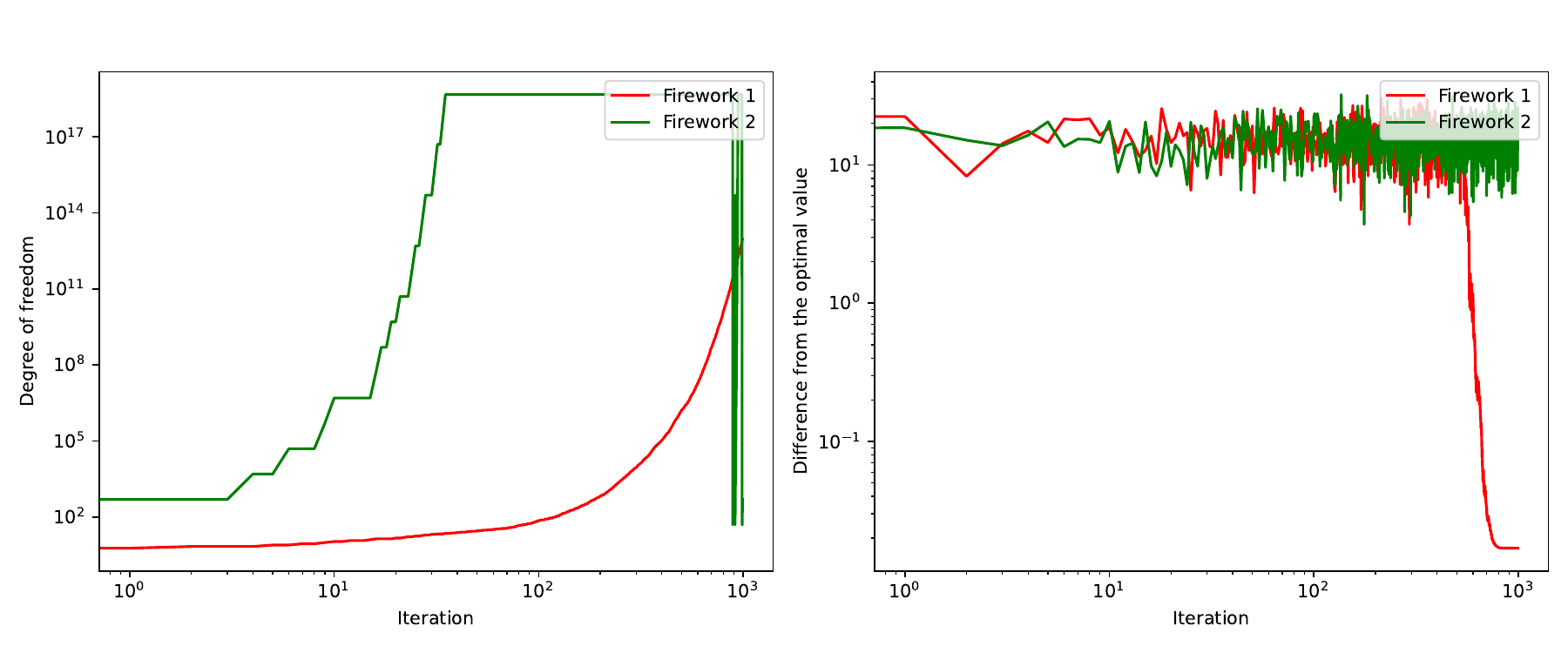}
\caption{CEC13 function 16}
\label{fig16}
\end{subfigure}
\caption{CEC2013 f9-f16}
\end{figure}

\begin{figure}[htbp]
\centering
\begin{subfigure}[b]{0.49\textwidth}
\includegraphics[width=\linewidth]{pics/CEC13_17.pdf}
\caption{CEC13 function 17}
\label{fig17}
\end{subfigure}
\begin{subfigure}[b]{0.49\textwidth}
\includegraphics[width=\linewidth]{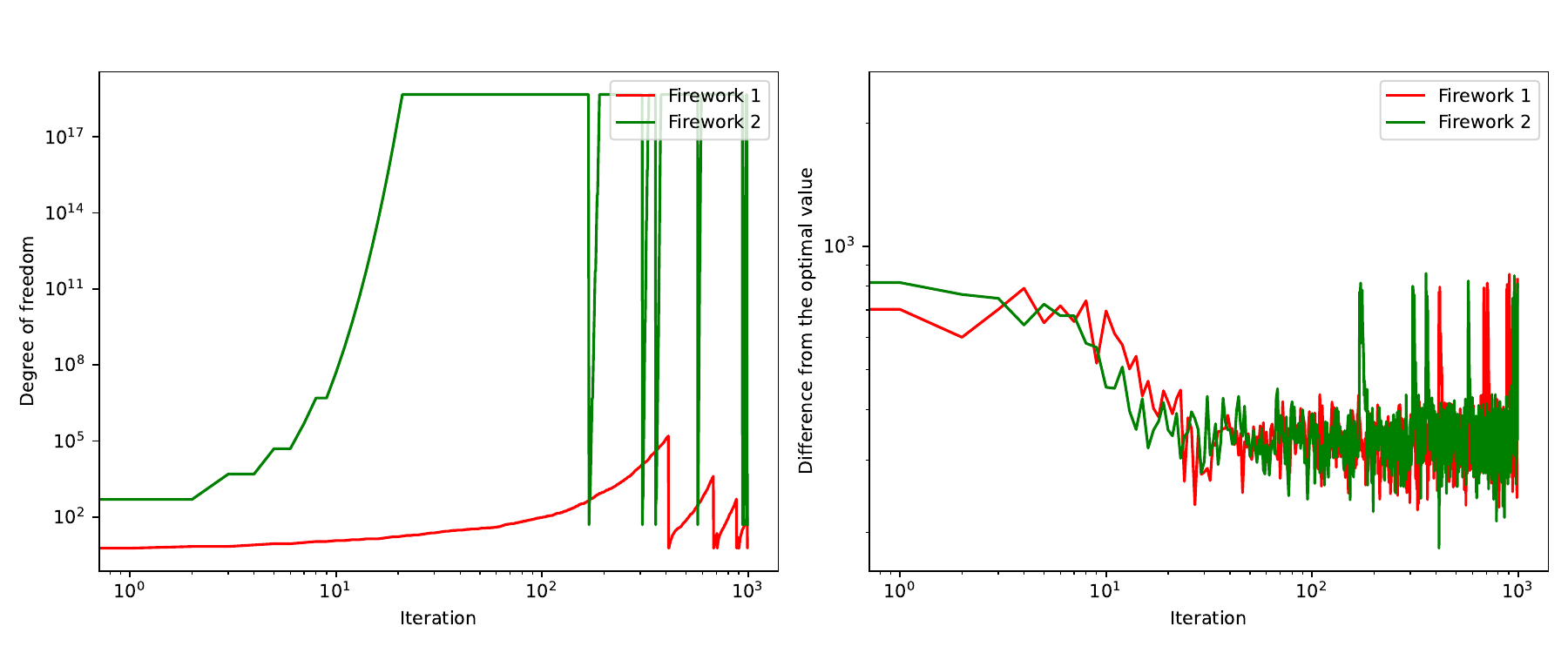}
\caption{CEC13 function 18}
\label{fig18}
\end{subfigure}
\begin{subfigure}[b]{0.49\textwidth}
\includegraphics[width=\linewidth]{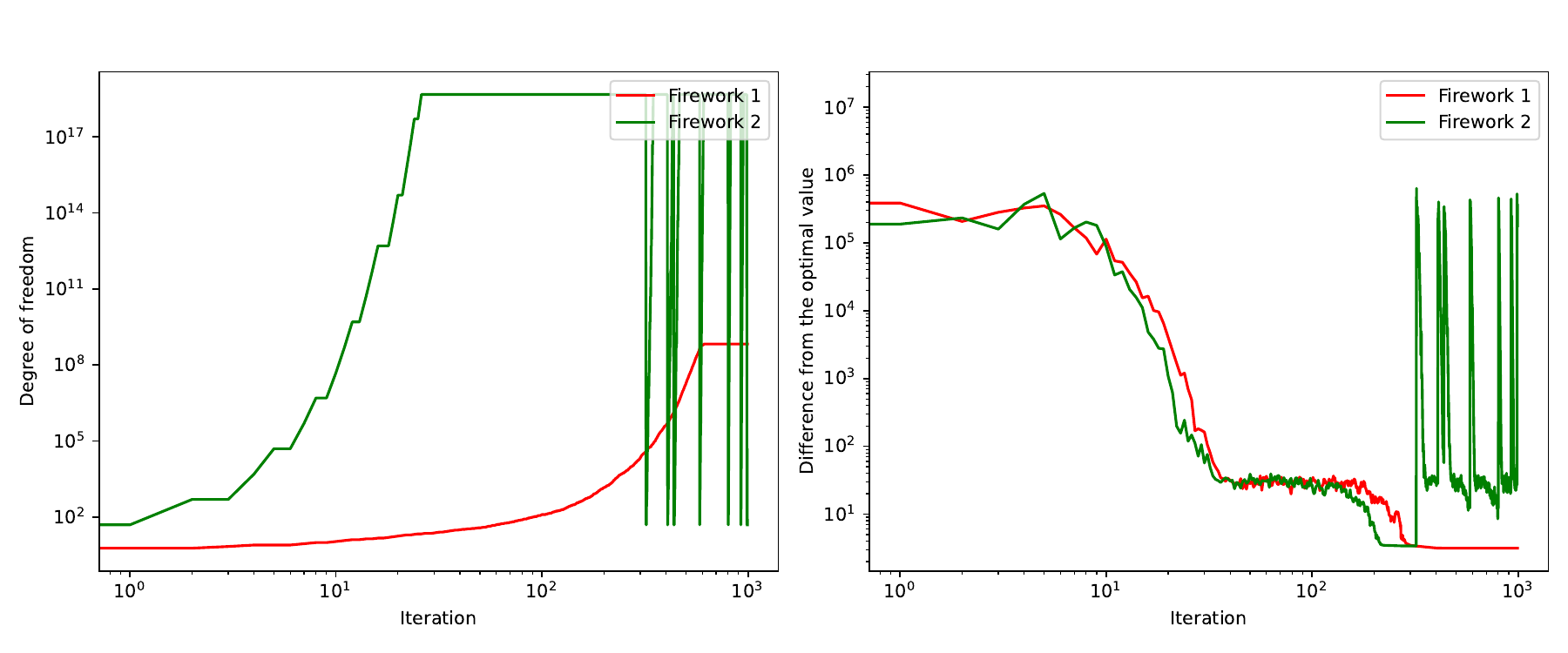}
\caption{CEC13 function 19}
\label{fig19}
\end{subfigure}
\begin{subfigure}[b]{0.49\textwidth}
\includegraphics[width=\linewidth]{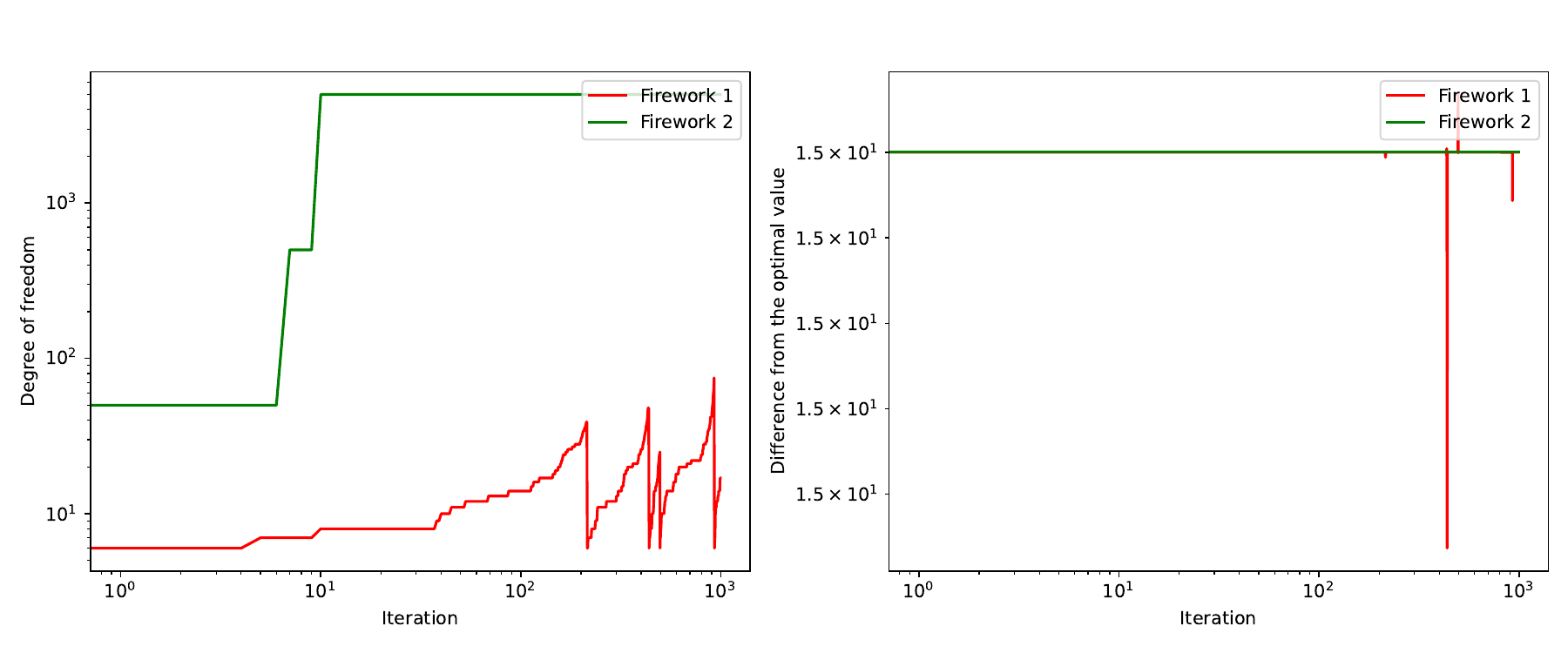}
\caption{CEC13 function 20}
\label{fig20}
\end{subfigure}
\begin{subfigure}[b]{0.49\textwidth}
\includegraphics[width=\linewidth]{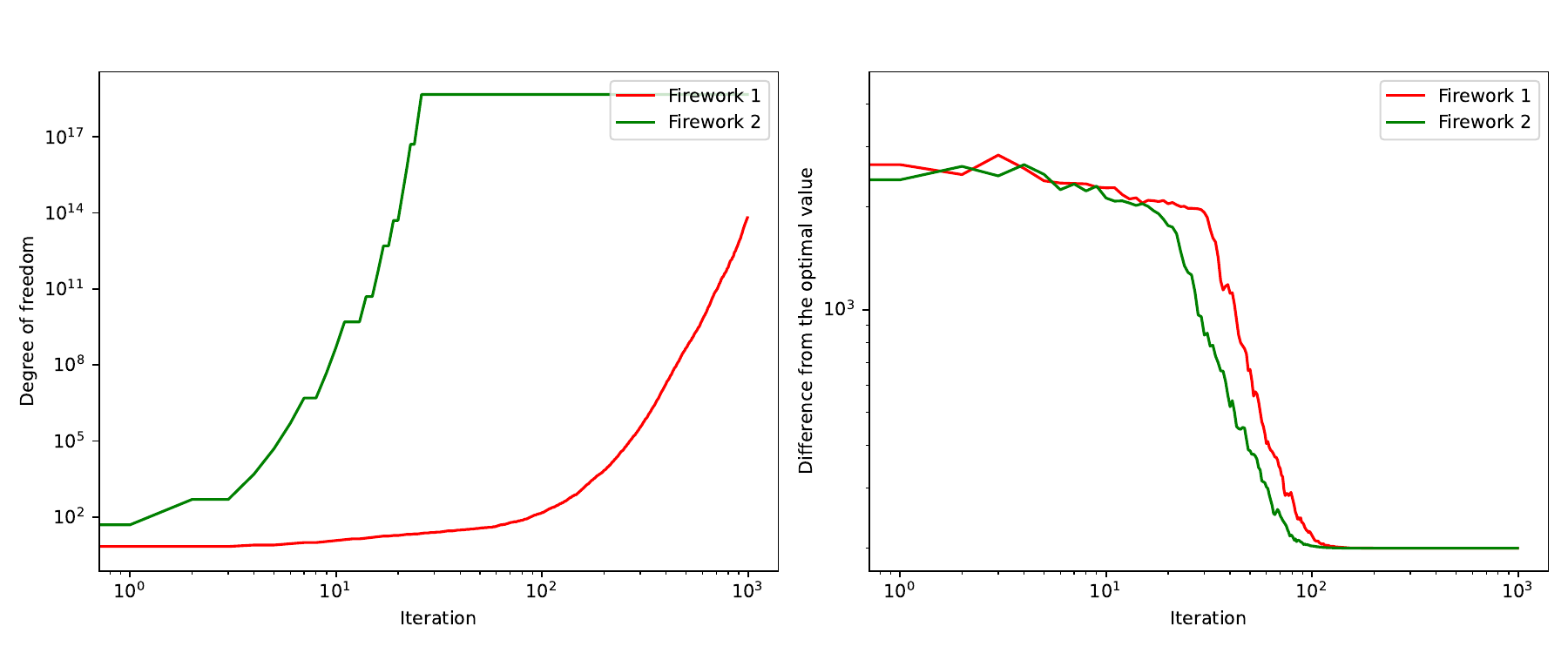}
\caption{CEC13 function 21}
\label{fig21}
\end{subfigure}
\begin{subfigure}[b]{0.49\textwidth}
\includegraphics[width=\linewidth]{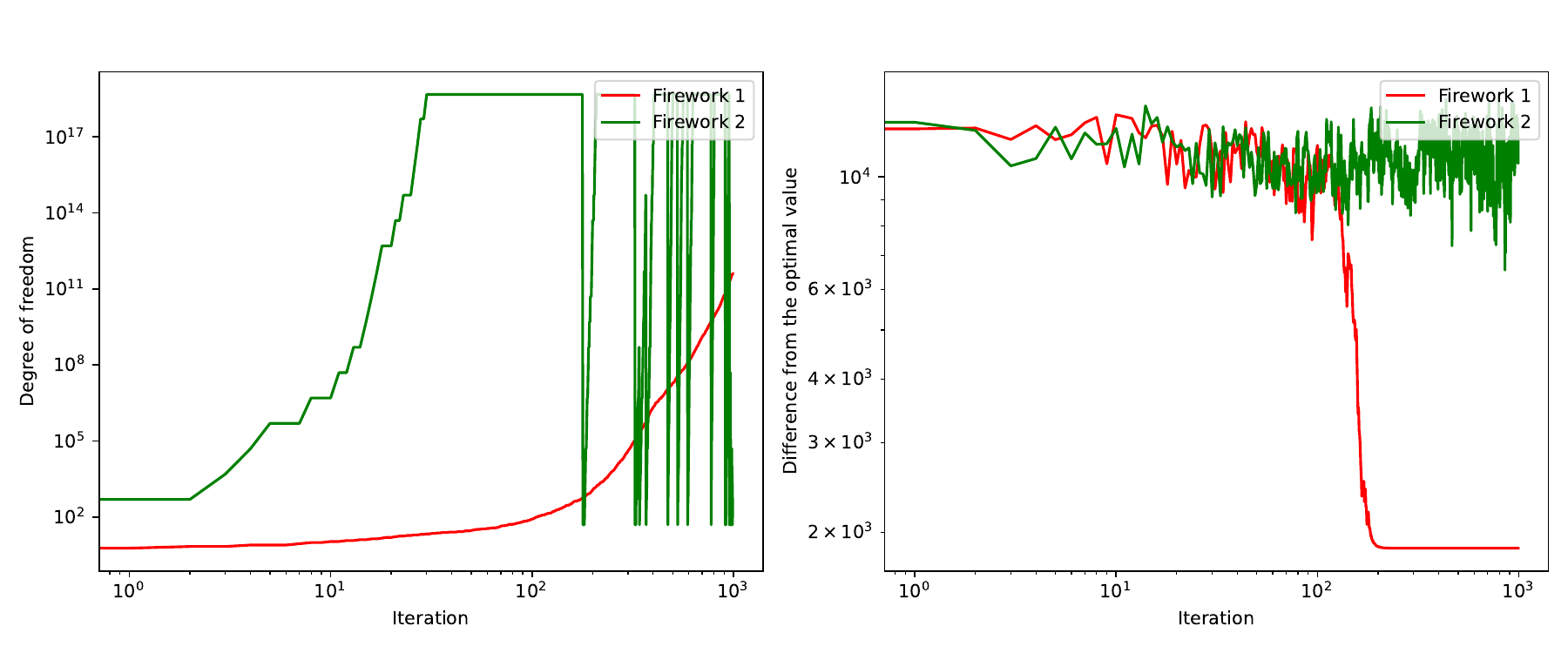}
\caption{CEC13 function 22}
\label{fig22}
\end{subfigure}
\begin{subfigure}[b]{0.49\textwidth}
\includegraphics[width=\linewidth]{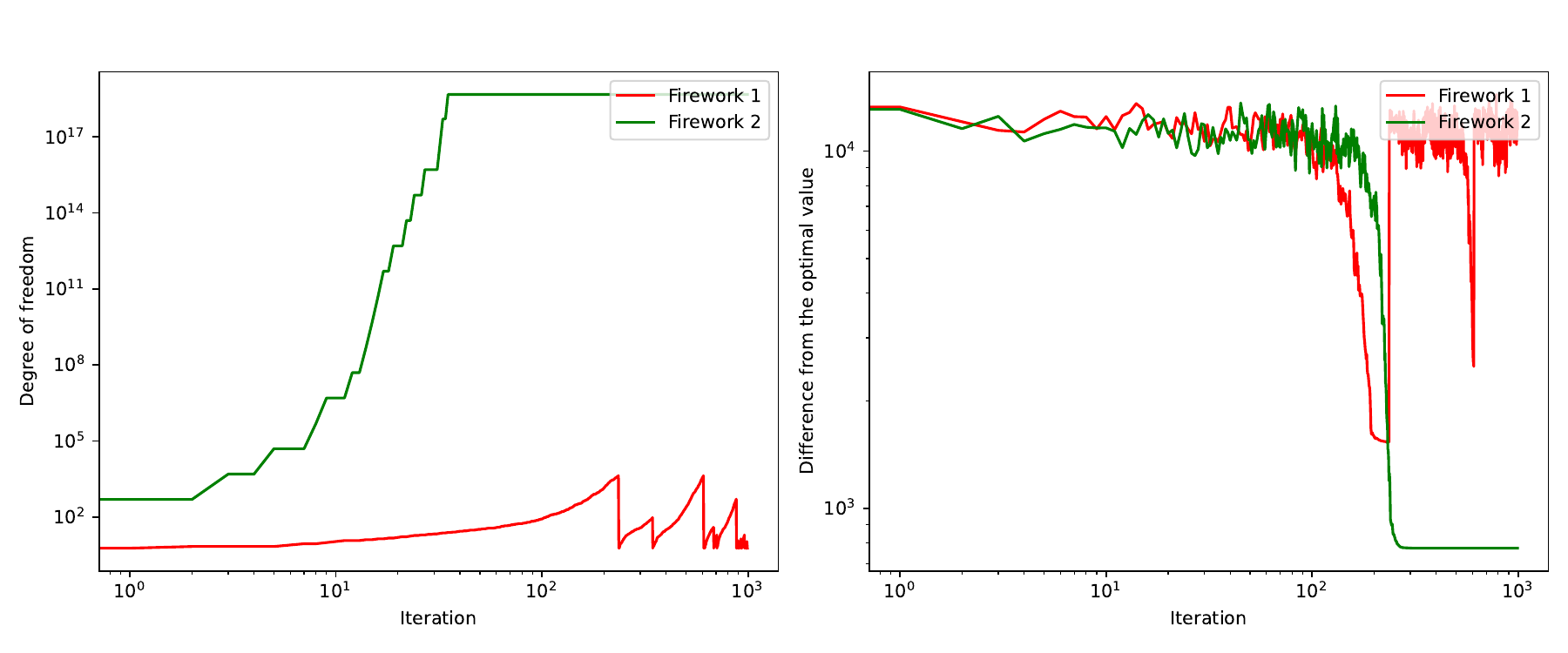}
\caption{CEC13 function 23}
\label{fig23}
\end{subfigure}
\begin{subfigure}[b]{0.49\textwidth}
\includegraphics[width=\linewidth]{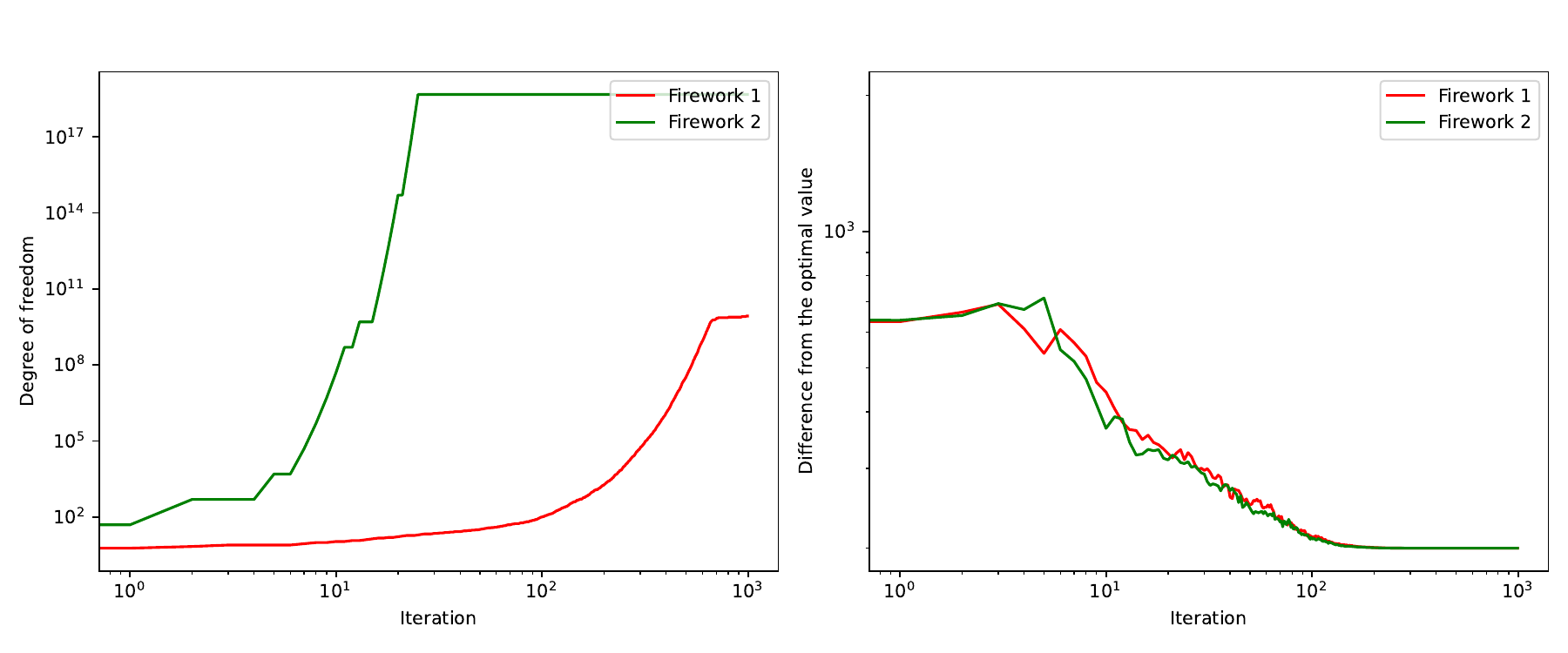}
\caption{CEC13 function 24}
\label{fig24}
\end{subfigure}
\caption{CEC2013 f17-f24}
\end{figure}

\begin{figure}[htbp]
\centering
\begin{subfigure}[b]{0.49\textwidth}
\includegraphics[width=\linewidth]{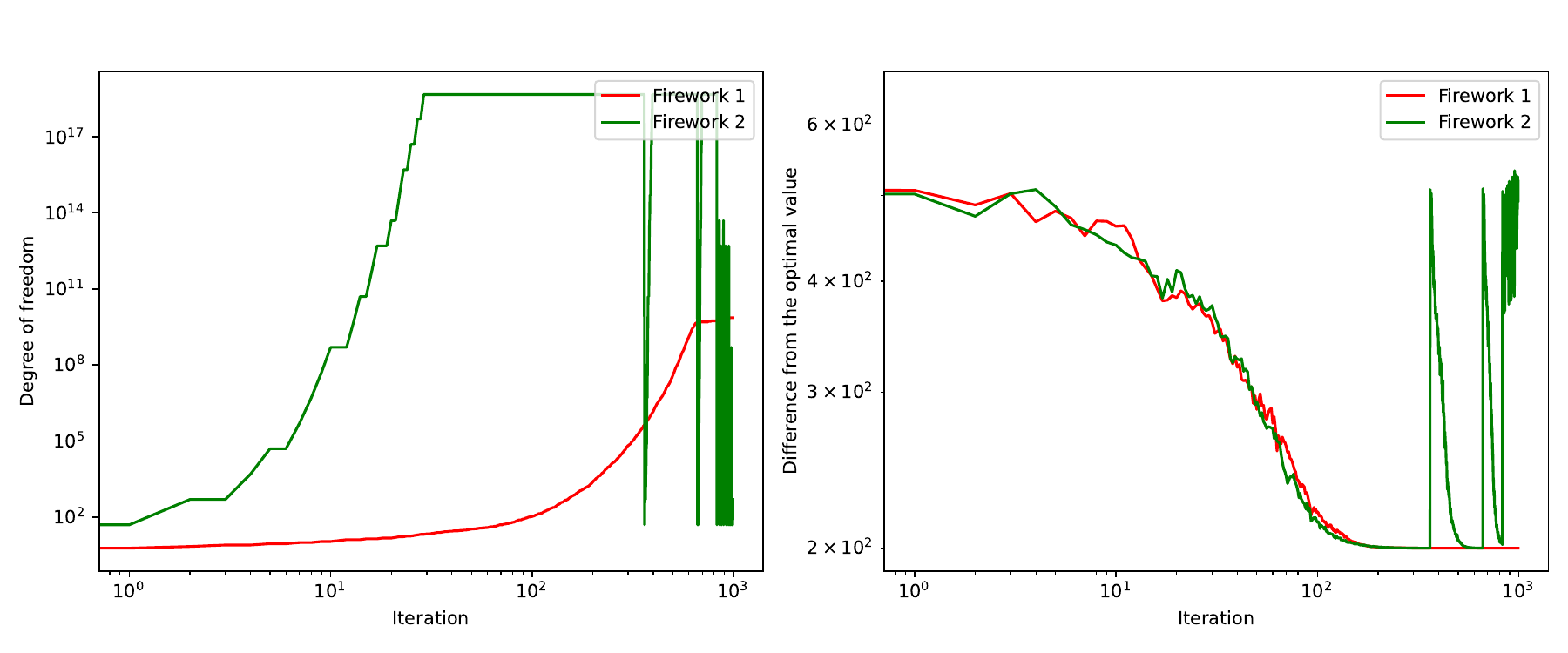}
\caption{CEC13 function 25}
\label{fig25}
\end{subfigure}
\begin{subfigure}[b]{0.49\textwidth}
\includegraphics[width=\linewidth]{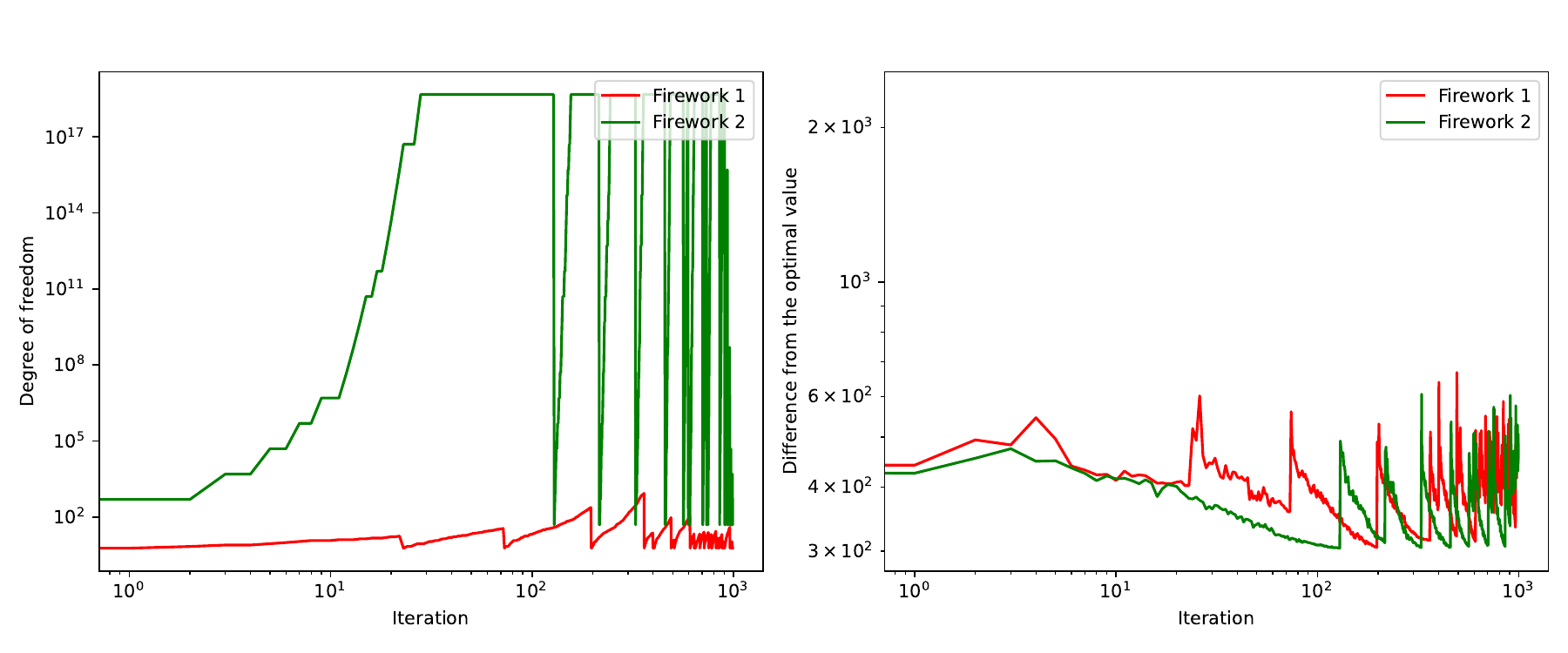}
\caption{CEC13 function 26}
\label{fig26}
\end{subfigure}
\begin{subfigure}[b]{0.49\textwidth}
\includegraphics[width=\linewidth]{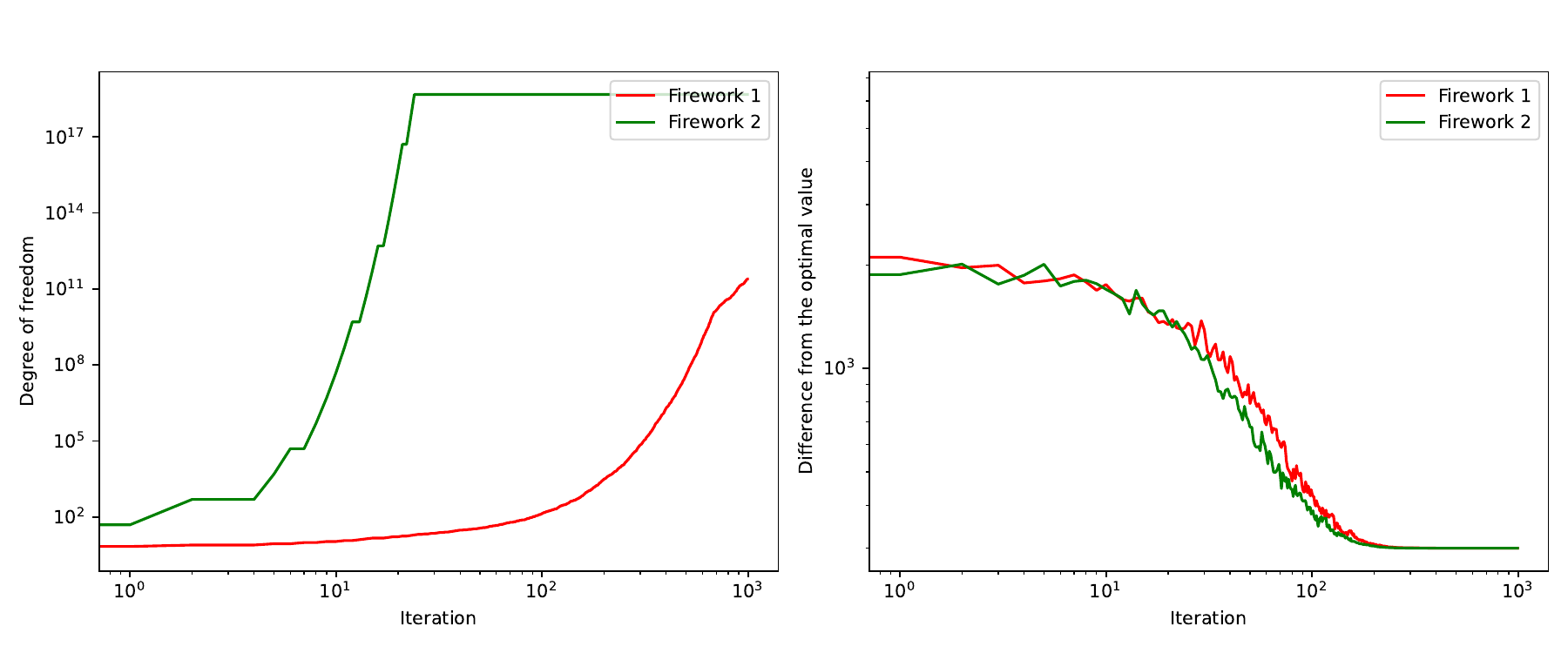}
\caption{CEC13 function 27}
\label{fig27}
\end{subfigure}
\begin{subfigure}[b]{0.49\textwidth}
\includegraphics[width=\linewidth]{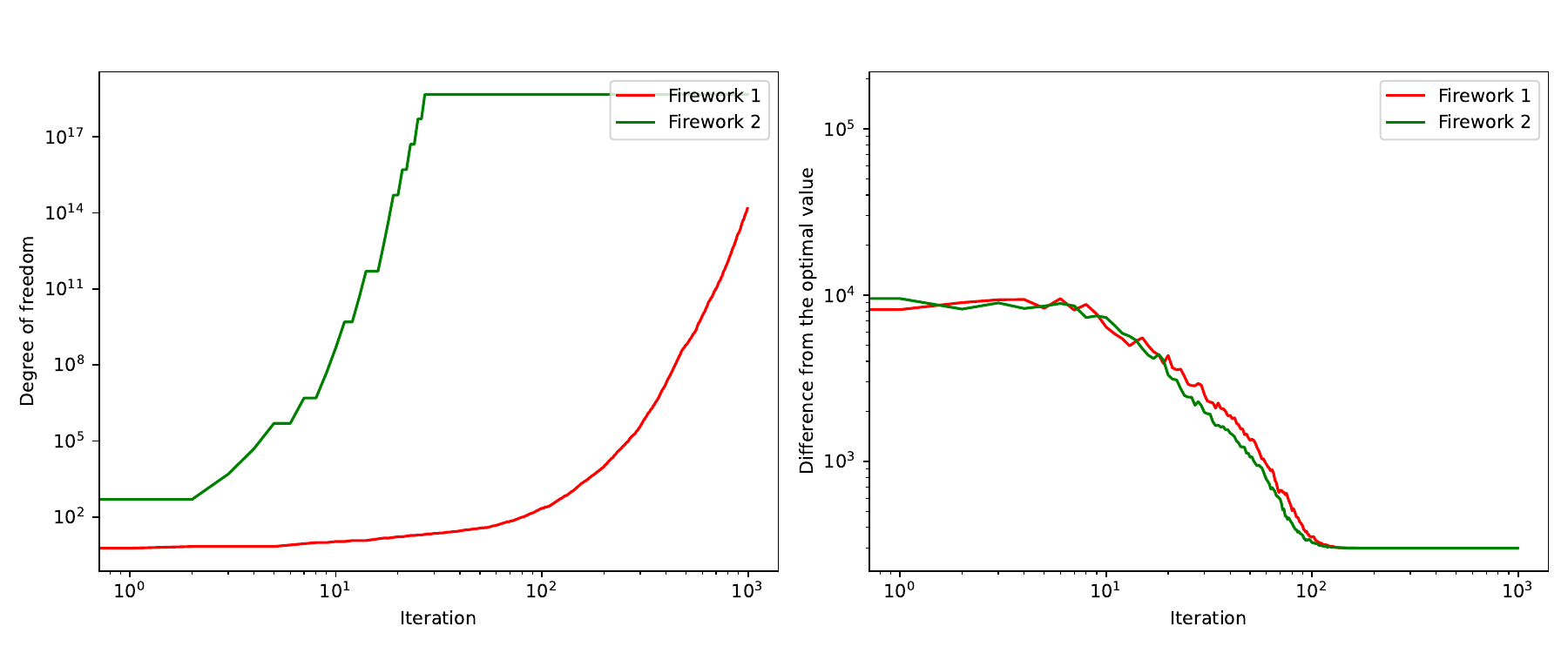}
\caption{CEC13 function 28}
\label{fig28}
\end{subfigure}
\caption{CEC2013 f25-f28}
\end{figure}

\begin{figure}[htbp]
\centering
\begin{subfigure}[b]{0.49\textwidth}
\includegraphics[width=\linewidth]{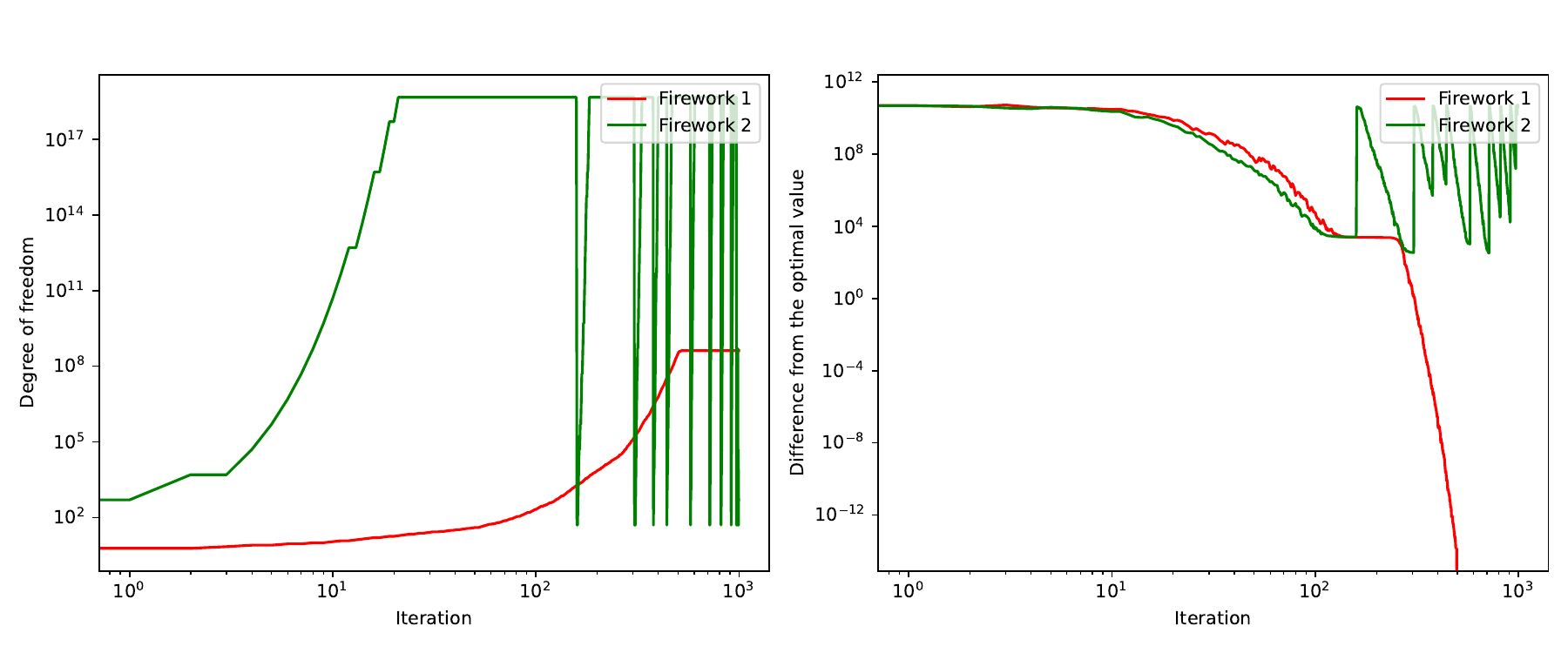}
\caption{CEC17 function 1}
\label{fig17_1}
\end{subfigure}
\begin{subfigure}[b]{0.49\textwidth}
\includegraphics[width=\linewidth]{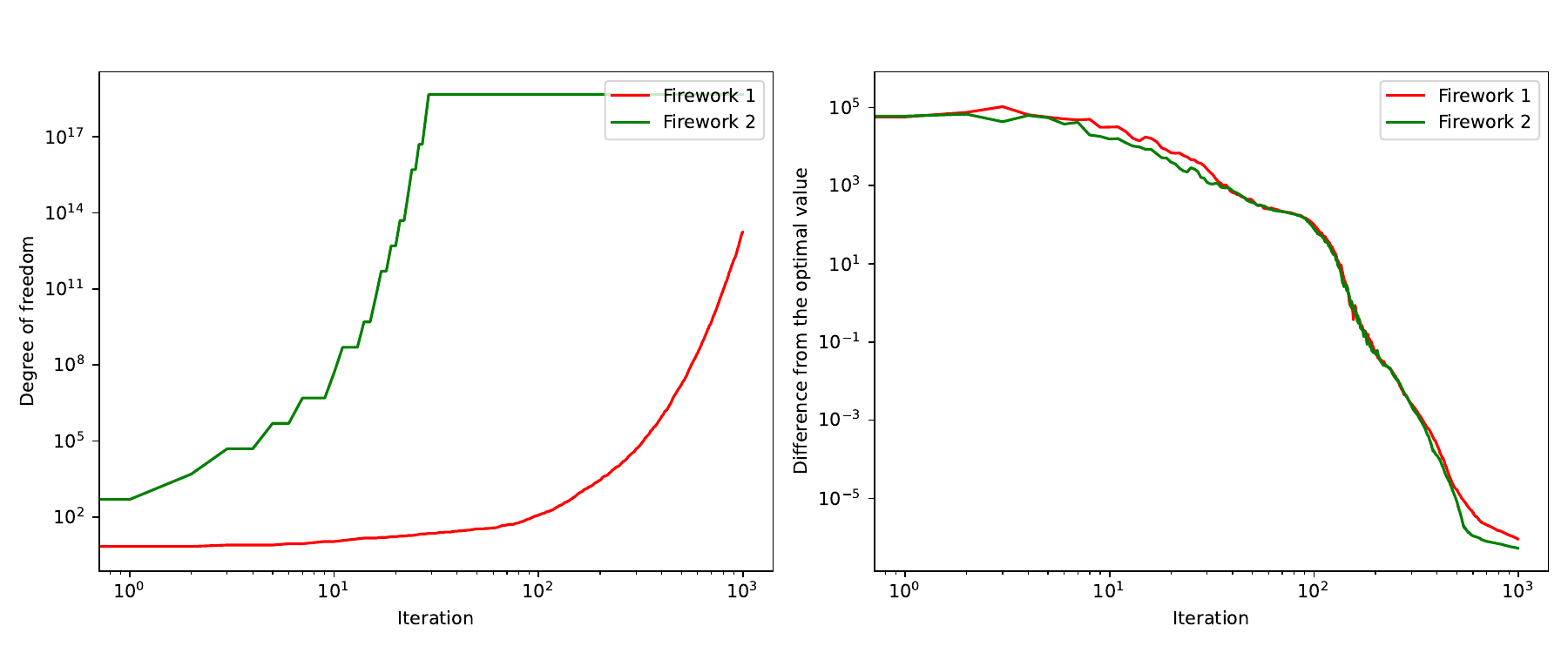}
\caption{CEC17 function 2}
\label{fig17_2}
\end{subfigure}
\begin{subfigure}[b]{0.49\textwidth}
\includegraphics[width=\linewidth]{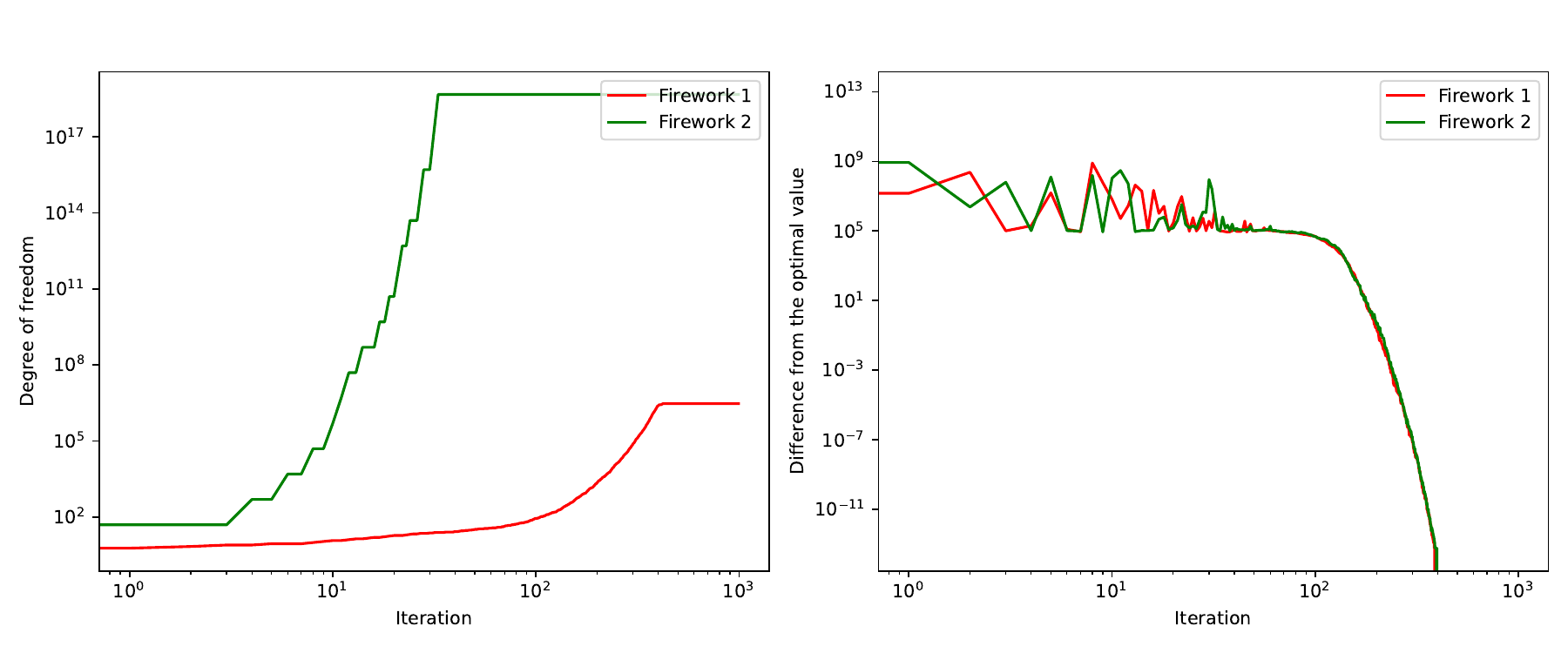}
\caption{CEC17 function 3}
\label{fig17_3}
\end{subfigure}
\begin{subfigure}[b]{0.49\textwidth}
\includegraphics[width=\linewidth]{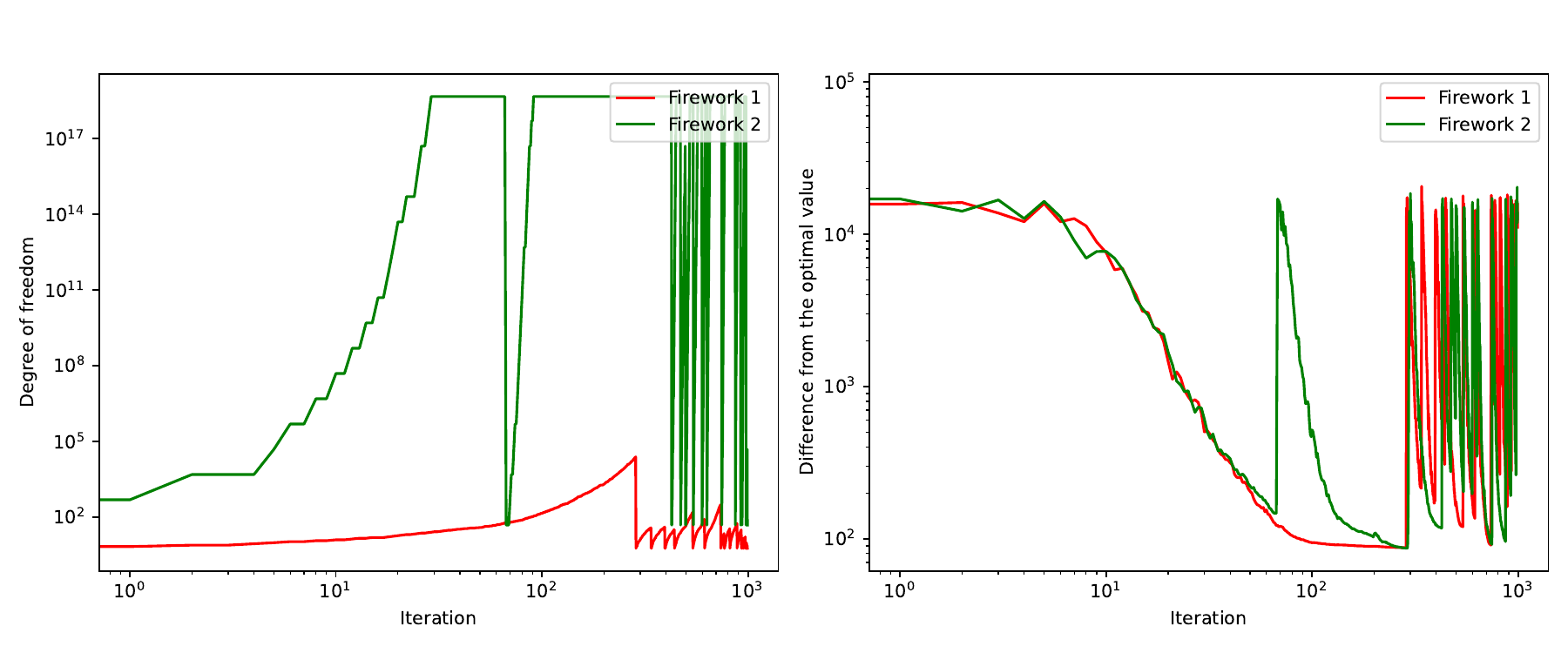}
\caption{CEC17 function 4}
\label{fig17_4}
\end{subfigure}
\begin{subfigure}[b]{0.49\textwidth}
\includegraphics[width=\linewidth]{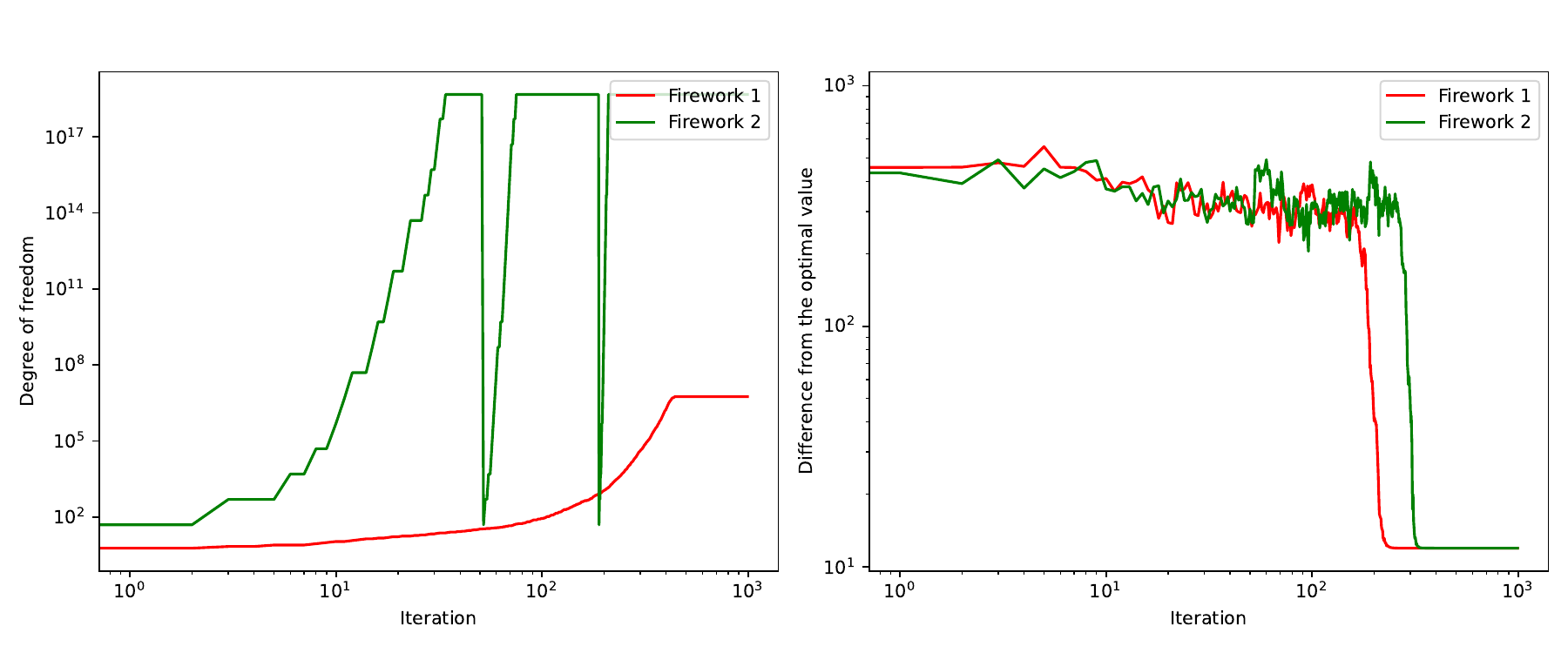}
\caption{CEC17 function 5}
\label{fig17_5}
\end{subfigure}
\begin{subfigure}[b]{0.49\textwidth}
\includegraphics[width=\linewidth]{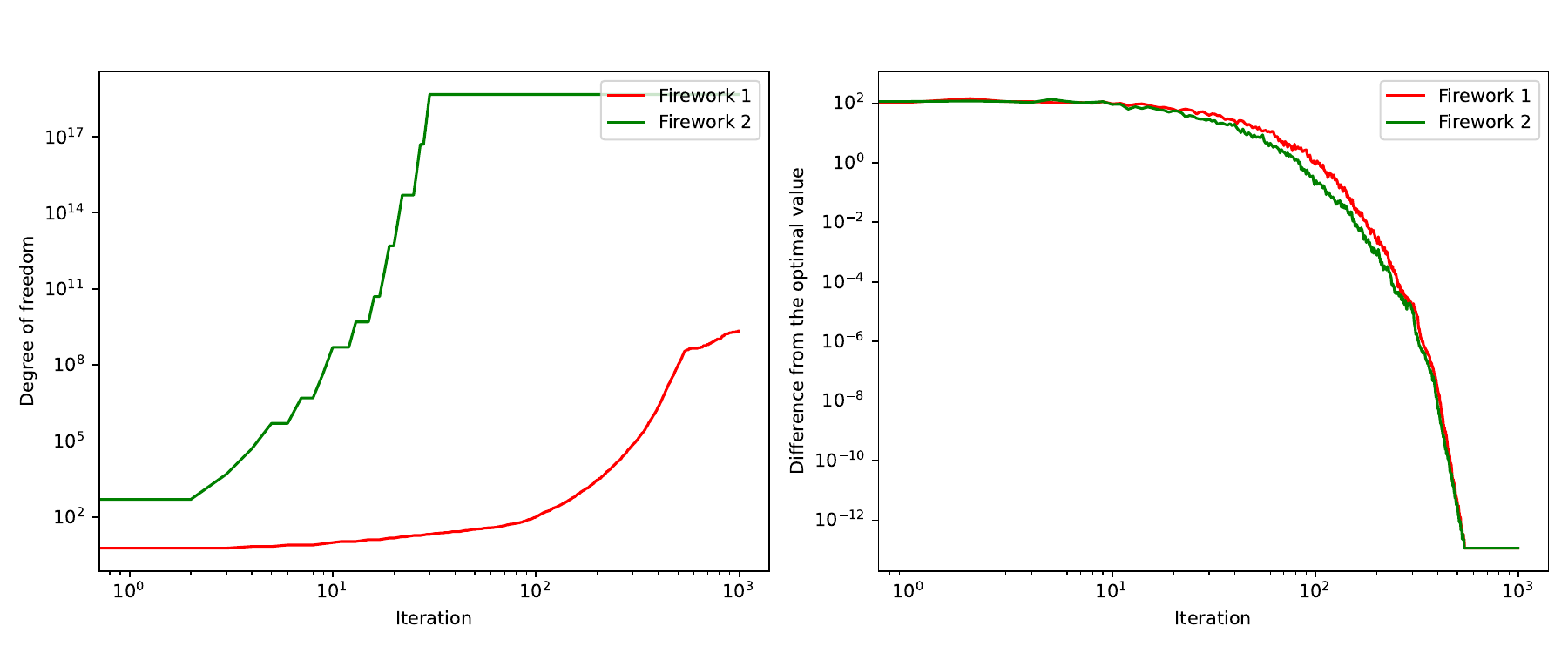}
\caption{CEC17 function 6}
\label{fig17_6}
\end{subfigure}
\begin{subfigure}[b]{0.49\textwidth}
\includegraphics[width=\linewidth]{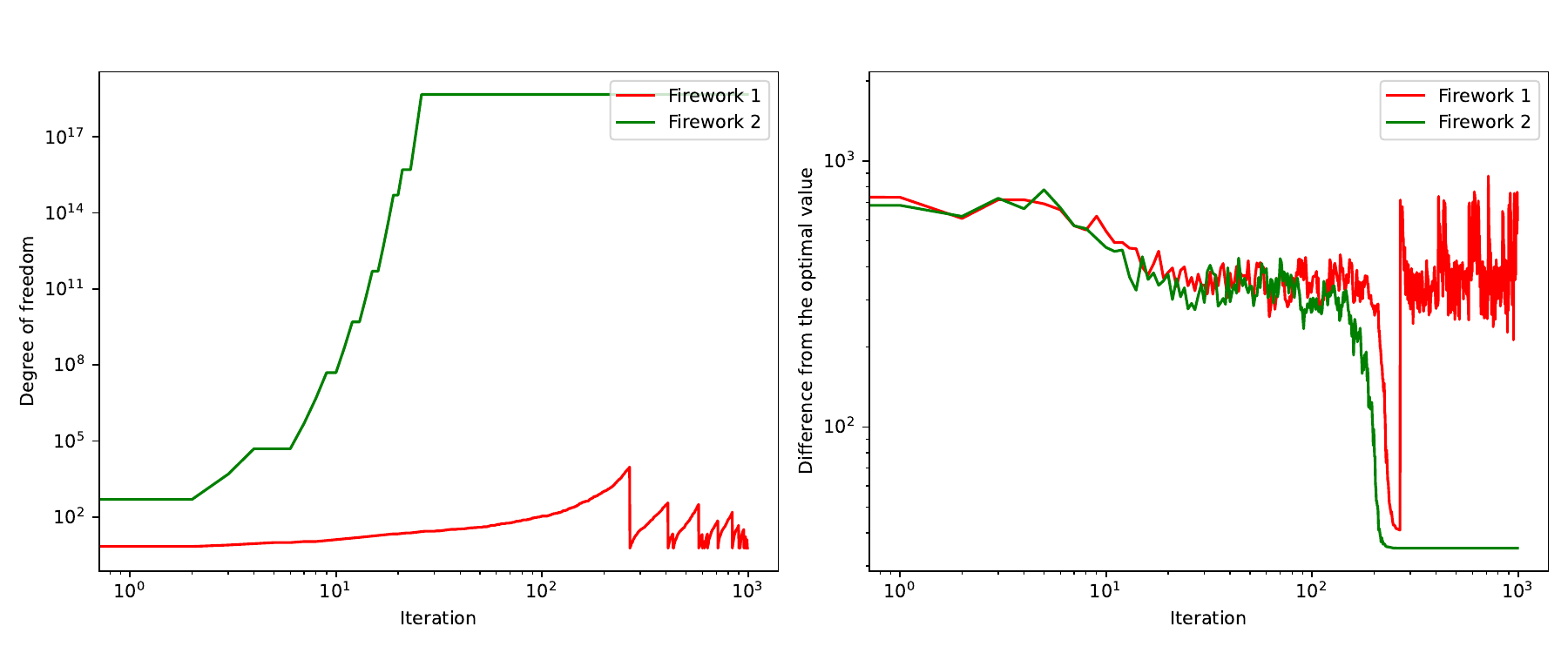}
\caption{CEC17 function 7}
\label{fig17_7}
\end{subfigure}
\begin{subfigure}[b]{0.49\textwidth}
\includegraphics[width=\linewidth]{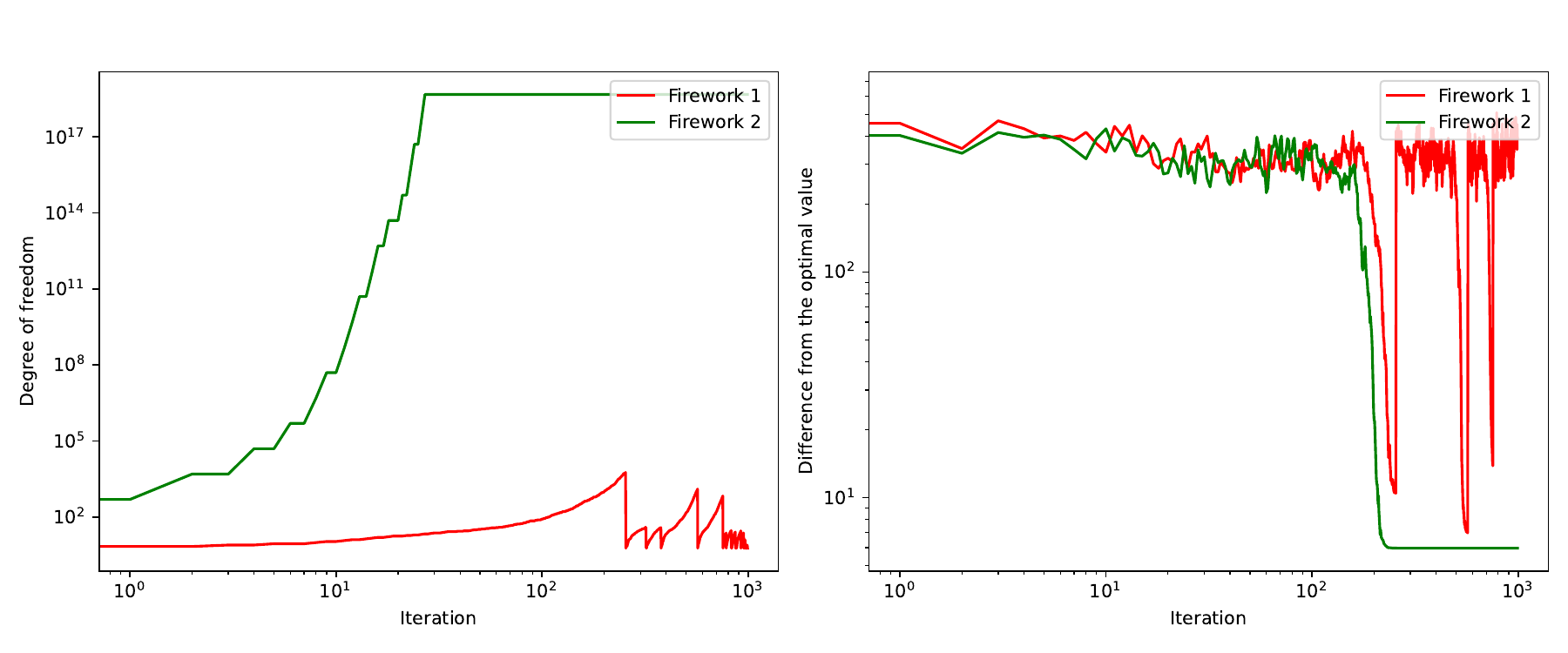}
\caption{CEC17 function 8}
\label{fig17_8}
\end{subfigure}
\caption{CEC2017 f1-f8}
\end{figure}

\begin{figure}[htbp]
\centering
\begin{subfigure}[b]{0.49\textwidth}
\includegraphics[width=\linewidth]{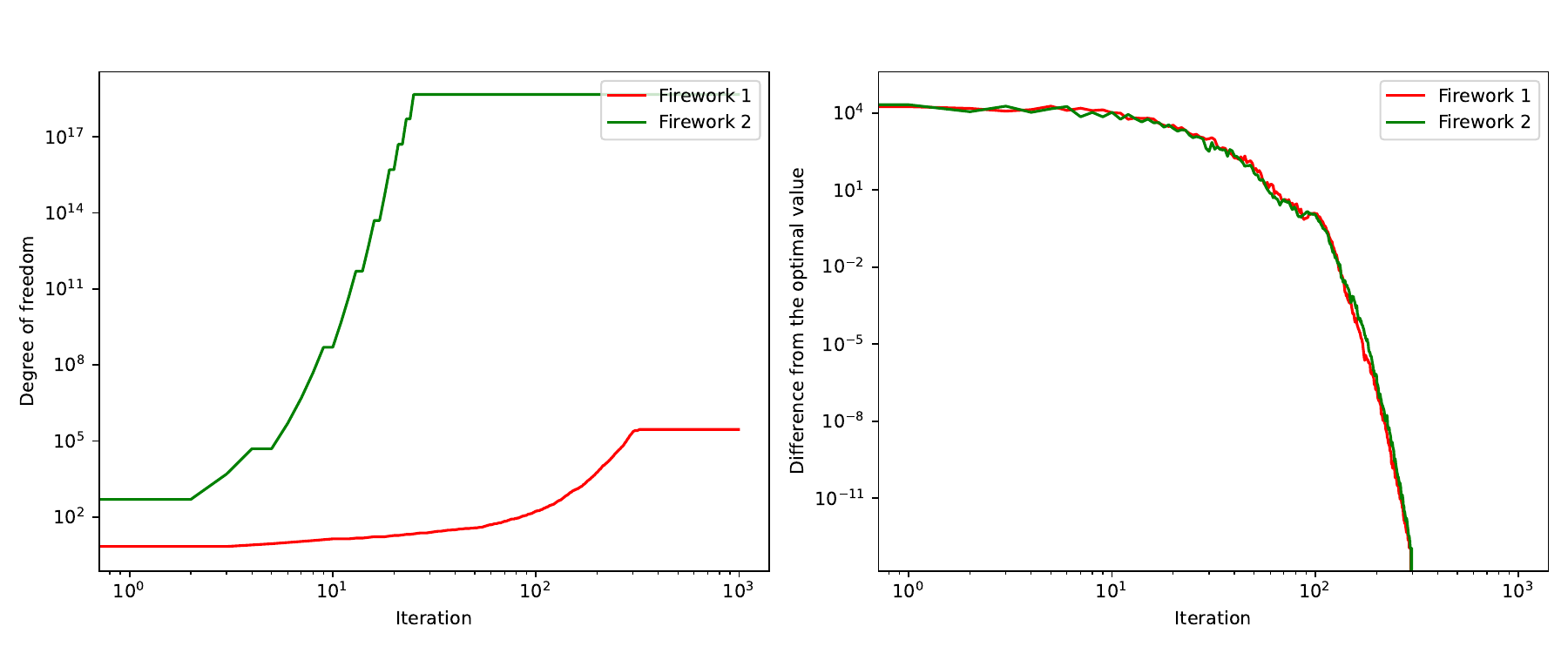}
\caption{CEC17 function 9}
\label{fig17_9}
\end{subfigure}
\begin{subfigure}[b]{0.49\textwidth}
\includegraphics[width=\linewidth]{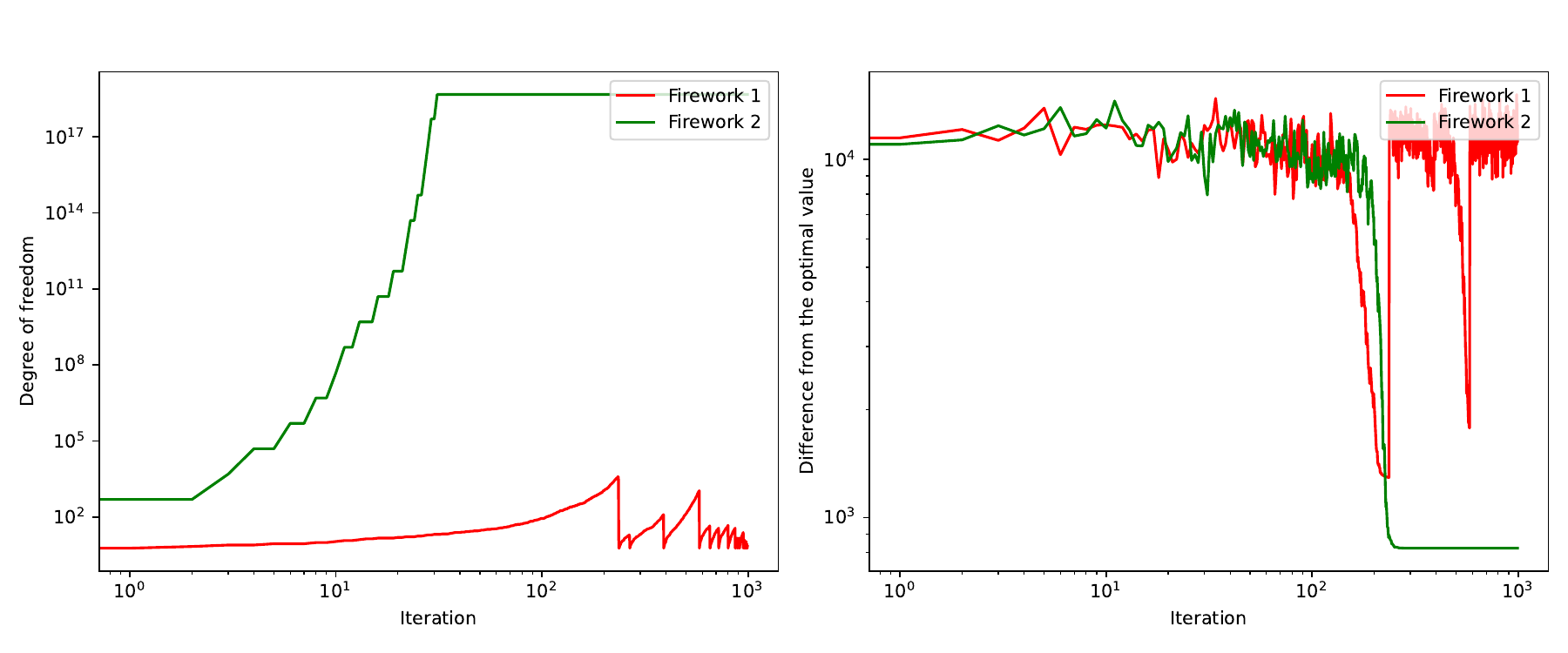}
\caption{CEC17 function 10}
\label{fig17_10}
\end{subfigure}
\begin{subfigure}[b]{0.49\textwidth}
\includegraphics[width=\linewidth]{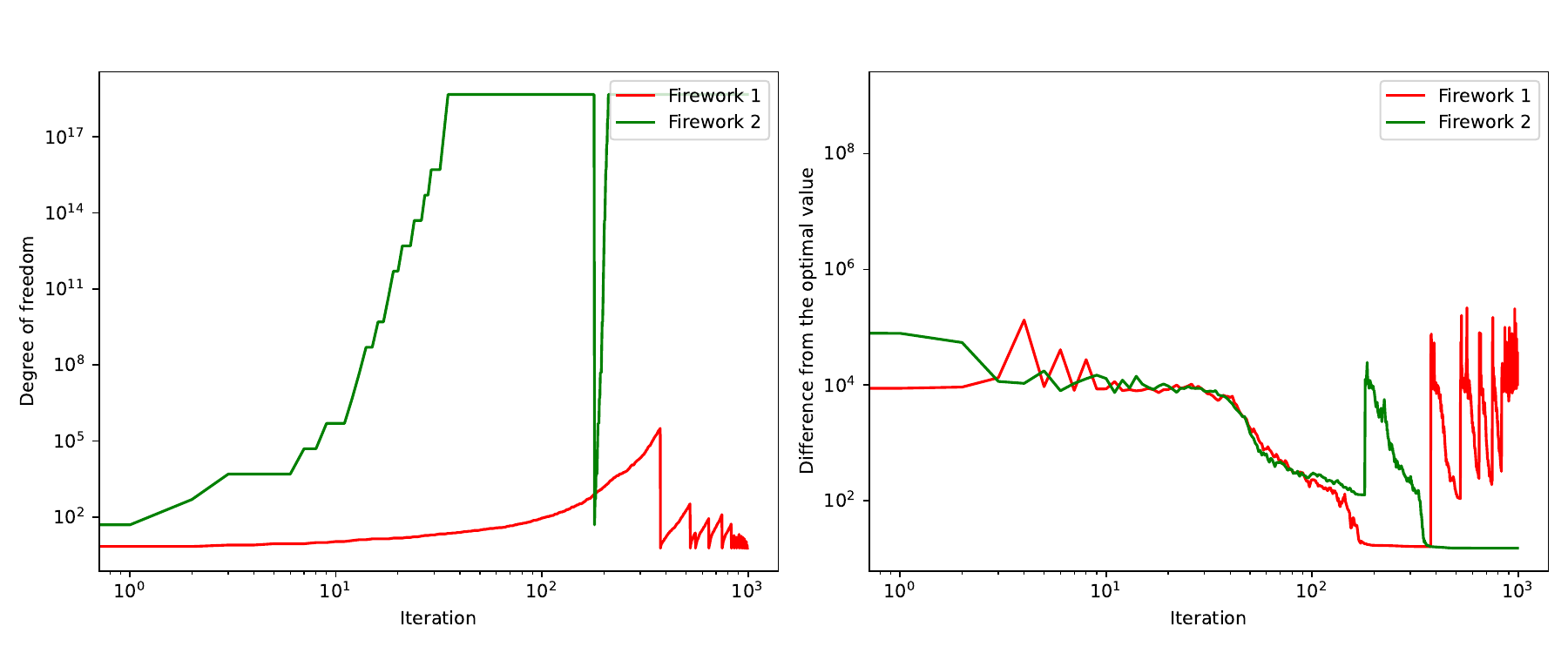}
\caption{CEC17 function 11}
\label{fig17_11}
\end{subfigure}
\begin{subfigure}[b]{0.49\textwidth}
\includegraphics[width=\linewidth]{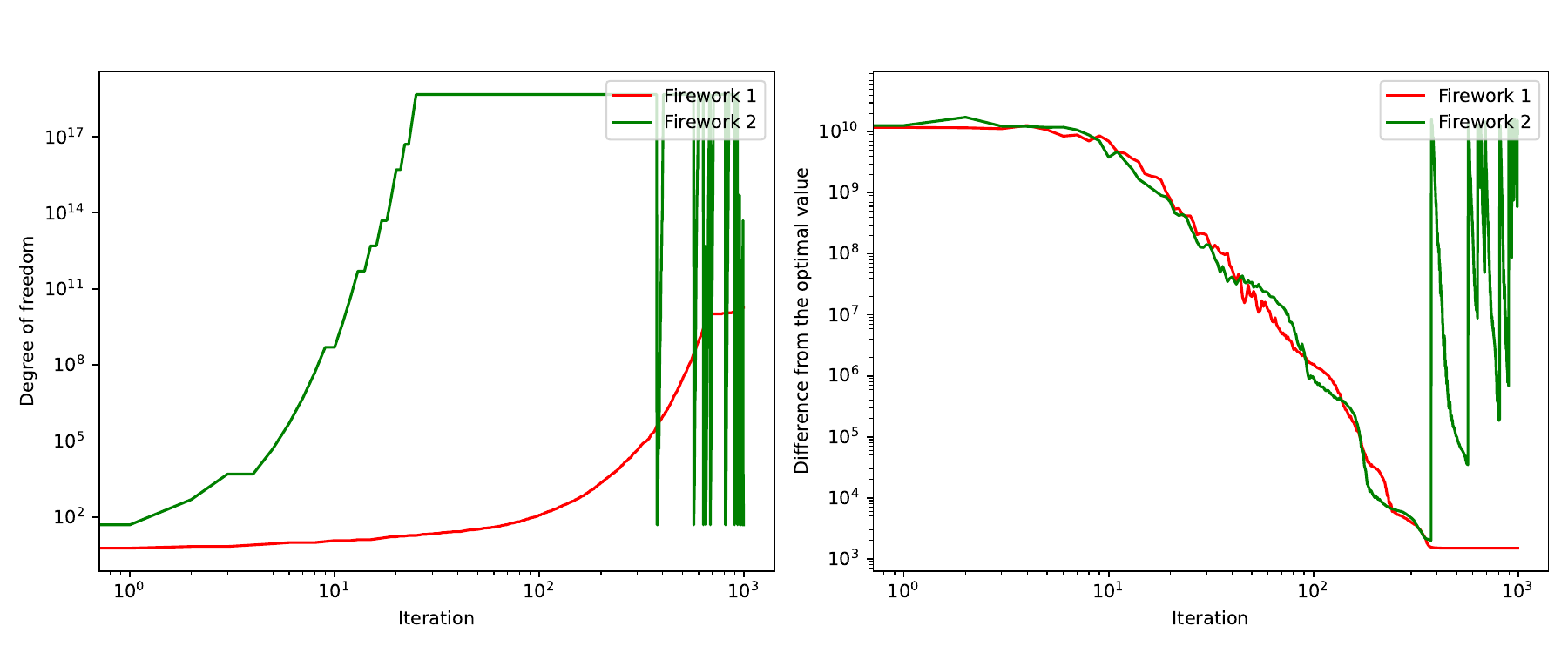}
\caption{CEC17 function 12}
\label{fig17_12}
\end{subfigure}
\begin{subfigure}[b]{0.49\textwidth}
\includegraphics[width=\linewidth]{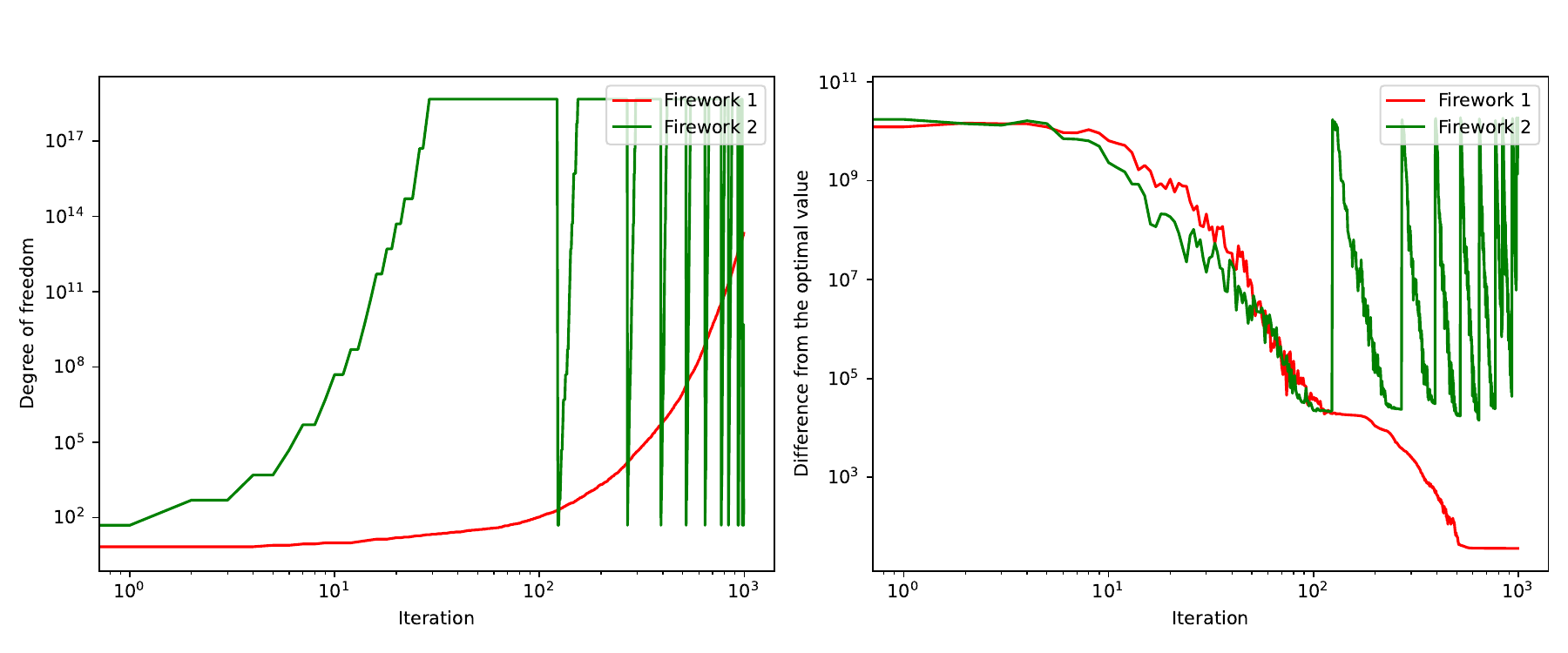}
\caption{CEC17 function 13}
\label{fig17_13}
\end{subfigure}
\begin{subfigure}[b]{0.49\textwidth}
\includegraphics[width=\linewidth]{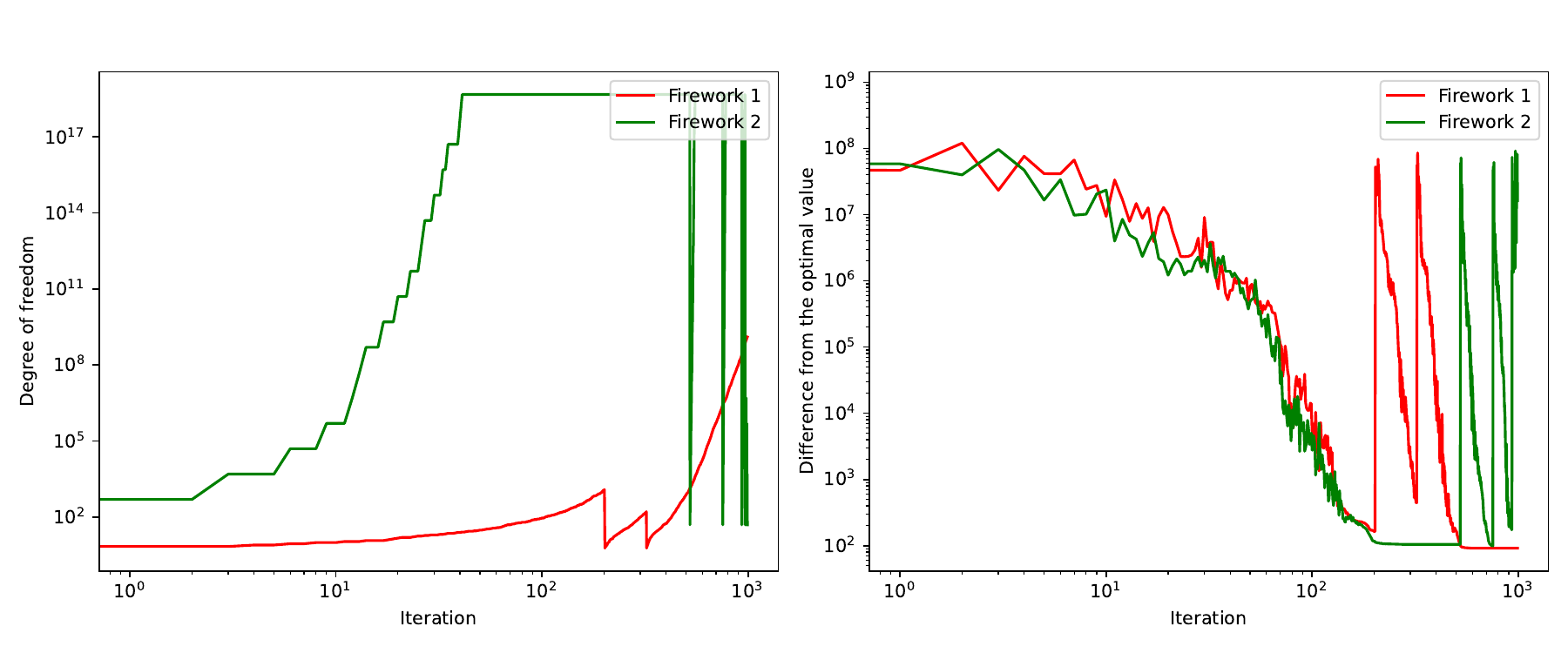}
\caption{CEC17 function 14}
\label{fig17_14}
\end{subfigure}
\begin{subfigure}[b]{0.49\textwidth}
\includegraphics[width=\linewidth]{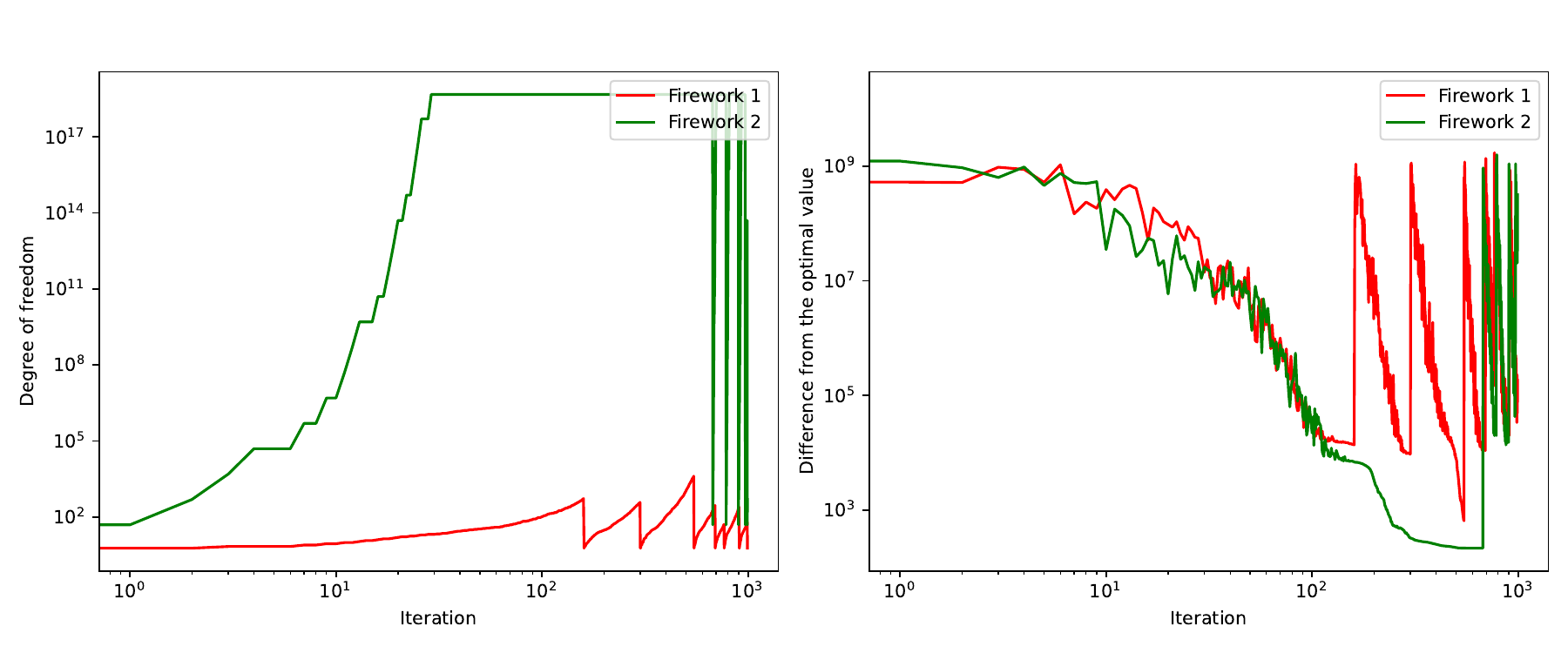}
\caption{CEC17 function 15}
\label{fig17_15}
\end{subfigure}
\begin{subfigure}[b]{0.49\textwidth}
\includegraphics[width=\linewidth]{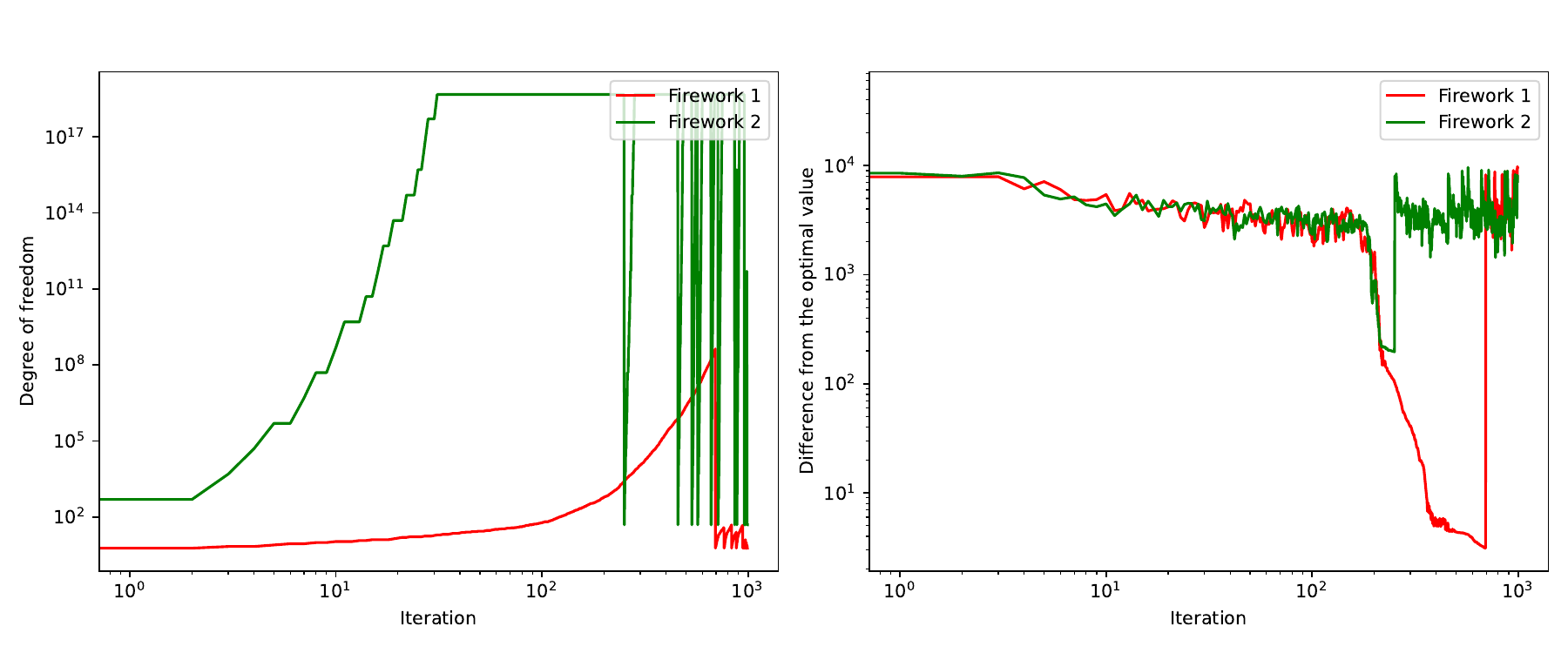}
\caption{CEC17 function 16}
\label{fig17_16}
\end{subfigure}
\caption{CEC2017 f9-f16}
\end{figure}

\begin{figure}[htbp]
\centering
\begin{subfigure}[b]{0.49\textwidth}
\includegraphics[width=\linewidth]{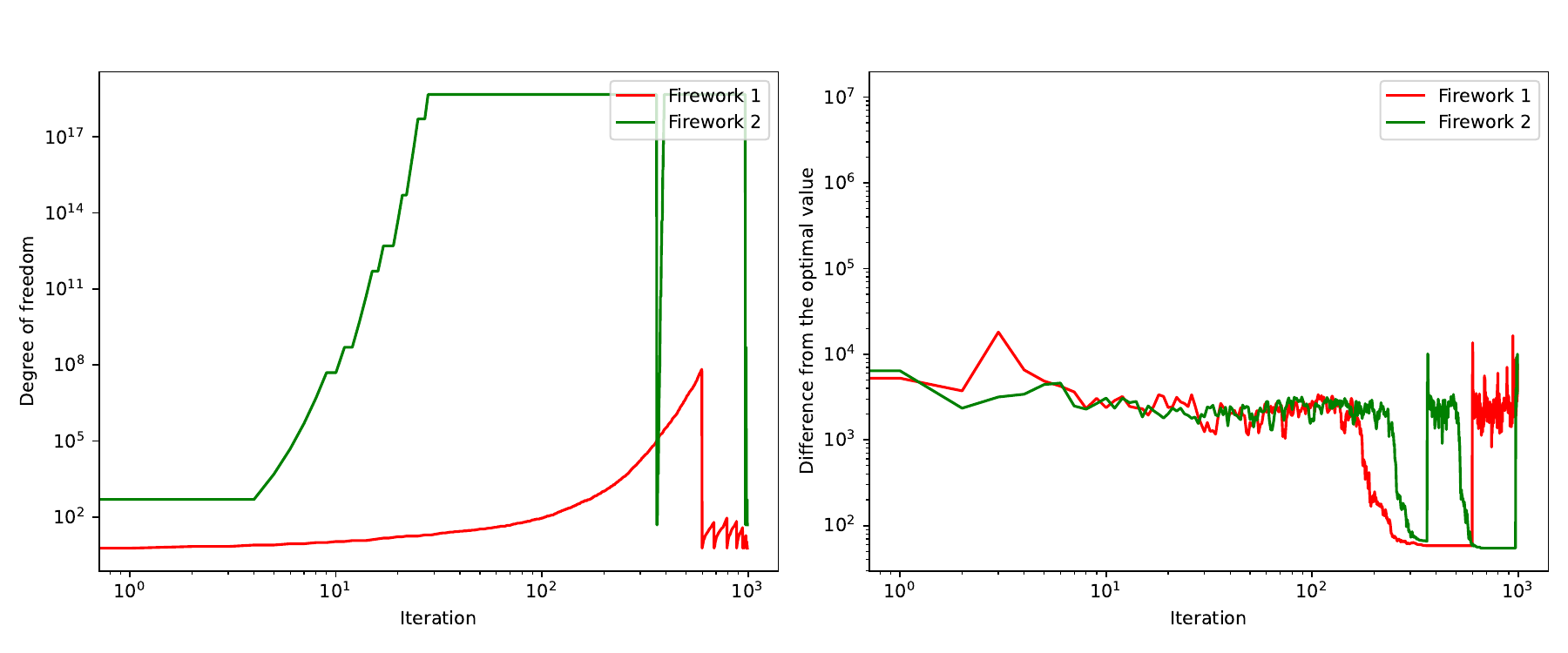}
\caption{CEC17 function 17}
\label{fig17_17}
\end{subfigure}
\begin{subfigure}[b]{0.49\textwidth}
\includegraphics[width=\linewidth]{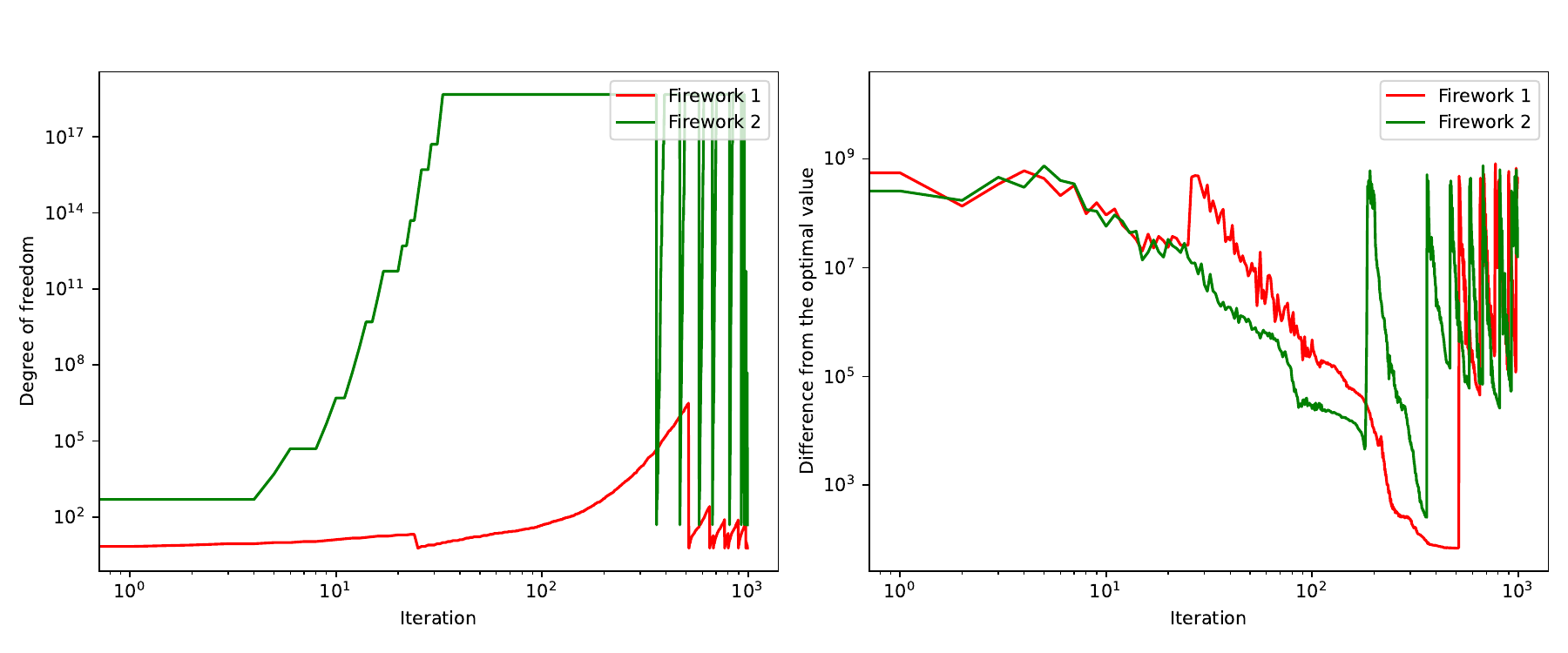}
\caption{CEC17 function 18}
\label{fig17_18}
\end{subfigure}
\begin{subfigure}[b]{0.49\textwidth}
\includegraphics[width=\linewidth]{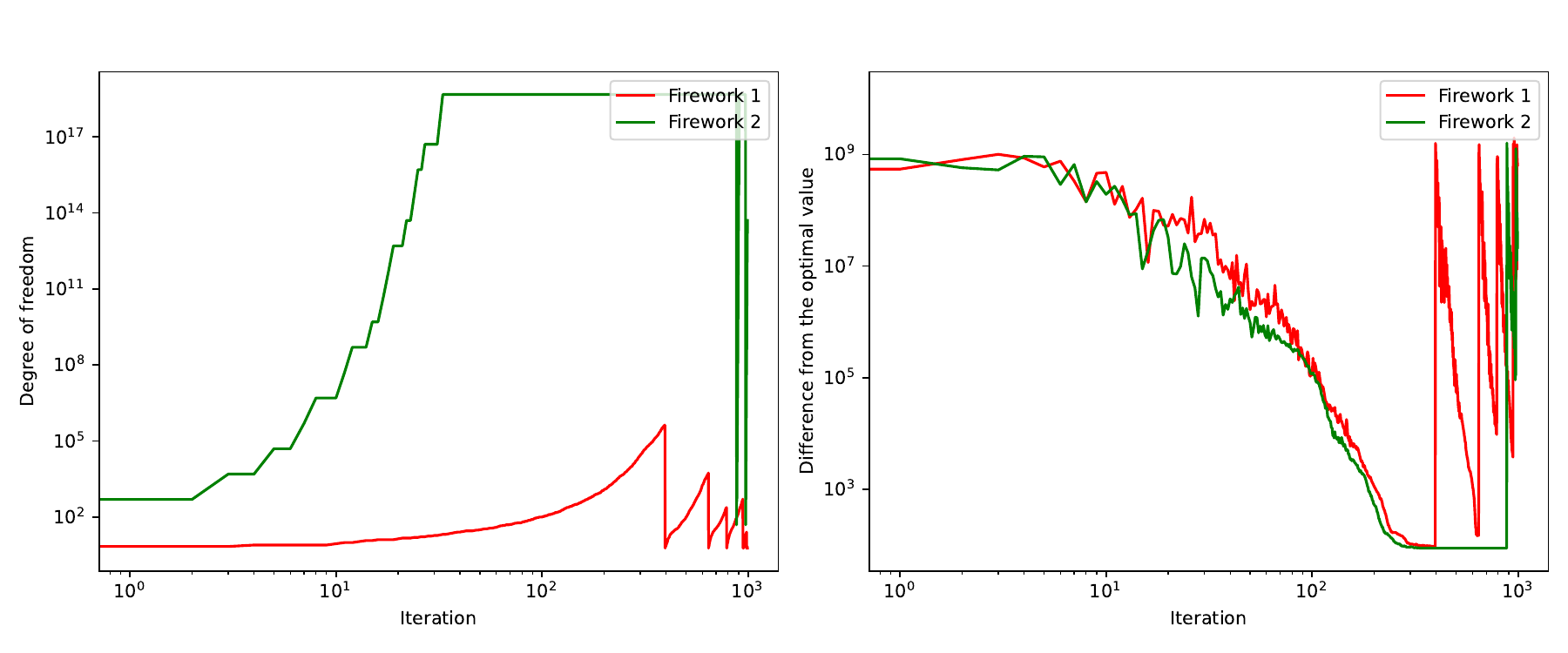}
\caption{CEC17 function 19}
\label{fig17_19}
\end{subfigure}
\begin{subfigure}[b]{0.49\textwidth}
\includegraphics[width=\linewidth]{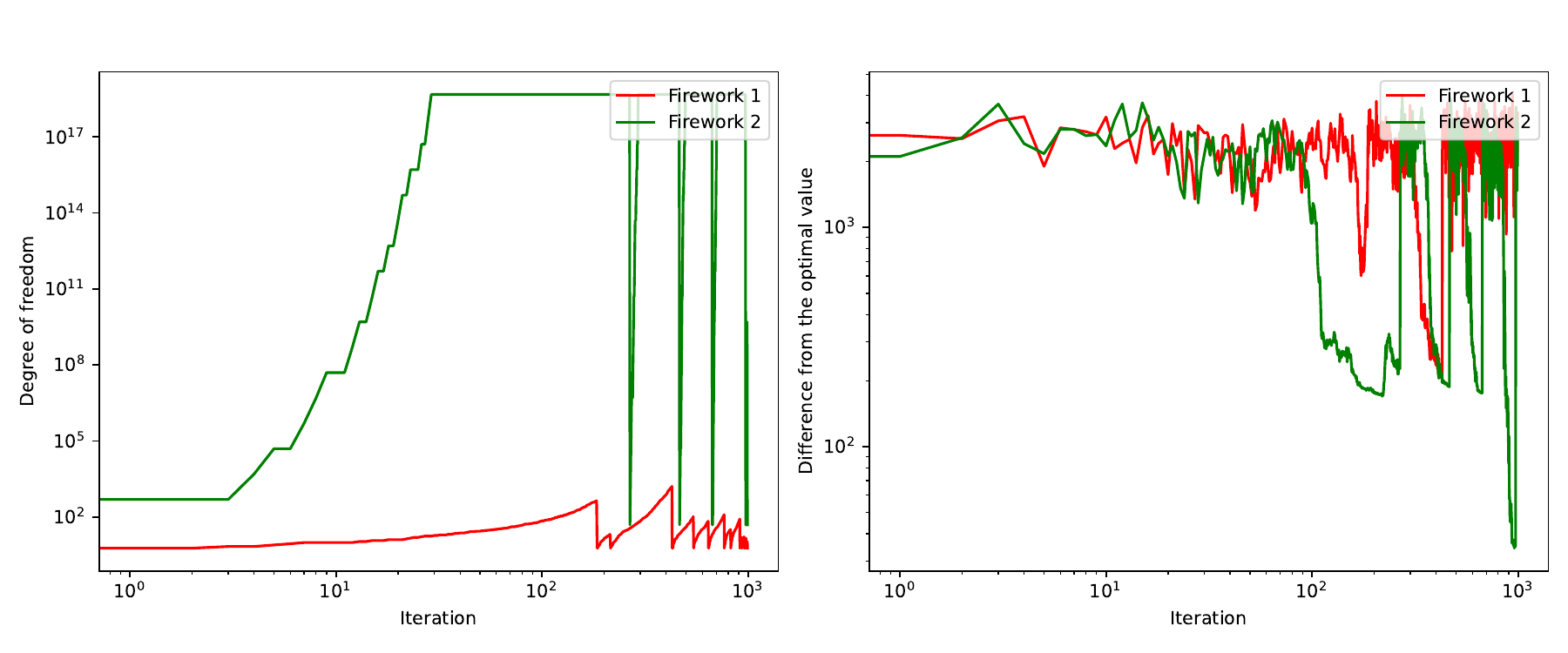}
\caption{CEC17 function 20}
\label{fig17_20}
\end{subfigure}
\begin{subfigure}[b]{0.49\textwidth}
\includegraphics[width=\linewidth]{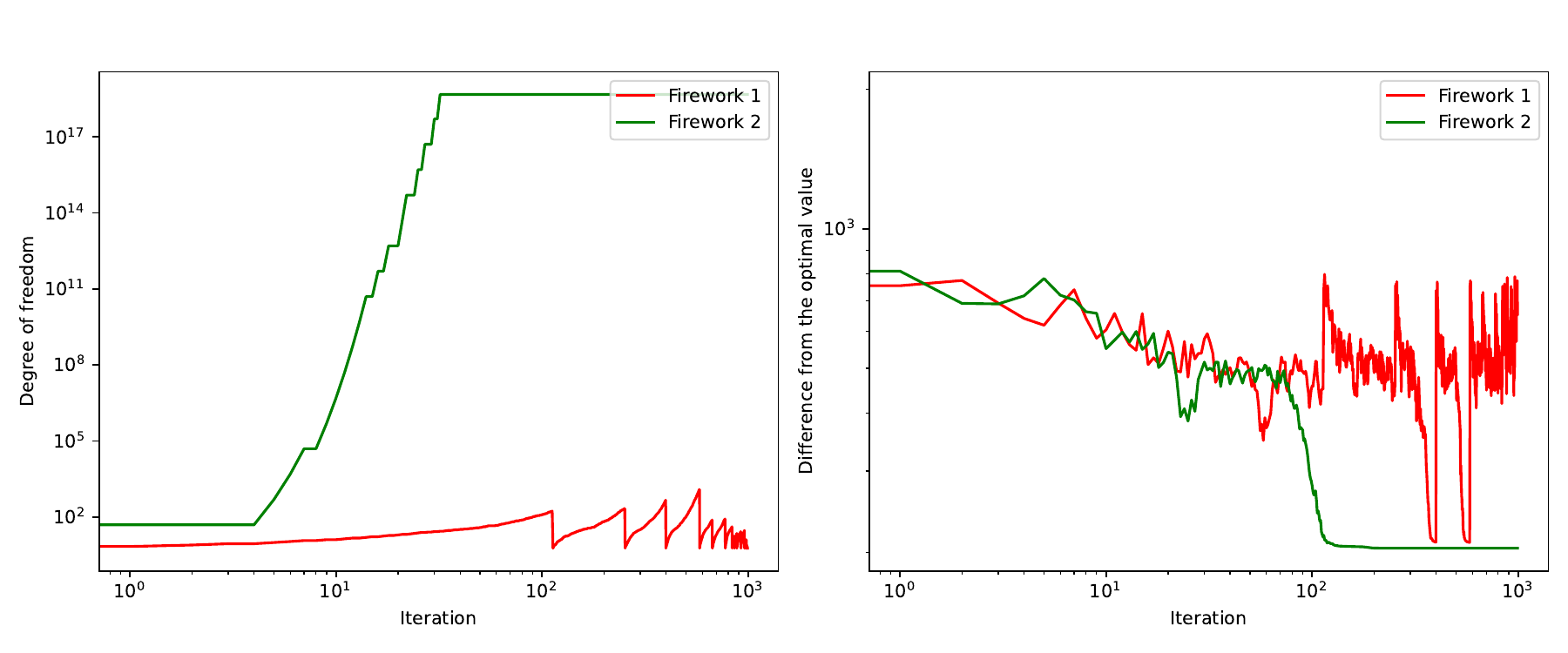}
\caption{CEC17 function 21}
\label{fig17_21}
\end{subfigure}
\begin{subfigure}[b]{0.49\textwidth}
\includegraphics[width=\linewidth]{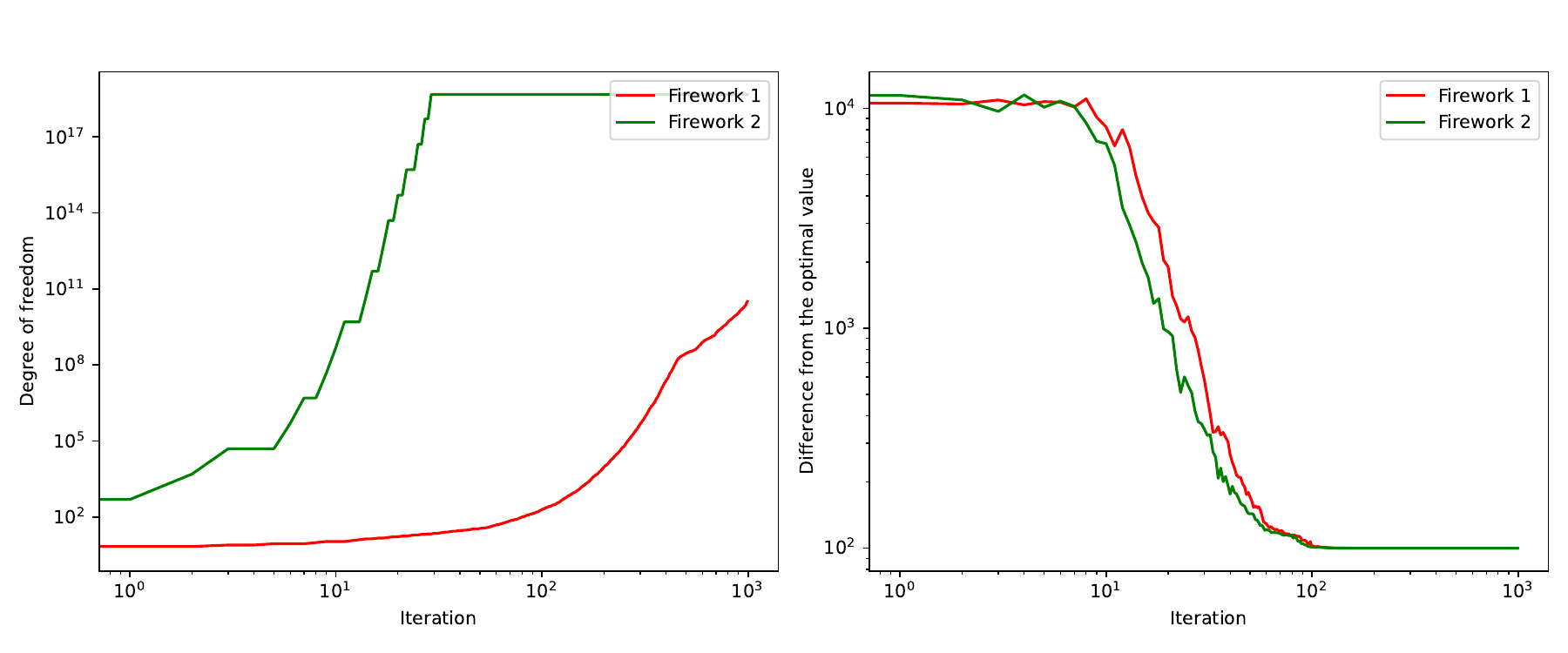}
\caption{CEC17 function 22}
\label{fig17_22}
\end{subfigure}
\begin{subfigure}[b]{0.49\textwidth}
\includegraphics[width=\linewidth]{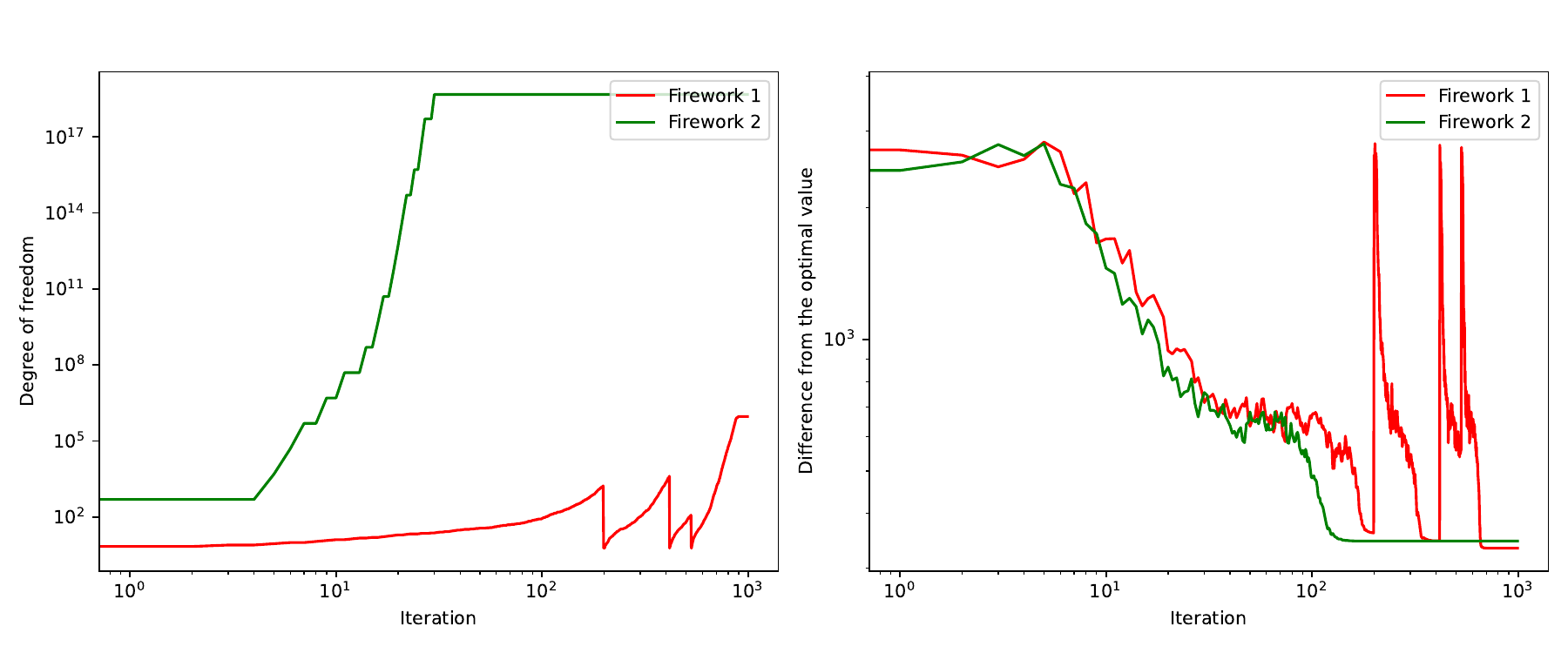}
\caption{CEC17 function 23}
\label{fig17_23}
\end{subfigure}
\begin{subfigure}[b]{0.49\textwidth}
\includegraphics[width=\linewidth]{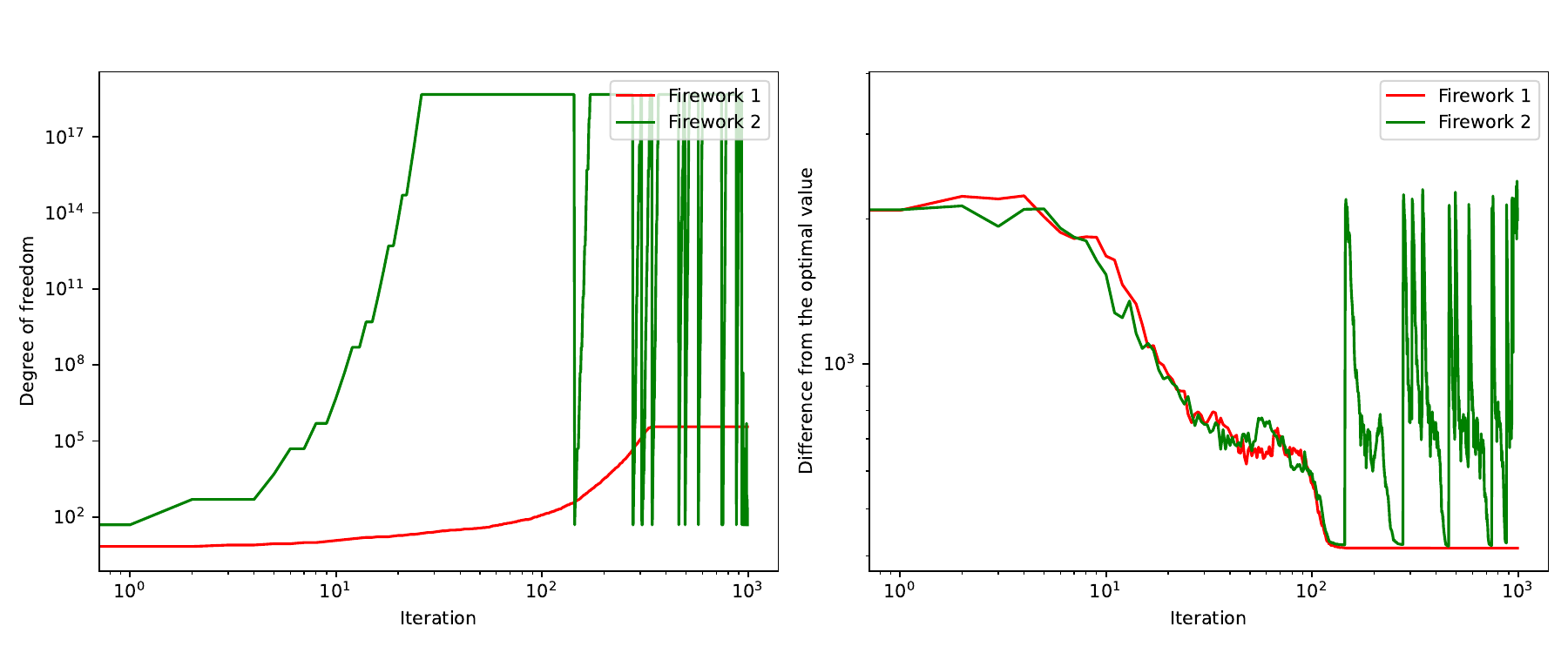}
\caption{CEC17 function 24}
\label{fig17_24}
\end{subfigure}
\caption{CEC2017 f17-f24}
\end{figure}

\begin{figure}[htbp]
\centering
\begin{subfigure}[b]{0.49\textwidth}
\includegraphics[width=\linewidth]{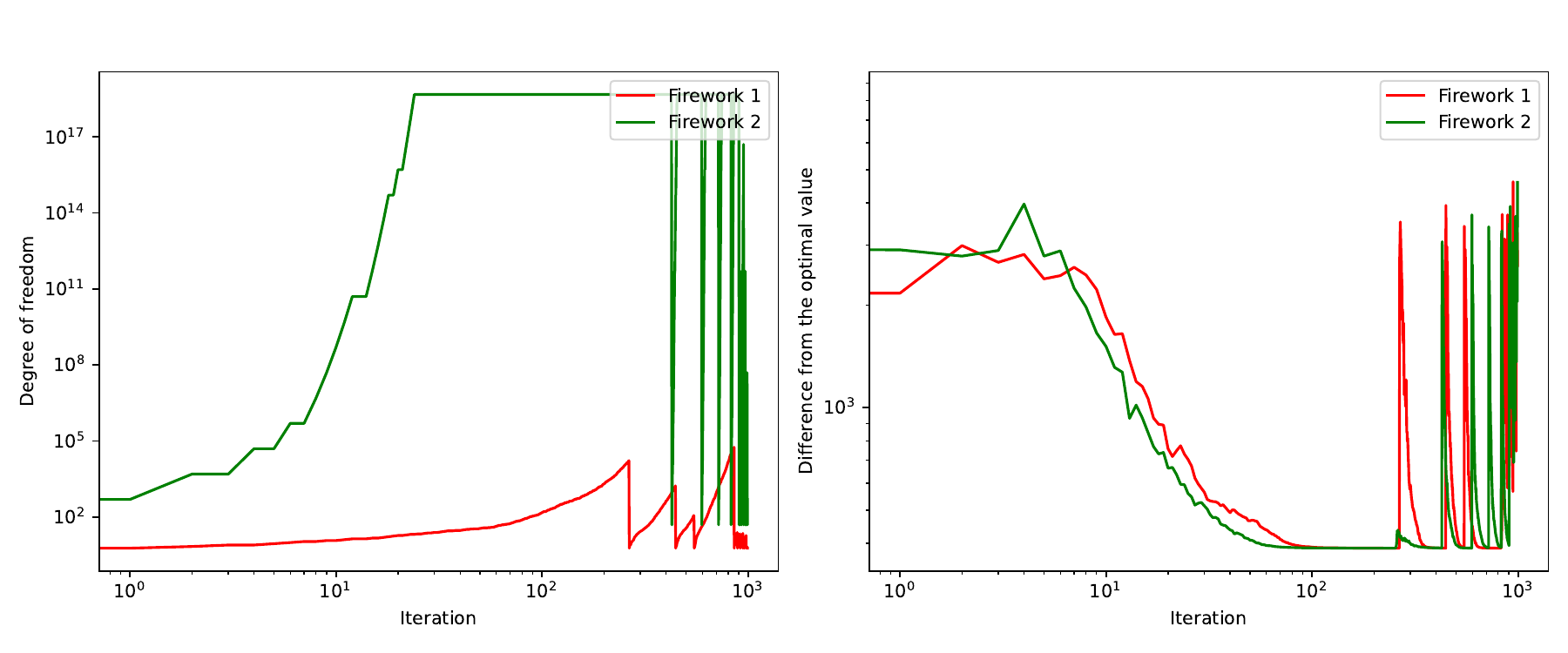}
\caption{CEC17 function 25}
\label{fig17_25}
\end{subfigure}
\begin{subfigure}[b]{0.49\textwidth}
\includegraphics[width=\linewidth]{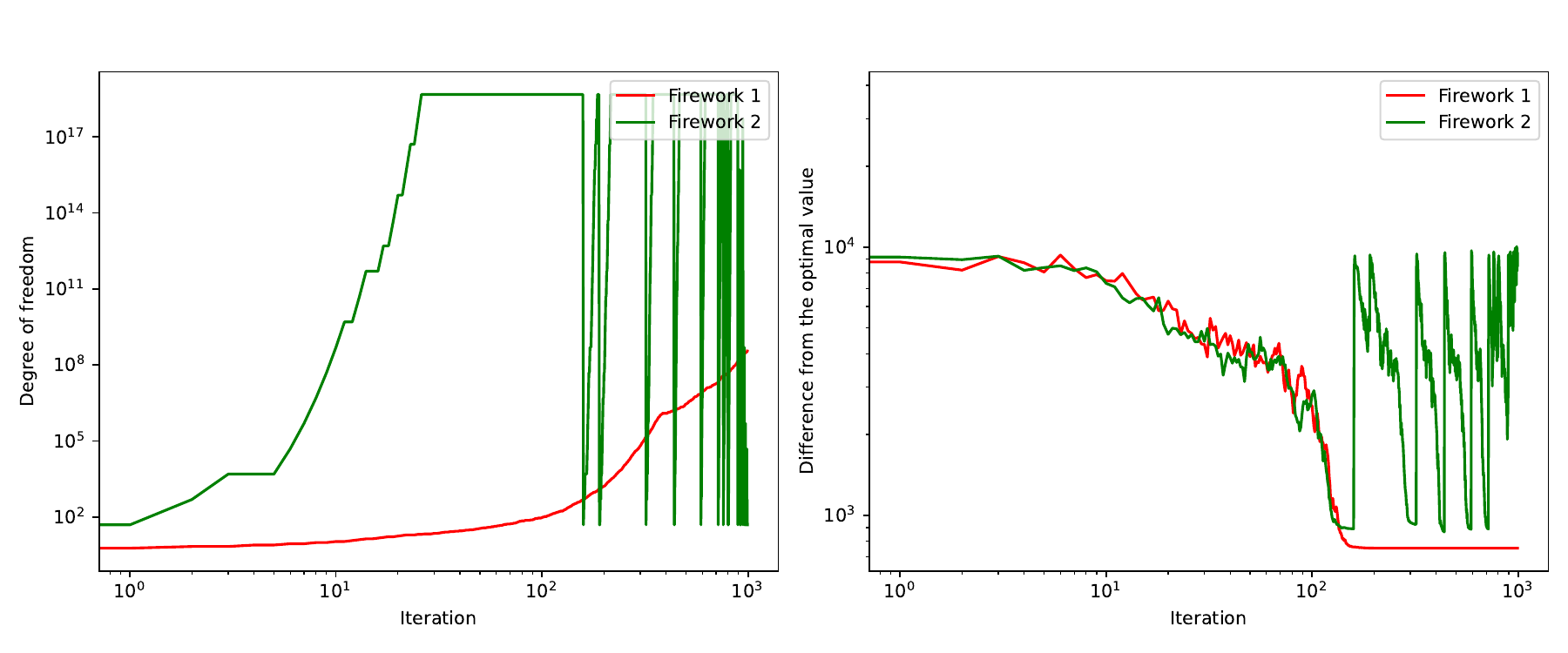}
\caption{CEC17 function 26}
\label{fig17_26}
\end{subfigure}
\begin{subfigure}[b]{0.49\textwidth}
\includegraphics[width=\linewidth]{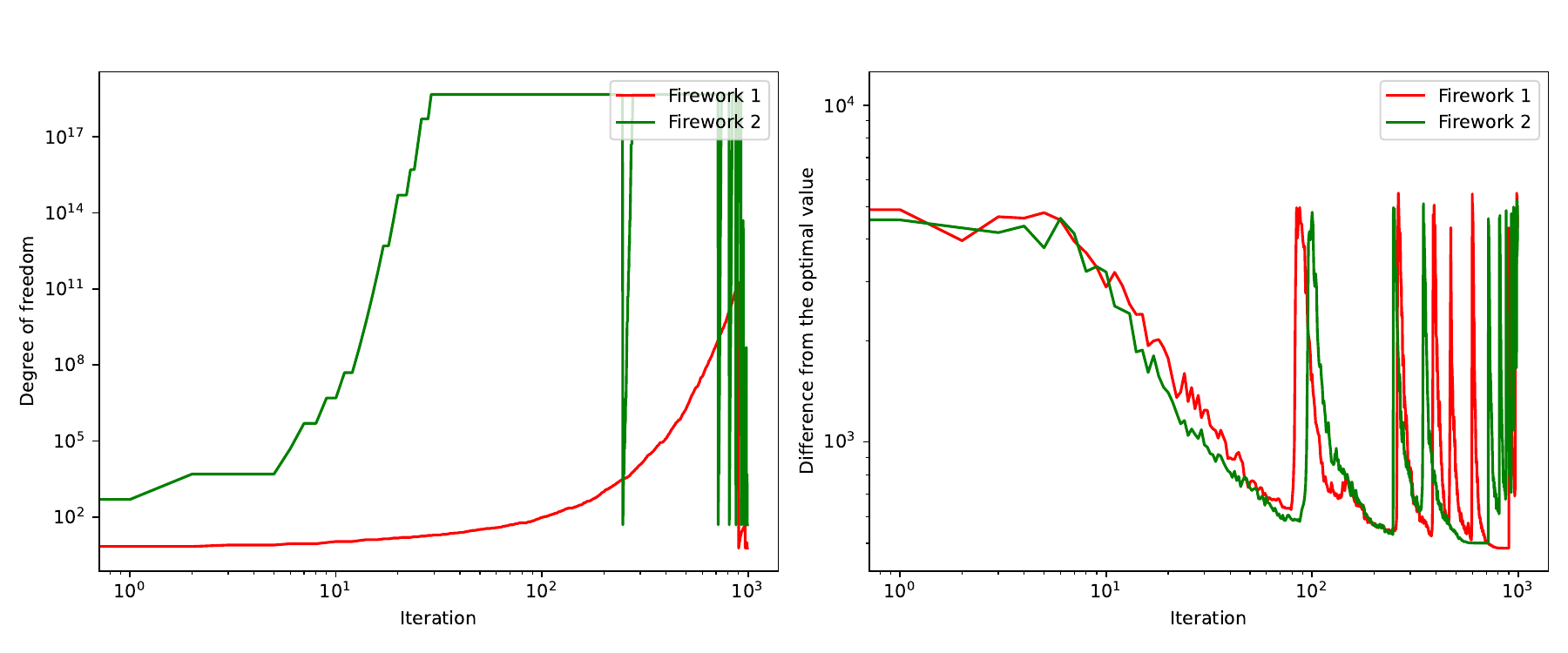}
\caption{CEC17 function 27}
\label{fig17_27}
\end{subfigure}
\begin{subfigure}[b]{0.49\textwidth}
\includegraphics[width=\linewidth]{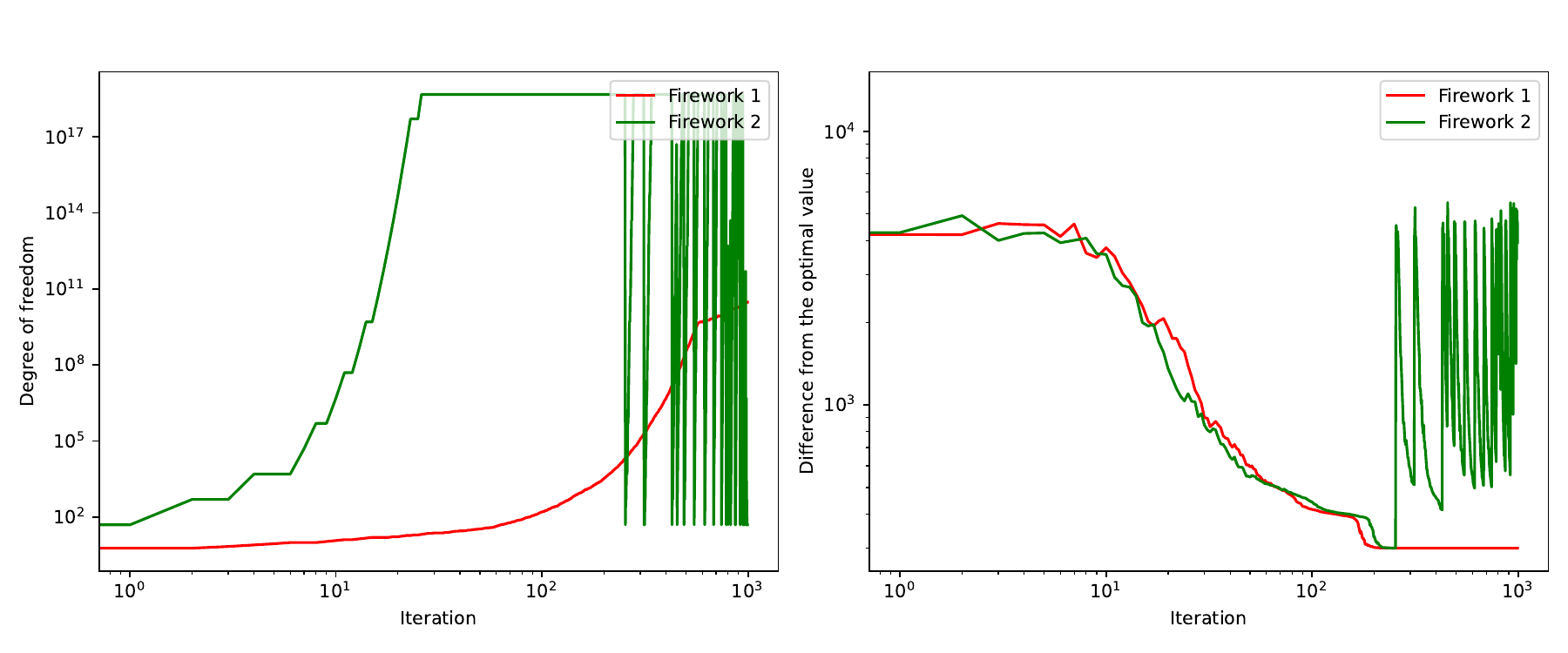}
\caption{CEC17 function 28}
\label{fig17_28}
\end{subfigure}
\begin{subfigure}[b]{0.49\textwidth}
\includegraphics[width=\linewidth]{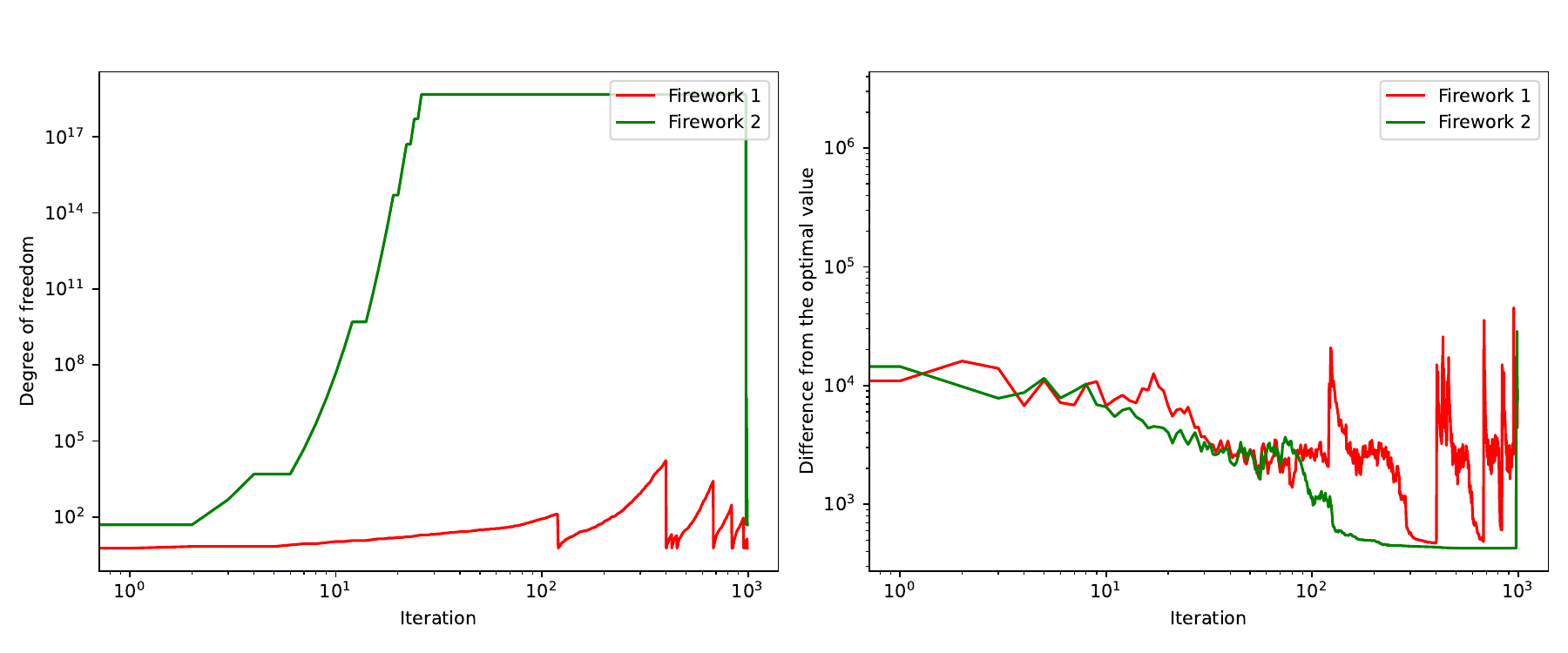}
\caption{CEC17 function 29}
\label{fig17_29}
\end{subfigure}
\begin{subfigure}[b]{0.49\textwidth}
\includegraphics[width=\linewidth]{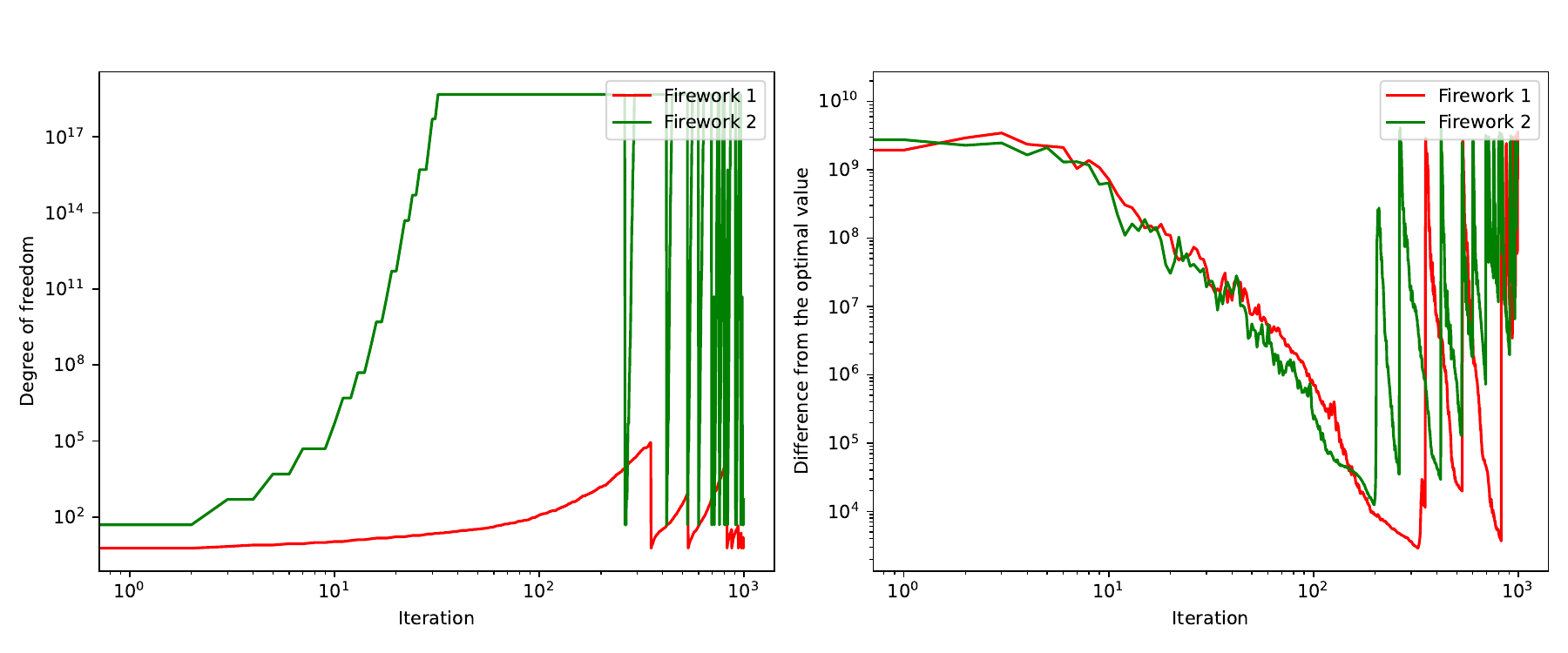}
\caption{CEC17 function 30}
\label{fig17_30}
\end{subfigure}
\caption{CEC2017 f25-f30}
\end{figure}

\end{document}